\documentclass[11pt]{article} 

\usepackage{hyperref}
\usepackage{url}
\usepackage{multirow}

\usepackage{amsfonts,epsfig,graphicx}
\usepackage{amsmath,amssymb,amsthm}
\usepackage{fullpage}
\usepackage{hhline}
\usepackage{sidecap}

\usepackage{epsf}
\usepackage{graphics}
\usepackage{amsfonts}
\usepackage{amsmath}
\usepackage{amsthm}
\usepackage{psfrag}

\usepackage{enumitem}
\usepackage{caption}
\usepackage{subcaption}

\usepackage{xcolor,colortbl}

\usepackage{wrapfig}
\usepackage{fullpage}
\usepackage{picinpar}

\let\oldbibliography\thebibliography
\renewcommand{\thebibliography}[1]{%
  \oldbibliography{#1}%
  \setlength{\itemsep}{0pt}%
}

\newtheorem{theorem}{Theorem} 
\newtheorem{lemma}{Lemma}
\newtheorem{proposition}{Proposition}

\newcommand{\delcrit}{\ensuremath{\delta_n}}

\newcommand{\Ball}{\ensuremath{\mathbb{B}}}
\newcommand{\BallF}{\ensuremath{\Ball_F}}

\newcommand{\DelHat}{\ensuremath{\widehat{\Delta}}}
\newcommand{\DelHatPi}{\ensuremath{\widehat{\Delta}_{\pi}}}

\newcommand{\tracer}[2]{\ensuremath{\langle \!\langle {#1}, \; {#2}
\rangle \!\rangle}}

\newcommand{\defn}{\ensuremath{:\,=}}
\newcommand{\Lnorm}[2]{\ensuremath{\matsnorm{#1}{\mbox{\tiny{#2}}}}}
\newcommand{\myLnorm}[2]{\ensuremath{\|#1\|_{#2}}}
\newcommand{\inprod}[2]{\ensuremath{\langle #1 , \, #2 \rangle}}
\newcommand{\Exs}{\ensuremath{\mathbb{E}}}

\newcommand{\argmin}{\operatornamewithlimits{arg~min}}

\newcommand{\identity}{\ensuremath{I}}
\newcommand{\kl}[2]{\ensuremath{D_{\mathrm{KL}}(#1\|#2)}}
\newcommand{\reals}{\ensuremath{\mathbb{R}}}
\newcommand{\mprob}{\ensuremath{\mathbb{P}}}
\newcommand{\hamming}{\ensuremath{\dham}}
\newcommand{\half}{\ensuremath{{\frac{1}{2}}}}
\newcommand{\colordiag}{\cellcolor{black!50}\half}

\newcommand{\marginalclass}{\ensuremath{\mathbb{C}_{\mbox{\scalebox{.5}{{FULL}}}}}}
\newcommand{\bias}{\ensuremath{\gamma}}
\newcommand{\noise}{\ensuremath{\Wmat}}
\newcommand{\covnum}{\ensuremath{N}}
\newcommand{\metent}{\ensuremath{\log \covnum}}
\newcommand{\wt}{\ensuremath{M}}
\newcommand{\wtstar}{\ensuremath{\wt^*}}
\newcommand{\wthat}{\ensuremath{\widehat{\wt}}}
\newcommand{\wthatUSVT}{\ensuremath{{\Mhat}_{\regparn}}}
\newcommand{\wthatUSVTp}{\ensuremath{{\Mhat}_{\regparnp}}}
\newcommand{\numitems}{\ensuremath{n}}
\newcommand{\obs}{\ensuremath{Y}}
\newcommand{\monoGW}{\ensuremath{{\cal F}}}
\newcommand{\permall}{\ensuremath{\Pi}}
\newcommand{\perm}{\ensuremath{\pi}}

\newcommand{\permhatclass}{\ensuremath{\widehat{\Pi}}}

\newcommand{\chatterjeeclass}{\ensuremath{\mathbb{C}_{\mbox{\scalebox{.5}{{SST}}}}}}
\newcommand{\paramclass}{\ensuremath{\mathbb{C}_{\mbox{\scalebox{.5}{{PAR}}}}}}

\newcommand{\chattRelaxed}{\ensuremath{\mathbb{C}_{\mbox{\scalebox{.5}{{BISO}}}}}}

\newcommand{\bisoclass}{\ensuremath{\chattRelaxed}}
\newcommand{\bisoclasssiva}[1]{\ensuremath{\chattRelaxed(#1)}}

 \newcommand{\plaincon}{c}
\newcommand{\wtparam}{w}
\newcommand{\wtparamstar}{\ensuremath{\wtparam^*}}

\newcommand{\wtparamhatML}{\ensuremath{\widehat{\wtparam}_{\mbox{\tiny{ML}}}}}

\newcommand{\packnum}{\ensuremath{T}}

 \newcommand{\ones}{\ensuremath{1}}
\newcommand{\packdmin}{\ensuremath{\alpha}}
\newcommand{\diff}{\ensuremath{x}} \newcommand{\cdf}{\ensuremath{F}}
\newcommand{\wmax}{\ensuremath{1}}

\newcommand{\glmcdf}{\ensuremath{F}}

\newcommand{\cardinality}[1]{\ensuremath{| #1 |}}
\newcommand{\factorial}[1]{\ensuremath{#1!}}

\newcommand{\AuxEvent}{\ensuremath{\mathcal{A}_t}}
\newcommand{\sign}{\mbox{sign}}

\newcommand{\numobs}{\numitems}

\makeatletter
\long\def\@makecaption#1#2{
        \vskip 0.8ex
        \setbox\@tempboxa\hbox{\small {\bf #1:} #2}
        \parindent 1.5em  
        \dimen0=\hsize
        \advance\dimen0 by -3em
        \ifdim \wd\@tempboxa >\dimen0
                \hbox to \hsize{
                        \parindent 0em
                        \hfil 
                        \parbox{\dimen0}{\def\baselinestretch{0.96}\small
                                {\bf #1.} #2
                                } 
                        \hfil}
        \else \hbox to \hsize{\hfil \box\@tempboxa \hfil}
        \fi
        }
\makeatother

\makeatletter \newenvironment{subtheorem}[1]{%
  \def\subtheoremcounter{#1}
  \protected@edef\theparentnumber{\csname the#1\endcsname}%
  \setcounter{parentnumber}{\value{#1}}
  \expandafter\def\csname the#1\endcsname{\theparentnumber.\Alph{#1}}%
  \ignorespaces }{%
  \setcounter{\subtheoremcounter}{\value{parentnumber}}%
  \ignorespacesafterend } \makeatother \newcounter{parentnumber}

\newcommand{\modStocTranClass}{\ensuremath{\mathbb{C}_{\mbox{\scalebox{.5}{{MST}}}}}}
\newcommand{\weakStocTranClass}{\ensuremath{\mathbb{C}_{\mbox{\scalebox{.5}{{WST}}}}}}
\newcommand{\chattDiffmx}{\ensuremath{D}}
\newcommand{\chattDiff}{\ensuremath{\mathbb{C}_{\mbox{\scalebox{.5}{{DIFF}}}}}}

\newcommand{\monoGWfn}{\ensuremath{g}}

\newcommand{\biisomx}{\ensuremath{\wt}}

\newcommand{\pp}{\ensuremath{p_{\mathrm{obs}}}}

\newcommand{\chatterjeeclassbias}{\ensuremath{\mathbb{C}_{\mbox{\scalebox{.5}{{HIGH}}}}}}

\newcommand{\matsnorm}[2]{|\!|\!| #1 | \! | \!|_{{#2}}}

\newcommand{\numitem}{\ensuremath{n}}
\newcommand{\order}{\ensuremath{\mathcal{O}}}

\long\def\comment#1{}

\newcommand{\rank}{\ensuremath{\operatorname{rank}}}
\newcommand{\indicator}[1]{\ensuremath{\mathbf{1}\{#1\}}}
\newcommand{\diag}[1]{\ensuremath{\mbox{diag}(#1)}}
\newcommand{\opnorm}[1]{\ensuremath{\matsnorm{#1}{\mbox{\tiny{op}}}}}
\newcommand{\frobnorm}[1]{\ensuremath{\matsnorm{#1}{\mbox{\tiny{F}}}}}

\newcommand{\Mhat}{\ensuremath{\widehat{M}}}
\newcommand{\Mstar}{\ensuremath{M^*}}
\newcommand{\Treg}{\ensuremath{T_{\regparn}}}
\newcommand{\Tregp}{\ensuremath{T_{\regparnp}}}
\newcommand{\Hreg}{\ensuremath{H_{\regparn}}}
\newcommand{\Wmat}{\ensuremath{W}}

\newcommand{\regparn}{\ensuremath{\lambda_{\numitem}}}
\newcommand{\regparnp}{\ensuremath{\lambda_{\numitem}}}
\newcommand{\regparnsq}{\ensuremath{\lambda^2_\numitem}}

\newcommand{\real}{\ensuremath{\mathbb{R}}}

\newcommand{\Order}{\mathcal{O}}

 \newcommand{\pistar}{\pi^*}
\newcommand{\trace}{\operatorname{trace}}
 \newcommand{\LapErdos}{L}
\newcommand{\eigenvalue}[2]{\lambda_{#1}(#2)}
\newcommand{\singularvalue}[2]{\sigma_{#1}(#2)}

\newenvironment{carlist} {\begin{list}{$\bullet$}
    {\setlength{\topsep}{0in} \setlength{\partopsep}{0in}
      \setlength{\parsep}{0in} \setlength{\itemsep}{\parskip}
      \setlength{\leftmargin}{0.07in} \setlength{\rightmargin}{0.08in}
      \setlength{\listparindent}{0in} \setlength{\labelwidth}{0.08in}
      \setlength{\labelsep}{0.1in} \setlength{\itemindent}{0in}}}
               {\end{list}}

\newcommand{\bcar}{\begin{carlist}} \newcommand{\ecar}{\end{carlist}}

\newcommand{\KCON}{\ensuremath{c}}
\newcommand{\ULOW}{\ensuremath{\KCON_\ell}}
\newcommand{\UUP}{\ensuremath{\KCON_u}}
\newcommand{\UHP}{\ensuremath{\KCON}}
\newcommand{\UNUM}{\ensuremath{\KCON_0}}

\newcommand{\Mtil}{\ensuremath{\widetilde{M}}}

\newcommand{\myhalf}{\ensuremath{\frac{1}{2}}}

\newcommand{\qhat}{\ensuremath{\widehat{q}}}

\newcommand{\obsinter}{\ensuremath{\bar{\obs}}}
\newcommand{\obstil}{\ensuremath{\widetilde{\obs}}}
\newcommand{\permfas}{\ensuremath{\widehat{\pi}_{\mbox{\tiny{FAS}}}}}

\newcommand{\BRACON}{\ensuremath{c}}

\newcommand{\DelFAS}{\ensuremath{\DelHat_{\mbox{\tiny{FAS}}}}}

\newcommand{\boo}{\ensuremath{b}} 

\newcommand{\Ind}{\mathbf{1}}

\newcommand{\dham}{\ensuremath{d_{\mbox{\tiny{H}}}}}

\newcommand{\GVCARD}{\ensuremath{\bar{T}}}

\newcommand{\HACKUV}{1 \leq u < v \leq \numitems}

\newcommand{\permid}{\perm_{\mbox{\tiny id}}}

\newcommand\blfootnote[1]{%
  \begingroup
  \renewcommand\thefootnote{}\footnote{#1}%
  \addtocounter{footnote}{-1}%
  \endgroup
}

\begin{document}

\begin{center} {\LARGE{\bf{
Stochastically Transitive Models for Pairwise Comparisons: Statistical and Computational Issues}}}

\vspace*{.3in}

{\large{
\begin{tabular}{c}
Nihar B. Shah$^{\ast}$, Sivaraman Balakrishnan$^{\sharp}$, Adityanand Guntuboyina$^{\dagger}$\\ and Martin J. Wainwright$^{\dagger\ast}$
\end{tabular}

\vspace*{.1in}

\begin{tabular}{ccc}
$^{\dagger}$Department of Statistics & & $^{\ast}$Department of EECS
\end{tabular}

\begin{tabular}{c}
University of California, Berkeley \\ Berkeley, CA 94720\\
\end{tabular}

~\\
\begin{tabular}{c}
$^{\sharp}$Department of Statistics \\
Carnegie Mellon University,\\5000 Forbes Ave, Pittsburgh, PA 15213
\end{tabular}

\vspace*{.2in} }}

\vspace*{.2in}

\date{}

\blfootnote{Author email addresses: nihar@eecs.berkeley.edu, siva@stat.cmu.edu, aditya@stat.berkeley.edu, \mbox{wainwrig@berkeley.edu}.}

\begin{abstract}
There are various parametric models for analyzing pairwise comparison
data, including the Bradley-Terry-Luce (BTL) and Thurstone models, but
their reliance on strong parametric assumptions is limiting.  In this
work, we study a flexible model for pairwise comparisons, under
which the probabilities of outcomes are required only to satisfy a
natural form of stochastic transitivity. This class includes 
parametric models including the BTL and Thurstone models as special
cases, but is considerably more general. We provide various examples of models in this broader
stochastically transitive class for which classical parametric models provide poor
fits.  Despite this greater flexibility, we show that the matrix of
probabilities can be estimated at the same rate as in standard
parametric models.  On the other hand, unlike in the BTL and Thurstone
models, computing the minimax-optimal estimator in the stochastically transitive
model is non-trivial, and we explore various computationally tractable
alternatives.  We show that a simple singular value thresholding
algorithm is statistically consistent but does not achieve the minimax
rate. We then propose and study algorithms that achieve the minimax
rate over interesting sub-classes of the full stochastically transitive class.  We
complement our theoretical results with thorough numerical
simulations.
\end{abstract}
\end{center}


\section{Introduction}
\label{sec:intro}

Pairwise comparison data is ubiquitous and arises naturally in a
variety of applications, including tournament play, voting, online
search rankings, and advertisement placement problems.  In rough
terms, given a set of $\numitems$ objects along with a collection of
possibly inconsistent comparisons between pairs of these objects, the
goal is to aggregate these comparisons in order to estimate underlying
properties of the population.  One property of interest is the
underlying matrix of pairwise comparison probabilities---that is, the
matrix in which entry $(i,j)$ corresponds to the probability that
object $i$ is preferred to object $j$ in a pairwise comparison.  The
Bradley-Terry-Luce~\cite{bradley1952rank,luce1959individual} and
Thurstone~\cite{thurstone1927law} models are mainstays in analyzing
this type of pairwise comparison data.  These models are parametric in
nature: more specifically, they assume the existence of an
$n$-dimensional weight vector that measures the quality or strength of
each item.  The pairwise comparison probabilities are then determined
via some fixed function of the qualities of the pair of
objects. Estimation in these models reduces to estimating the
underlying weight vector, and a large body of prior work has focused
on these models (e.g., see the papers~\cite{ negahban2012iterative,
	hajek2014minimax, shah2015estimation}). However, such models enforce
strong relationships on the pairwise comparison probabilities that
often fail to hold in real applications.  Various
papers~\cite{davidson1959experimental,mclaughlin1965stochastic,tversky1972elimination,ballinger1997decisions}
have provided experimental results in which these parametric modeling
assumptions fail to hold.

Our focus in this paper is on models that have their roots in social
science and psychology (e.g., see Fishburn~\cite{fishburn1973binary}
for an overview), in which the only coherence assumption imposed on
the pairwise comparison probabilities is that of \emph{strong
	stochastic transitivity}, or SST for short. These models include the
parametric models as special cases but are considerably more
general. The SST model has been validated by several empirical
analyses, including those in a long line of
work~\cite{davidson1959experimental,mclaughlin1965stochastic,tversky1972elimination,ballinger1997decisions}.
The conclusion of Ballinger et al.~\cite{ballinger1997decisions} is
especially strongly worded:
\begin{quotation}
	{\textit{All of these parametric c.d.f.s are soundly rejected by our
			data. However, SST usually survives scrutiny.}}
\end{quotation}
We are thus provided with strong empirical motivation for studying the
fundamental properties of pairwise comparison probabilities satisfying
SST.

In this paper, we focus on the problem of estimating the matrix of
pairwise comparison probabilities---that is, the probability that an
item $i$ will beat a second item $j$ in any given comparison.
Estimates of these comparison probabilities are useful in various
applications.  For instance, when the items correspond to players or
teams in a sport, the predicted odds of one team beating the other are
central to betting and bookmaking operations.  In a supermarket or an
ad display, an accurate estimate of the probability of a customer
preferring one item over another, along with the respective profits
for each item, can effectively guide the choice of which product to
display.  Accurate estimates of the pairwise comparison probabilities
can also be used to infer partial or full rankings of the underlying
items.

\paragraph{Our contributions:} 
We begin by studying the performance
of optimal methods for estimating matrices in the SST class: our first
main result (Theorem~\ref{ThmMinimax}) characterizes the minimax rate
in squared Frobenius norm up to logarithmic factors.  This result
reveals that even though the SST class of matrices is considerably
larger than the classical parametric class, surprisingly, it is
possible to estimate any SST matrix at nearly the same rate as the
classical parametric family.  On the other hand, our achievability
result is based on an estimator involving prohibitive computation, as
a brute force approach entails an exhaustive search over permutations.
Accordingly, we turn to studying computationally tractable estimation
procedures.  Our second main result (Theorem~\ref{ThmImprovedUSVT})
applies to a polynomial-time estimator based on soft-thresholding the
singular values of the data matrix.  An estimator based on
hard-thresholding was studied previously in this context by
Chatterjee~\cite{chatterjee2014matrix}. We sharpen and generalize this
previous analysis, and give a tight characterization of the rate
achieved by both hard and soft-thresholding estimators.  Our third
contribution, formalized in Theorems~\ref{ThmHighSNR}
and~\ref{thm:parametric}, is to show how for certain interesting
subsets of the full SST class, a combination of parametric maximum
likelihood~\cite{shah2015estimation} and noisy sorting
algorithms~\cite{braverman2008noisy} leads to a tractable two-stage
method that achieves the minimax rate.  Our fourth contribution is to
supplement our minimax lower bound with lower bounds for various known
estimators, including those based on thresholding singular
values~\cite{chatterjee2014matrix}, noisy
sorting~\cite{braverman2008noisy}, as well as parametric
estimators~\cite{negahban2012iterative,hajek2014minimax,shah2015estimation}.
These lower bounds show that none of these tractable estimators
achieve the minimax rate uniformly over the entire class. The lower
bounds also show that the minimax rates for any of these subclasses is
no better than the full SST class.


\paragraph{Related work:} The literature on ranking and estimation
from pairwise comparisons is vast and we refer the reader to various
surveys~\cite{fligner1993probability,marden1996analyzing,cattelan2012models}
for a more detailed overview. Here we focus our literature review on
those papers that are most closely related to our contributions.  Some
recent
work~\cite{negahban2012iterative,hajek2014minimax,shah2015estimation}
studies procedures and minimax rates for estimating the latent quality
vector that underlie parametric models.  Theorem~\ref{thm:parametric}
in this paper provides an extension of these results, in particular by
showing that an optimal estimate of the latent quality vector can be
used to construct an optimal estimate of the pairwise comparison
probabilities.  Chatterjee~\cite{chatterjee2014matrix} analyzed matrix
estimation based on singular value thresholding, and obtained results
for the class of SST matrices.  In Theorem~\ref{ThmImprovedUSVT}, we
provide a sharper analysis of this estimator, and show that our upper
bound is---in fact---unimprovable.

In past work, various authors~\cite{kenyon2007rank,
	braverman2008noisy} have considered the noisy sorting problem, in
which the goal is to infer the underlying order under a so-called high
signal-to-noise ratio (SNR) condition.  The high SNR condition means
that each pairwise comparison has a probability of agreeing with the
underlying order that is bounded away from $\half$ by a fixed
constant. Under this high SNR condition, these authors provide
polynomial-time algorithms that, with high probability, return an
estimate of true underlying order with a prescribed accuracy.  Part of
our analysis leverages an algorithm from the
paper~\cite{braverman2008noisy}; in particular, we extend their
analysis in order to provide guarantees for estimating pairwise
comparison probabilities as opposed to estimating the underlying
order.

As will be clarified in the sequel, the assumption of strong
stochastic transitivity has close connections to the statistical
literature on shape constrained inference
(e.g.,~\cite{silvapulle2011constrained}), particularly to the problem
of bivariate isotonic regression.  In our analysis of the
least-squares estimator, we leverage metric entropy bounds from past
work in this area (e.g.,~\cite{gao2007entropy,
	chatterjee2015biisotonic}).

In Appendix~\ref{AppRelationsModels} of the present paper, we study
estimation under two popular models that are closely related to the
SST class, and make even weaker assumptions. We show that under these
moderate stochastic transitivity (MST) and weak stochastic
transitivity (WST) models, the Frobenius norm error of any estimator,
measured in a uniform sense over the class, must be almost as bad as
that incurred by making no assumptions whatsoever. Consequently, from
a statistical point of view, these assumptions are not strong enough
to yield reductions in estimation error.  We note that the ``low noise
model'' studied in the paper~\cite{rajkumar2014statistical} is
identical to the WST class.


\paragraph*{Organization:}
The remainder of the paper is organized as follows. We begin by providing a background and a formal description of the problem in Section~\ref{SecBackground}. In Section~\ref{SecMainResults}, we present the main theoretical results of the paper. We then present results from numerical simulations in Section~\ref{SecSimulations}. We present proofs of our main results in Section~\ref{SecProofs}. We conclude the paper in Section~\ref{SecDiscussion}.


\section{Background and problem formulation}
\label{SecBackground}

Given a collection of $\numitem \geq 2$ items, suppose that we have
access to noisy comparisons between any pair $i \neq j$ of distinct
items.  The full set of all possible pairwise comparisons can be
described by a probability matrix $\Mstar \in [0,1]^{\numitem \times
	\numitem}$, in which $\Mstar_{ij}$ is the probability that item $i$
is preferred to item $j$. The upper and lower halves of the
probability matrix $\Mstar$ are related by the
\emph{shifted-skew-symmetry condition}\footnote{In other words, the
	shifted matrix $\Mstar - \half$ is skew-symmetric.}
\mbox{$\Mstar_{ji} = 1 - \Mstar_{ij}$} for all \mbox{$i, j \in
	[\numitems]$.}  For concreteness, we set $\Mstar_{ii} = 1/2$ for all
$i \in [\numitems]$.

\subsection{Estimation of pairwise comparison probabilities}
For any matrix $\Mstar \in [0,1]^{\numitem \times \numitem}$ with
$\wtstar_{ij} = 1-\wtstar_{ji}$ for every $(i,j)$, suppose that we
observe a random matrix $\obs \in \{0,1 \}^{\numitem \times \numitem}$
with (upper-triangular) independent Bernoulli entries, in particular,
with $\mprob[\obs_{ij} = 1] = \Mstar_{ij}$ for every $1 \leq i \leq j
\leq \numitems$ and $\obs_{ji} = 1-\obs_{ij}$.  Based on observing
$Y$, our goal in this paper is to recover an accurate estimate, in the
squared Frobenius norm, of the full matrix $\Mstar$. 

Our primary focus in this paper will be on the setting where for $\numitem$ items we observe
the outcome of a single pairwise comparison for each pair.  We will subsequently (in Section~\ref{SecPartial}) also address the more general case when 
we have partial observations, that is, when each pairwise comparison is observed with a fixed
probability.

For future
reference, note that we can always write the Bernoulli observation
model in the linear form
\begin{align}
\label{EqnObservationModel}
\obs & = \Mstar + \noise,
\end{align}
where $\noise \in [-1,1]^{\numitem \times \numitem}$ is a random
matrix with independent zero-mean entries for every $i \geq j$ given by
\begin{align}
\label{EqnDefnWmat}
\noise_{ij} & \sim \begin{cases} 1 - \Mstar_{ij} & \mbox{with
    probability $\Mstar_{ij}$} \\
-\Mstar_{ij} & \mbox{with probability $1 - \Mstar_{ij}$,}
\end{cases}
\end{align}
and $\noise_{ji} = - \noise_{ij}$ for every $i < j$.  This
linearized form of the observation model is convenient for subsequent
analysis.


\subsection{Strong stochastic transitivity}
\label{sec:models_SST}
Beyond the previously mentioned constraints on the matrix
$\Mstar$---namely that $\Mstar_{ij} \in [0,1]$ and \mbox{$\Mstar_{ij}
	= 1 - \Mstar_{ij}$}---more structured and interesting models are
obtained by imposing further restrictions on the entries of $\Mstar$.
We now turn to one such condition, known as \emph{strong stochastic
	transitivity} (SST), which reflects the natural transitivity of any
complete ordering.  Formally, suppose that the full collection of
items $[\numitem]$ is endowed with a complete ordering $\pistar$.  We
use the notation $\pistar(i) < \pistar(j)$ to convey that item $i$ is
preferred to item $j$ in the total ordering $\pistar$.  Consider some
triple $(i,j,k)$ such that $\pistar(i) < \pistar(j)$.  A matrix
$\Mstar$ satisfies the SST condition if the inequality $\Mstar_{ik}
\geq \Mstar_{jk}$ holds for every such triple.  The intuition
underlying this constraint is the following: since $i$ dominates $j$
in the true underlying order, when we make noisy comparisons, the
probability that $i$ is preferred to $k$ should be at least as large
as the probability that $j$ is preferred to $k$.  The SST condition
was first described\footnote{We note that the psychology literature
	has also considered what are known as weak and moderate stochastic
	transitivity conditions.  From a statistical standpoint, pairwise
	comparison probabilities cannot be consistently estimated in a
	minimax sense under these conditions. We establish this formally in
	Appendix~\ref{AppRelationsModels}.}  in the psychology literature
(e.g.,~\cite{fishburn1973binary,davidson1959experimental}).

The SST condition is characterized by the existence of
a permutation such that the permuted matrix has entries that increase
across rows and decrease down columns.  More precisely, for a given
permutation $\pistar$, let us say that a matrix $\wt$ is
$\pistar$-faithful if for every pair $(i,j)$ such that $\pistar(i) <
\pistar(j)$, we have $M_{ik} \geq M_{jk}$ for all $k \in [\numitems]$.
With this notion, the set of SST matrices is given by
\begin{align}
\label{EqnDefnSST}
\chatterjeeclass = \big \{ M \in [0,1]^{\numitem \times \numitem} \,
\mid \, &\mbox{ $M_{ba} = 1 - M_{ab} \; \forall \; (a,b)$, and
  $\exists$ perm. $\pistar$ s.t. $M$ is $\pistar$-faithful} \big \}.
\end{align}
Note that the stated inequalities also guarantee that for any pair
with $\pistar(i) < \pistar(j)$, we have $M_{ki} \leq M_{kj}$ for
all $k$, which corresponds to a form of column ordering. The class
$\chatterjeeclass$ is our primary focus in this paper.


\subsection{Classical parametric models}
\label{sec:models_parametric}

Let us now describe a family of classical parametric models, one which
includes Bradley-Terry-Luce and Thurstone (Case V)
models~\cite{bradley1952rank,luce1959individual,thurstone1927law}.  In
the sequel, we show that these parametric models induce a relatively
small subclass of the SST matrices $\chatterjeeclass$.

In more detail, parametric models are described by a weight vector
$\wtparamstar \in \real^{\numitem}$ that corresponds to the notional
qualities of the $\numitem$ items.  Moreover, consider any
non-decreasing function $\glmcdf: \real \mapsto [0,1]$ such that
\mbox{$\glmcdf(t) = 1 - \glmcdf(-t)$} for all $t \in \real$; we refer
to any such function $\glmcdf$ as being \emph{valid}.  Any such pair
$(\glmcdf, \wtparamstar)$ induces a particular pairwise comparison
model in which
\begin{align}
\label{EqnInduce}
\Mstar_{ij} = \glmcdf(\wtparamstar_i - \wtparamstar_j) \qquad
\mbox{for all pairs $(i,j)$}.
\end{align}
For each valid choice of $\glmcdf$, we define
\begin{align}
\label{EqnDefnParamClass}
\paramclass(\glmcdf) = \Big \{ M \in [0,1]^{\numitem \times \numitem}
\, \mid \, & \mbox{$M$ induced by Equation~\eqref{EqnInduce} for some
  $\wtparamstar \in \real^\numitems$} \Big \}.
\end{align}
For any choice of $\glmcdf$, it is straightforward to verify that
$\paramclass(\glmcdf)$ is a subset of $\chatterjeeclass$, meaning that
any matrix $M$ induced by the relation~\eqref{EqnInduce} satisfies all
the constraints defining the set $\chatterjeeclass$.  As particular
important examples, we recover the Thurstone model by setting
$\glmcdf(t) = \Phi(t)$ where $\Phi$ is the Gaussian CDF, and the
Bradley-Terry-Luce model by setting $\glmcdf(t) = \frac{e^t}{1 +
	e^t}$, corresponding to the sigmoid function.

\paragraph{Remark:}  One can impose further constraints on the vector
$\wtparamstar$ without reducing the size of the class $\{
\paramclass(\glmcdf), \; \mbox{for some valid $\glmcdf$} \}$.  In
particular, since the pairwise probabilities depend only on the
differences $\wtparamstar_i - \wtparamstar_j$, we can assume without loss of
generality that $\inprod{\wtparamstar}{1} = 0$.  Moreover, since the
choice of $\glmcdf$ can include rescaling its argument, we can also
assume that $\|\wtparamstar\|_\infty \leq 1$.  Accordingly, we assume
in our subsequent analysis that $\wtparamstar$ belongs to the set
\begin{align*}
\big \{\wtparam \in \real^\numitems \, \mid \, \mbox{such
  that $\inprod{\wtparam}{1} = 0$ and $\|\wtparam\|_\infty \leq 1$.}
\big \}.
\end{align*}


\subsection{Inadequacies of parametric models}
\label{SecParametricBad}

As noted in the introduction, a large body of past work
(e.g.,~\cite{davidson1959experimental, mclaughlin1965stochastic,
  tversky1972elimination, ballinger1997decisions}) has shown that
parametric models, of the form~\eqref{EqnDefnParamClass} for some
choice of $\glmcdf$, often provide poor fits to real-world data.  What
might be a reason for this phenomenon? Roughly, parametric models
impose the very restrictive assumption that the choice of an item
depends on the value of a single latent factor (as indexed by
$\wtparamstar$)---e.g., that our preference for cars depends only on
their fuel economy, or that the probability that one hockey team beats
another depends only on the skills of the goalkeepers.

This intuition can be formalized to construct matrices $\Mstar \in
\chatterjeeclass$ that are far away from \emph{every valid parametric
  approximation} as summarized in the following result:
\begin{proposition}
	\label{PropParametricBreak}
	There exists a universal constant $\KCON > 0$ such that for every
	$\numitems \geq 4$, there exist matrices $\Mstar \in \chatterjeeclass$
	for which
	\begin{align}
	\label{EqnParametricBreak}
	\frac{1}{\numitems^2} \inf_{\mathrm{valid}~\glmcdf} \; \inf_{\wt \in
		\paramclass(\glmcdf)} \frobnorm{\wt - \wtstar}^2 \geq \KCON.
	\end{align}
\end{proposition}
\noindent Given that every entry of matrices in $\chatterjeeclass$ lies in the interval $[0,1]$, the Frobenius norm
diameter of the class $\chatterjeeclass$ is at most $\numitems^2$, so
that the scaling of the lower
bound~\eqref{EqnParametricBreak} cannot be sharpened. See Appendix~\ref{AppPropParametricBreak} for a proof of Proposition~\ref{PropParametricBreak}.

\begin{figure}[t]
\begin{center}
\begin{tabular}{ccc}
\raisebox{1in}{
\parbox{.38\textwidth}{
\begin{equation*}
\Mstar \defn \frac{1}{8}
\begin{bmatrix}
4 & 6 & 7 & 8\\ 2 & 4 & 7 & 8\\ 1 & 1 & 4 & 5\\ 0 & 0 & 3 & 4
\end{bmatrix} \; \in \real^{\numitems \times \numitems}
\end{equation*}
} 
} 
& &
\includegraphics[width =
  .58\textwidth]{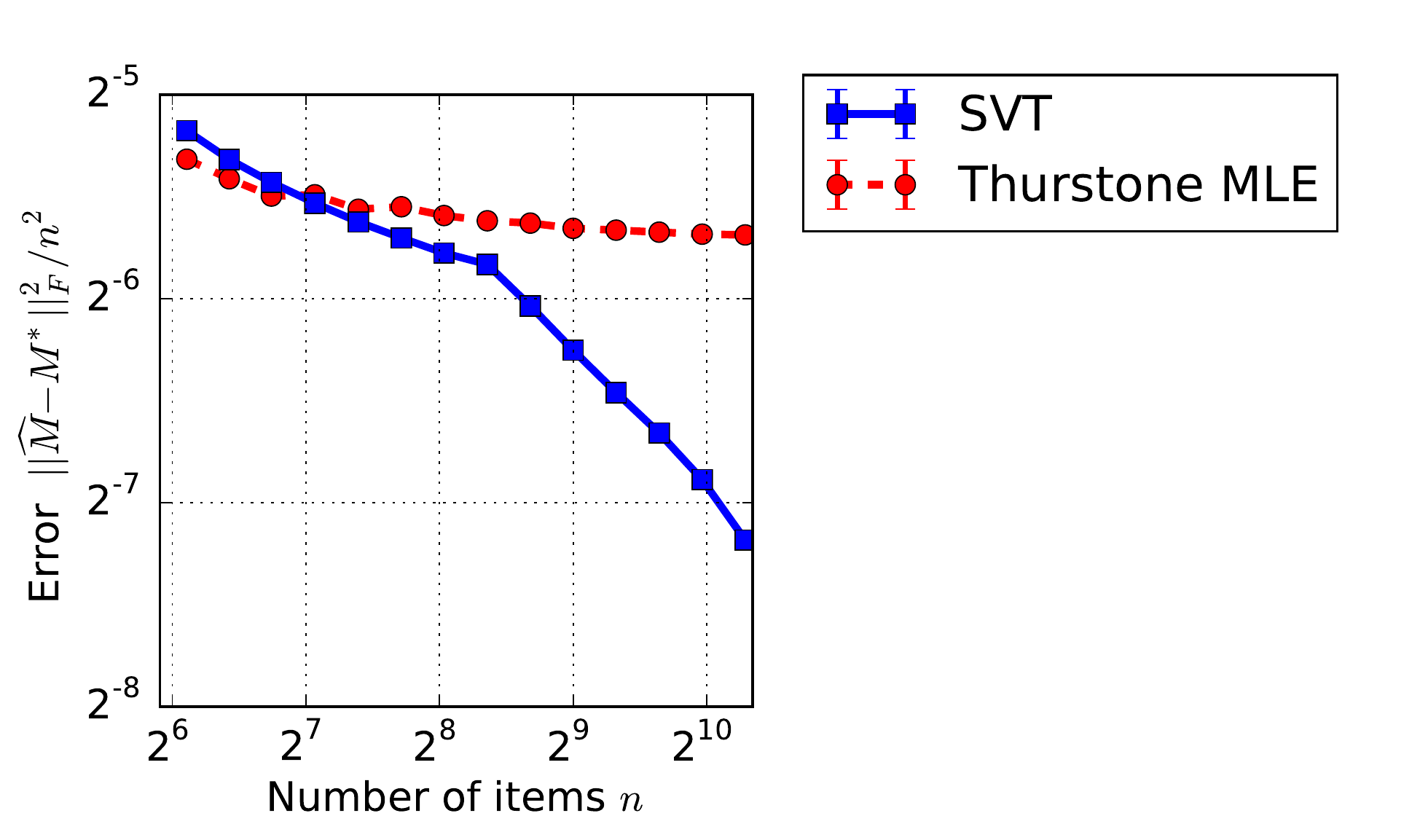}~~ \\
(a) & & (b)
\end{tabular}
\end{center}
\caption{ (a) Construction of a ``bad'' matrix: for $\numitems$
  divisible by $4$, form the matrix $\Mstar \in \real^{\numitems
    \times \numitems}$ shown, where each block has dimensions
  $\numitems/4 \times \numitems/4$. (b) Estimation for a class of SST
  matrices that are far from the parametric models. The parametric
  model (Thurstone MLE) yields a poor fit, whereas fitting using the
  singular value thresholding (SVT) estimator, which allows for
  estimates over the full SST class, leads to consistent estimation.}
\label{FigParametricBreak}
\end{figure}

What sort of matrices $\Mstar$ are ``bad'' in the sense of satisfying
a lower bound of the form~\eqref{EqnParametricBreak}?  Panel (a) of
Figure~\ref{FigParametricBreak} describes the construction of one
``bad'' matrix $\Mstar$.  In order to provide some intuition, let us return to
the analogy of rating cars. A key property of any parametric model 
is that if we prefer car 1 to car 2 more than we prefer car 3 to car 4,
then we must also prefer car 1 to car 3 more than we prefer car 2 to
car 4.\footnote{This condition follows from the proof of Proposition~\ref{PropParametricBreak}.}  This condition is potentially satisfied if there is a single
determining factor across all cars---for instance, their fuel economy.  

This ordering condition is, however, violated by the pairwise comparison matrix $\wtstar$
from Figure~\ref{FigParametricBreak}(a). In
this example, we have $\wtstar_{12} = \frac{6}{8} > \frac{5}{8} =
\wtstar_{34}$ and $\wtstar_{13} = \frac{7}{8} < \frac{8}{8} =
\wtstar_{24}$. 
Such an occurrence can be explained by a simple
two-factor model: suppose the fuel economies of cars $1,2,3$ and $4$ are $20$, $18$, $12$ and $6$ kilometers per liter 
respectively, and
the comfort levels of the four cars are also ordered $1 \succ 2
\succ 3 \succ 4$, with $i \succ j$ meaning that $i$ is more comfortable than $j$. Suppose that in a pairwise comparison of two cars,
if one car is more fuel efficient by at least 10 kilometers per liter, it is 
always chosen.
Otherwise the choice is governed by a parametric 
choice model acting on the respective comfort levels of the
two cars. 
Observe that while the comparisons between the pairs
$(1,2)$, $(3,4)$ and $(1,3)$ of cars can be explained by this
parametric model acting on their respective comfort levels, the
preference between cars $1$ and $4$,
as well as between cars $2$ and $4$, is governed by their fuel economies. This
two-factor model accounts for the said values of $\wtstar_{12}$,
$\wtstar_{34}$, $\wtstar_{24}$ and $\wtstar_{13}$, which violate
parametric requirements. 

While this was a simple hypothetical example, there is a more ubiquitous phenomenon
underlying our example. It is often the case that our preferences are decided by comparing items
on a multitude of dimensions. In any situation where a
single (latent) parameter per item does not adequately explain our preferences,
one can expect that the class of parametric models to provide a poor
fit to the pairwise preference probabilities.

The lower bound on approximation
quality guaranteed by Proposition~\ref{PropParametricBreak} means that
any parametric estimator of the matrix $\wtstar$ should perform
poorly.  This expectation is confirmed by the simulation results in
panel (b) of Figure~\ref{FigParametricBreak}.  After generating
observations from a ``bad matrix'' over a range of $\numitems$, we
fit the data set using either the maximum likelihood estimate in the
Thurstone model, or the singular value thresholding (SVT) estimator,
to be discussed in Section~\ref{sec:SVD}.  For each estimator $\Mhat$,
we plot the rescaled Frobenius norm error $\frac{\frobnorm{\Mhat -
    \Mstar}^2}{\numitems^2}$ versus the sample size.  Consistent with
the lower bound~\eqref{EqnParametricBreak}, the error in the
Thurstone-based estimator stays bounded above a universal constant.
In contrast, the SVT error goes to zero with $\numitems$, and as our
theory in the sequel shows, the rate at which the error 
decays is at least as fast as $1/\sqrt{\numitems}$.


\section{Main results}
\label{SecMainResults}

Thus far, we have introduced two classes of models for matrices of
pairwise comparison probabilities.  Our main results characterize the
rates of estimation associated with different subsets of these
classes, using either optimal estimators (that we suspect are not
polynomial-time computable), or more computationally efficient
estimators that can be computed in polynomial-time.

\subsection{Characterization of the minimax risk}
\label{sec:minimax_risk}

We begin by providing a result that characterizes the minimax risk in
squared Frobenius norm over the class $\chatterjeeclass$ of SST
matrices.  The minimax risk is defined by taking an infimum over the
set of all possible estimators, which are measurable functions $Y
\mapsto \Mtil \in [0,1]^{\numitem \times \numitem}$.  Here the data
matrix $Y \in \{0,1\}^{\numitem \times \numitem }$ is drawn from the
observation model~\eqref{EqnObservationModel}.

\begin{theorem}
\label{ThmMinimax}
There are universal constants $0 < \ULOW < \UUP$ such that
\begin{align}
\label{EqnMinimax}
\frac{\ULOW}{\numitems} \; \leq \; \inf_{\Mtil} \sup_{\Mstar \in
  \chatterjeeclass} \frac{1}{\numitems^2} \Exs[ \frobnorm{\Mtil -
    \Mstar}^2] \; \leq \; \UUP \frac{\log^2(\numitems)}{\numitems},
\end{align}
where the infimum ranges over all measurable functions $\Mtil$ of the
observed matrix $\obs$.
\end{theorem}
\noindent

We prove this theorem in Section~\ref{AppThmMinimax}.  The proof of
the lower bound is based on extracting a particular subset of the
class $\chatterjeeclass$ such that any matrix in this subset has at
least $\numitem$ positions that are unconstrained, apart from having
to belong to the interval $[\half, 1]$.  We can thus conclude that
estimation of the full matrix is at least as hard as estimating
$\numitem$ Bernoulli parameters belonging to the interval $[\half,
1]$ based on a single observation per number.  This reduction leads
to an $\Omega(\numitem^{-1})$ lower bound, as stated.

Proving an upper bound requires substantially more effort.  In
particular, we establish it via careful analysis of the constrained
least-squares estimator
\begin{subequations}
\begin{align}
\label{EqnLeastSquares}
\Mhat \in \argmin_{M \in \chatterjeeclass} \frobnorm{Y - M}^2.
\end{align}
In particular, we prove that there are universal constants $(c_0, c_1,
c_2)$ such that, for any matrix \mbox{$\Mstar \in \chatterjeeclass$,}
this estimator satisfies the high probability bound
\begin{align}
\label{EqnHighProbability}
\mprob \Big[ \frac{1}{\numitems^2} \frobnorm{\wthat - \Mstar}^2 \geq
  c_0 \frac{\log^2(\numitems)}{\numitems} \Big] & \leq c_1 e^{-c_2
  \numitems}.
\end{align}
\end{subequations}
Since the entries of $\wthat$ and $\Mstar$ all lie in the interval
$[0,1]$, integrating this tail bound leads to the stated upper
bound~\eqref{EqnMinimax} on the expected mean-squared error.  Proving
the high probability bound~\eqref{EqnHighProbability} requires sharp
control on a quantity known as the localized Gaussian complexity of
the class $\chatterjeeclass$.  We use Dudley's entropy integral
(e.g.,~\cite[Corollary 2.2.8]{van1996weak}) in order to derive an
upper bound that is sharp up to a logarithmic factor; doing so in turn
requires deriving upper bounds on the metric entropy of the class
$\chatterjeeclass$ for which we leverage the prior work of Gao and
Wellner~\cite{gao2007entropy}.

We do not know whether the constrained least-squares
estimator~\eqref{EqnLeastSquares} is computable in time polynomial in
$\numitems$, but we suspect not. This complexity is a consequence of
the fact that the set $\chatterjeeclass$ is not convex, but is a union
of $\numitems!$ convex sets.  Given this issue, it becomes interesting
to consider the performance of alternative estimators that can be
computed in polynomial-time.


\subsection{Sharp analysis of singular value thresholding (SVT)} 
\label{sec:SVD}

The first polynomial-time estimator that we consider is a simple
estimator based on thresholding singular values of the observed matrix $Y$,
and reconstructing its truncated singular value decomposition.
For the full
class $\chatterjeeclass$, Chatterjee~\cite{chatterjee2014matrix}
analyzed the performance of such an estimator and proved that the
squared Frobenius error decays as $\order(\numitems^{-\frac{1}{4}})$
uniformly over $\chatterjeeclass$.  In this section, we prove that its
error decays as $\order(\numitems^{-\half})$, again uniformly over
$\chatterjeeclass$, and moreover, that this upper bound is
unimprovable.

Let us begin by describing the estimator.  Given the observation
matrix $Y \in \real^{\numitem \times \numitem}$, we can write its
singular value decomposition as $\obs = U D V^T$, where the
$(\numitem \times \numitem)$ matrix $D$ is diagonal, whereas the
$(\numitem \times \numitem)$ matrices $U$ and $V$ are orthonormal.  Given a
threshold level $\regparn > 0$, the soft-thresholded version of a
diagonal matrix $D$ is the diagonal matrix $\Treg(D)$ with values
\begin{align}
\label{EqnDefnSoftSVT}
[\Treg(D)]_{jj} & = \max\{0, D_{jj} - \regparn\} \quad \mbox{for every
  integer $j \in [1,\numitems]$}.
\end{align}
With this
notation, the soft singular-value-thresholded (soft-SVT) version of
$\obs$ is given by $\Treg(\obs) = U \Treg(D) V^T$.  The following
theorem provides a bound on its squared Frobenius error:
\begin{theorem}
\label{ThmImprovedUSVT}
There are universal positive constants $(\UUP, \plaincon_0,
\plaincon_1)$ such that the soft-SVT estimator \mbox{$\wthatUSVT =
  \Treg(\obs)$} with $\regparn = 2.1 \sqrt{\numitem}$ satisfies the
bound
\begin{subequations}
\begin{align}
\label{EqnImprovedUSVT}
\mprob \Big[ \frac{1}{\numitems^2} \frobnorm{\wthatUSVT - \wtstar}^2
  \geq \frac{\UUP}{\sqrt{\numitems}} \Big] & \leq \plaincon_0 e^{-
  \plaincon_1 \numitems}
\end{align}
for any $\Mstar \in \chatterjeeclass$.  Moreover, there is a universal
constant $\ULOW > 0$ such that for \emph{any choice of $\regparn$}, we
have
\begin{align}
\label{EqnImprovedUSVTLower}
\sup_{\wtstar \in \chatterjeeclass} \frac{1}{\numitems^2}
\frobnorm{\wthatUSVT - \wtstar}^2 \geq \frac{\ULOW}{
  \sqrt{\numitems}}.
\end{align}
\end{subequations}
\end{theorem}

A few comments on this result are in order.  Since the matrices
$\wthatUSVT$ and $\wtstar$ have entries in the unit interval $[0,1]$,
the normalized squared error $\frac{1}{\numitems^2}
\frobnorm{\wthatUSVT - \wtstar}^2$ is at most $1$.  Consequently, by
integrating the the tail bound~\eqref{EqnImprovedUSVT}, we find that
\begin{align*}
\sup_{\Mstar \in \chatterjeeclass} \Exs[ \frac{1}{\numitems^2}
  \frobnorm{\wthatUSVT - \wtstar}^2] & \leq
\frac{\UUP}{\sqrt{\numitems}} + \plaincon_0 e^{- \plaincon_1
  \numitems} \; \leq \frac{\UUP'}{\sqrt{\numitems}}.
\end{align*}
On the other hand, the matching lower
bound~\eqref{EqnImprovedUSVTLower} holds with probability one, meaning
that the soft-SVT estimator has squared error of the order
$1/\sqrt{\numitems}$, irrespective of the realization of the noise.

To be clear, Chatterjee~\cite{chatterjee2014matrix} actually analyzed
the hard-SVT estimator, which is based on the hard-thresholding
operator
\begin{align*}
[\Hreg(D)]_{jj} & = D_{jj} \; \indicator{D_{jj} \geq \regparn}.
\end{align*}
Here $\indicator{\cdot}$ denotes the 0-1-valued indicator
function.
In this setting, the hard-SVT estimator is simply, $\Hreg(Y) = U \Hreg(D) V^T$.
With essentially the same choice of $\regparn$ as above, Chatterjee
showed that the estimate $\Hreg(Y)$ has a mean-squared error of 
$\Order(\numitems^{-1/4})$. One can verify that the proof of
Theorem~\ref{ThmImprovedUSVT} in our paper goes through for the
hard-SVT estimator as well. Consequently the performance of
the hard-SVT estimator is of the order
$\Theta(\numitems^{-1/2})$, and is identical to that of the
soft-thresholded version up to universal constants.

Note that the hard and soft-SVT estimators return matrices that may not lie in the SST class $\chatterjeeclass$. In a companion paper~\cite{shah2016feeling}, we provide an alternative computationally-efficient estimator with similar statistical guarantees that is guaranteed to return a matrix in the SST class.

Together the upper and lower bounds of Theorem~\ref{ThmImprovedUSVT}
provide a sharp characterization of the behavior of the soft/hard SVT
estimators. On the positive side, these are easily implementable
estimators that achieve a mean-squared error bounded by
$\Order(1/\sqrt{\numobs})$ uniformly over the entire class $\chatterjeeclass$.
On the negative side, this rate is slower than the $\Order(\log^2 \numobs/\numobs)$ rate
achieved by the least-squares estimator, as in Theorem~\ref{ThmMinimax}.

In conjunction, Theorems~\ref{ThmMinimax} and~\ref{ThmImprovedUSVT}
raise a natural open question: is there a polynomial-time estimator
that achieves the minimax rate uniformly over the family
$\chatterjeeclass$?  We do not know the answer to this
question, but the following subsections provide some partial answers
by analyzing some polynomial-time estimators that (up to logarithmic
factors) achieve the optimal $\tilde{\Order}(1/\numobs)$-rate over some interesting
sub-classes of $\chatterjeeclass$.  In the next two sections, we turn
to results of this type.


\subsection{Optimal rates for high SNR subclass}
\label{SecHighSNR}

In this section, we describe a multi-step polynomial-time estimator
that (up to logarithmic factors) can achieve the optimal $\tilde{\Order}(1/\numobs)$
rate over an interesting subclass of $\chatterjeeclass$. This subset
corresponds to matrices $M$ that have a relatively high
signal-to-noise ratio (SNR), meaning that no entries of $M$ fall
within a certain window of $1/2$.  More formally, for some $\bias \in
(0,\half)$, we define the class
\begin{align}
\label{EqnDefnBoundedAwayClass}
\chatterjeeclassbias(\bias) =  \big\{ \wt \in \chatterjeeclass \, \mid \,
\max (\wt_{ij}, \wt_{ji}) \geq 1/2 + \bias~~\forall~~i \neq j 
\big\}.
\end{align}
By construction, for any matrix $\chatterjeeclassbias(\bias)$, the amount of information
contained in each observation is bounded away from zero uniformly in
$\numitems$, as opposed to matrices in which some large subset of
entries have values equal (or arbitrarily close) to $\half$.  In terms
of estimation difficulty, this SNR restriction does not make the
problem substantially easier: as the following theorem shows, the
minimax mean-squared error remains lower bounded by a constant
multiple of $1/\numitems$.  Moreover, we can demonstrate a
polynomial-time algorithm that achieves this optimal mean squared error up to
logarithmic factors.

The following theorem applies to any fixed $\bias \in (0, \half]$
  independent of $\numitem$, and involves constants $(\ULOW, \UUP,
  \plaincon)$ that may depend on $\bias$ but are independent of
  $\numobs$.
\begin{theorem}
\label{ThmHighSNR}
There is a constant $\ULOW > 0$ such that
\begin{subequations}
\begin{align}
\label{EqnTwoStageLower}
\inf_{\Mtil} \sup_{\Mstar \in \chatterjeeclassbias(\bias)}
\frac{1}{\numitems^2} \Exs \big[ \frobnorm{\Mtil - \Mstar}^2] & \geq
\frac{\ULOW}{\numitems},
\end{align}
where the infimum ranges over all estimators.  Moreover, there is a
two-stage estimator $\Mhat$, computable in polynomial-time, for which
\begin{align}
\label{EqnTwoStageUpper}
\mprob \Big[ \frac{1}{\numitems^2} \frobnorm{\Mhat - \Mstar}^2 \geq
  \frac{\UUP \log^2(\numitem)}{\numitem} \Big] & \leq
\frac{\plaincon}{\numitems^2},
\end{align}
valid for any $\Mstar \in \chatterjeeclassbias(\bias)$.
\end{subequations}
\end{theorem}
\noindent As before, since the ratio $\frac{1}{\numitems^2}
\frobnorm{\Mhat - \Mstar}^2$ is at most $1$, so the tail
bound~\eqref{EqnTwoStageUpper} implies that
\begin{align}
\sup_{\Mstar \in \chatterjeeclassbias(\bias)}   \frac{1}{\numitem^2} \Exs [
  \frobnorm{\Mhat - \Mstar}^2] & \leq
\frac{\UUP \log^2(\numitem)}{\numitem} + \frac{\plaincon}{\numitems^2}
\; \leq \; \frac{\UUP' \log^2(\numitem)}{\numitem}.
\end{align}
We provide the proof of this theorem in Section~\ref{AppThmHighSNR}. As
with our proof of the lower bound in Theorem~\ref{ThmMinimax}, we
prove the lower bound by considering the sub-class of matrices that
are free only on the two diagonals just above and below the main
diagonal.  We now provide a brief sketch for the proof
of the upper bound~\eqref{EqnTwoStageUpper}. It is
based on analyzing the following two-step procedure:

\begin{enumerate}[leftmargin=*]
\item 
In the first step of algorithm, we find a permutation $\permfas$ of
the $\numitems$ items that minimizes the total number of disagreements
with the observations.  (For a given ordering, we say that any pair of
items $(i,j)$ are in disagreement with the observation if either $i$
is rated higher than $j$ in the ordering and $\obs_{ij}=0$, or if $i$
is rated lower than $j$ in the ordering and $\obs_{ij}=1$.)  The
problem of finding such a disagreement-minimizing permutation
$\permfas$ is commonly known as the minimum feedback arc set (FAS)
problem. It is known to be NP-hard in the worst-case~\cite{ailon2008aggregating,alon2006ranking}, but our set-up has
additional probabilistic structure that allows for polynomial-time
solutions with high probability. In
particular, we call upon a polynomial-time algorithm due to Braverman
and Mossel~\cite{braverman2008noisy} that, under the
model~\eqref{EqnDefnBoundedAwayClass}, is guaranteed to find the exact
solution to the FAS problem with high probability.  Viewing the FAS
permutation $\permfas$ as an approximation to the true permutation
$\pistar$, the novel technical work in this first step is show that
$\permfas$ is ``good enough'' for Frobenius norm estimation, in the
sense that for any matrix $\Mstar \in \chatterjeeclassbias(\bias)$, it satisfies
the bound
\begin{subequations}
\begin{align}
\frac{1}{\numitem^2} \frobnorm{\pistar(\Mstar) - \permfas(\Mstar)}^2 &
\leq \frac{\KCON \log \numitems}{\numitems}
\end{align}
with high probability. 
In this statement, for any given permutation
$\pi$, we have used $\pi(\wtstar)$ to denote the matrix obtained by
permuting the rows and columns of $\wtstar$ by $\pi$.
The term $\frac{1}{\numitem^2} \frobnorm{\pistar(\Mstar) - \permfas(\Mstar)}^2$
can be viewed in some sense as the \emph{bias}
in estimation incurred from using $\permfas$ in place of $\pistar$.
\item 
Next we define $\bisoclass$ as the class of ``bivariate isotonic'' matrices, that is, matrices $\wt \in [0,1]^{\numitem
  \times \numitem}$ that satisfy the linear constraints $M_{ij} = 1-
M_{ji}$ for all \mbox{$(i,j) \in [\numitems]^2$,} and $M_{k \ell} \geq
M_{ij}$ whenever $k \leq i$ and $\ell \geq j$.  This class corresponds
to the subset of matrices $\chatterjeeclass$ that are faithful with
respect to the identity permutation.  Letting
\mbox{$\permfas(\bisoclass) = \{ \permfas(M), \, M \in \bisoclass \}$}
denote the image of this set under $\permfas$, the second step
involves computing the constrained least-squares estimate
\begin{align}
\label{eq:bounded_away_algo_step2}
\wthat \in \argmin_{\wt \in \permfas(\bisoclass)} \frobnorm{\obs -
  \wt}^2.
\end{align}
\end{subequations}
Since the set $\permfas(\bisoclass)$ is a convex polytope, with a
number of facets that grows polynomially in $\numitems$, the
constrained quadratic program~\eqref{eq:bounded_away_algo_step2} can
be solved in polynomial-time.  The final step in the proof of
Theorem~\ref{ThmHighSNR} is to show that the estimator $\wthat$ also
has mean-squared error that is upper bounded by a constant multiple of
$\frac{\log^2(\numitems)}{\numitems}$.
\end{enumerate}

Our analysis shows that for any fixed $\bias \in (0, \frac{1}{2}]$,
  the proposed two-step estimator works well for any matrix $\Mstar
  \in \chatterjeeclassbias(\bias)$. Since this two-step estimator is based
  on finding a minimum feedback arc set (FAS) in the first step, it is
  natural to wonder whether an FAS-based estimator works well over the
  full class $\chatterjeeclass$.  Somewhat surprisingly 
  the answer to this question turns
  out to be negative: we refer the reader to
  Appendix~\ref{SecFASoverAllSST} for more intuition and details on
  why the minimal FAS estimator does not perform well over the full
  class.


\subsection{Optimal rates for parametric subclasses}

Let us now return to the class of parametric models
$\paramclass(\glmcdf)$ introduced earlier in
Section~\ref{sec:models_parametric}. As shown previously in
Proposition~\ref{PropParametricBreak}, this class is much smaller than
the class $\chatterjeeclass$, in the sense that there are models in
$\chatterjeeclass$ that cannot be well-approximated by any parametric
model.  Nonetheless, in terms of minimax rates of estimation, these
classes differ only by logarithmic factors.  An advantage of the
parametric class is that it is possible to achieve the $1/\numitem$
minimax rate by using a simple, polynomial-time estimator.  In
particular, for any log concave function $\glmcdf$, the maximum likelihood
estimate $\wtparamhatML$ can be obtained by solving a convex program.
This MLE induces a matrix estimate $\wt(\wtparamhatML)$
via Equation~\eqref{EqnInduce}, and the following result shows that
this estimator is minimax-optimal up to constant factors.

\begin{theorem}
\label{thm:parametric} 
Suppose that $\glmcdf$ is strictly increasing, strongly log-concave
and twice differentiable.  Then there is a constant $\ULOW >
0$, depending only on $\glmcdf$, such that the minimax risk over $\paramclass(\glmcdf)$ is lower
bounded as
\begin{subequations}
\begin{align}
\label{EqnParametricLower}
\inf_{\Mtil} \sup_{\Mstar \in \paramclass(\glmcdf) }
\frac{1}{\numitems^2} \Exs[ \frobnorm{\Mtil - \Mstar}^2 ] \geq
\frac{\ULOW}{\numitems},
\end{align}
Conversely, there is a constant $\UUP \geq \ULOW$, depending only on $\glmcdf$,
such that
the matrix estimate $\wt(\wtparamhatML)$ induced by the MLE satisfies
the bound
\begin{align}
\label{EqnParametricUpper}
\sup_{\wtstar \in \paramclass(\glmcdf)} \frac{1}{\numitems^2}
\Exs[\frobnorm{\wt(\wtparamhatML) - \wtstar}^2] \leq
\frac{\UUP}{\numitems}.
\end{align}
\end{subequations}
\end{theorem}
\noindent 
To be clear, the constants $(\ULOW, \UUP)$ in this theorem are
independent of $\numitems$, but they do
depend on the specific properties of the given function $\glmcdf$.
We note that the stated conditions on $\glmcdf$ are true for many popular
parametric models, including (for instance) the Thurstone and BTL
models.

We provide the proof of Theorem~\ref{thm:parametric} in
Section~\ref{AppThmParametric}.  The lower
bound~\eqref{EqnParametricLower} is, in fact, stronger than the the
lower bound in Theorem~\ref{ThmMinimax}, since the supremum is taken
over a smaller class.  The proof of the lower bound in
Theorem~\ref{ThmMinimax} relies on matrices that cannot be realized by
any parametric model, so that we pursue a different route to
establish the bound~\eqref{EqnParametricLower}.  On the other hand, in
order to prove the upper bound~\eqref{EqnParametricUpper}, we make use
of bounds on the accuracy of the MLE $\wtparamhatML$ from our own past
work (see the paper~\cite{shah2015estimation}).


\subsection{Extension to partial observations}
\label{SecPartial}

We now consider the extension of our results to the setting in which
not all entries of $\obs$ are observed. Suppose instead that every
entry of $\obs$ is observed independently with probability $\pp$. In
other words, the set of pairs compared is the set of edges of an
Erd\H{o}s-R\'enyi graph $\mathcal{G}(\numitems, \pp)$ that has the
$\numitems$ items as its vertices. 

In this setting, we obtain an upper bound on the minimax risk of
estimation by first setting $\obs_{ij} = \half$ whenever the pair
$(i,j)$ is not compared, then forming a new  $(\numitems \times
\numitems)$ matrix $\obs'$ as
\begin{subequations}
\begin{align}
\label{EqnDefnObsPartial}
\obs' \defn \frac{1}{\pp} \obs - \frac{1 - \pp}{2\pp} \ones \ones^T,
\end{align}
and finally computing the least squares solution
\begin{align}
\label{EqnDefnLSEPartial}
\wthat \in \argmin \limits_{\wt \in \chatterjeeclass} \frobnorm{ \obs' - \wt}^2.
\end{align}
\end{subequations}
Likewise, the computationally-efficient singular value thresholding
estimator is also obtained by thresholding the singular values of
$\obs'$ with a threshold $\regparnp = 3
  \sqrt{\frac{\numitems}{\pp}}$.  See our discussion following
Theorem~\ref{ThmPartialObservations} for the motivation underlying the
transformed matrix $\obs'$.

The parametric estimators continue to operate on the original
(partial) observations, first computing a maximum likelihood estimate
$\wtparamhatML$ of $\wtstar$ using the observed data, and then
computing the associated pairwise-comparison-probability matrix
$\wt(\wtparamhatML)$ via~\eqref{EqnInduce}.

\begin{theorem}
In the setting where each pair is observed with a probability $\pp$, there are positive universal constants $\ULOW$, $\UUP$ and $\UNUM$
  such that:
\label{ThmPartialObservations}
\begin{enumerate}
\item[(a)] The minimax risk is sandwiched as
\begin{subequations}
\begin{align}
\label{EqnPartialMinimax}
\frac{\ULOW}{\pp \numitems} \leq \inf_{\Mtil} \sup_{\wtstar \in \chatterjeeclass}
\frac{1}{\numitems^2} \Exs [ \frobnorm{\Mtil - \wtstar}^2 ]\leq \frac{\UUP (\log \numitems)^2}{\pp \numitems},
\end{align}
when $\pp \geq \frac{\UNUM}{\numitems}$.
	\item[(b)] The soft-SVT estimator, $\wthatUSVTp$ with $\regparnp =
	  3 \sqrt{\frac{\numitems}{\pp}}$, satisfies the bound
	\begin{align}
	\label{EqnPartialSVT}
	 \sup_{\wtstar \in \chatterjeeclass} \frac{1}{\numitems^2} \Exs [
	   \frobnorm{\wthatUSVTp - \wtstar}^2 ] & \leq \frac{\UUP}{
	   \sqrt{\numitems \pp}}.
	\end{align}
\item[(c)] For a parametric sub-class based on a strongly log-concave
  and smooth $\glmcdf$, the estimator $\wt(\wtparamhatML)$ induced by
  the maximum likelihood estimate $\wtparamhatML$ of the parameter
  vector $\wtparamstar$ has mean-squared error upper bounded as
\begin{align}
\label{EqnPartialParametric}
 \sup_{\wtstar \in \paramclass(\glmcdf)} \frac{1}{\numitems^2} \Exs[
   \frobnorm{\wt(\wtparamhatML) - \wtstar}^2 ] \leq \frac{\UUP}{\pp \numitems},
\end{align}
 when $\pp \geq  \frac{\UNUM (\log
     \numitems)^2}{\numitems}$.
\end{subequations}
\end{enumerate}
\end{theorem}

The intuition behind the transformation~\eqref{EqnDefnObsPartial} is
that the matrix $\obs'$ can equivalently be written in a linearized
form as
\begin{subequations}
\begin{align}
\label{EqnDefnObsPrime}
\obs' = \wtstar + \frac{1}{\pp} W',
\end{align}
where $W'$ has entries that are independent on and above the diagonal,
satisfy skew-symmetry, and are distributed as
\begin{align}
[W']_{ij} = 
\begin{cases} \pp(
\half - [\wtstar]_{ij}) + \frac{1}{2} & \qquad \mbox{with probability
} \pp [\wtstar]_{ij}\\ 
\pp (\half - [\wtstar]_{ij}) - \frac{1}{2} & \qquad \mbox{with
  probability } \pp (1-[\wtstar]_{ij})\\ 
\pp (\half - [\wtstar]_{ij}) & \qquad \mbox{with probability } 1-\pp .
\end{cases}
\label{EqnDefnWprimePartial}
\end{align}
\end{subequations}
The proofs of the upper bounds exploit the specific
relation~\eqref{EqnDefnObsPrime} between the observations $\obs'$ and
the true matrix $\wtstar$, and the specific form of the additive
noise~\eqref{EqnDefnWprimePartial}.

The result of Theorem~\ref{ThmPartialObservations}(b) yields an affirmative answer to the question, originally posed by Chatterjee~\cite{chatterjee2014matrix}, of whether or not the singular value thresholding estimator can yield a vanishing error when $\pp \leq \frac{1}{\sqrt{\numitems}}$.

We note that we do not have an analogue of the high-SNR result in the
partial observations case since having partial observations reduces
the SNR. In general, we are interested in scalings of $\pp$ which
allow $\pp \rightarrow 0$ as $n \rightarrow \infty$. The noisy-sorting
algorithm of Braverman and Mossel~\cite{braverman2008noisy} for the
high-SNR case has computational complexity scaling as
$e^{\bias^{-4}}$, and hence is not computable in time polynomial in
$\numitems$ when $\bias < (\log \numitems)^{-\frac{1}{4}}$.  This
restriction disallows most interesting scalings of $\pp$ with
$\numitems$.


\section{Simulations}
\label{SecSimulations}

In this section, we present results from simulations to gain a further
understanding of the problem at hand, in particular to understand the
rates of estimation under specific generative models.  We investigate
the performance of the soft-SVT estimator (Section~\ref{sec:SVD}) and the maximum likelihood
estimator under the Thurstone model
(Section~\ref{sec:models_parametric}).\footnote{We could not compare
  the algorithm that underlies Theorem~\ref{ThmHighSNR}, since it is
  not easily implementable.  In particular, it relies on the algorithm
  due to Braverman and Mossel~\cite{braverman2008noisy} to compute the
  feedback arc set minimizer.  The computational complexity of this
  algorithm, though polynomial in $\numitems$, has a large
  polynomial degree which precludes it from being implemented for
  matrices
  of any reasonable size.  
  
  The simulations in this section add to the
  simulation results of Section~\ref{SecParametricBad}
  (Figure~\ref{FigParametricBreak}) demonstrating a large class of
  matrices in the SST class that cannot be represented by any
  parametric class.} The output of the SVT estimator need not lie in
the set $[0,1]^{\numitems \times \numitems}$ of matrices; in our
implementation, we take a projection of the output of the SVT
estimator on this set, which gives a constant factor reduction in the
error. \\

\noindent In our simulations, we generate the ground truth $\wtstar$
in the following five ways:
\begin{itemize}[leftmargin=*]
  \setlength\itemsep{1em}
\item \underline{Uniform:} The matrix $\Mstar$ is generated by drawing
  ${\numitems \choose 2}$ values independently and uniformly at random
  in $[\half,1]$ and sorting them in descending order. The values are
  then inserted above the diagonal of an $(\numitems \times
  \numitems)$ matrix such that the entries decrease down a column or
  left along a row. We then make the matrix skew-symmetric and permute
  the rows and columns.
\item \underline{Thurstone:} The matrix $\wtstar \in [-1,1]^\numitems$ is generated
  by first choosing $\wtparamstar$ uniformly at random from the set
  satisfying $\inprod{\wtparamstar}{1} = 0$.  The matrix $\Mstar$ is then
  generated from $\wtparamstar$ via Equation~\eqref{EqnInduce} with $\glmcdf$ chosen as the CDF of
  the standard normal distribution.
\item \underline{Bradley-Terry-Luce (BTL):} Identical to the Thurstone
  case, except that $\glmcdf$ is given by the sigmoid function.
\item \underline{High SNR:} A setting studied previously by Braverman
  and Mossel~\cite{braverman2008noisy}, in which the noise is
  independent of the items being compared.  Some global order is fixed
  over the $\numitems$ items, and the comparison matrix $\wtstar$
  takes the values $\wtstar_{ij} = 0.9 = 1-\wtstar_{ji}$ for every
  pair $(i,j)$ where $i$ is ranked above $j$ in the underlying
  ordering. The entries on the diagonal are $0.5$.
\item \underline{Independent bands:} The matrix $\Mstar$ is chosen
  with diagonal entries all equal to $\frac{1}{2}$.  Entries on
  diagonal band immediately above the diagonal itself are chosen
  i.i.d. and uniformly at random from $[\half,1]$. The
  band above is then chosen uniformly at random from the allowable
  set, and so on. The choice of any entry in this process is only
  constrained to be upper bounded by $1$ and lower bounded by the
  entries to its left and below. The entries below the diagonal are
  chosen to make the matrix skew-symmetric.

\end{itemize}

\begin{figure}[t]
\centering
\begin{subfigure}{.24 \textwidth}
\includegraphics[width=\textwidth]{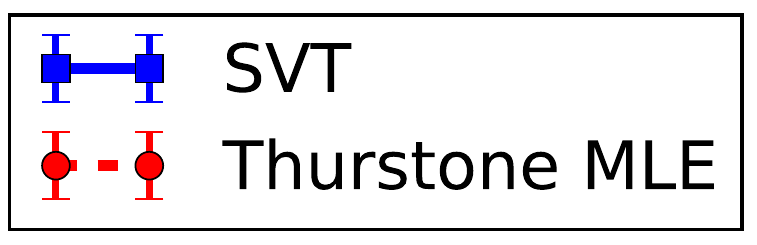}\\~\\~\\
\end{subfigure}
\begin{subfigure}{.32 \textwidth}
\includegraphics[width=\textwidth]{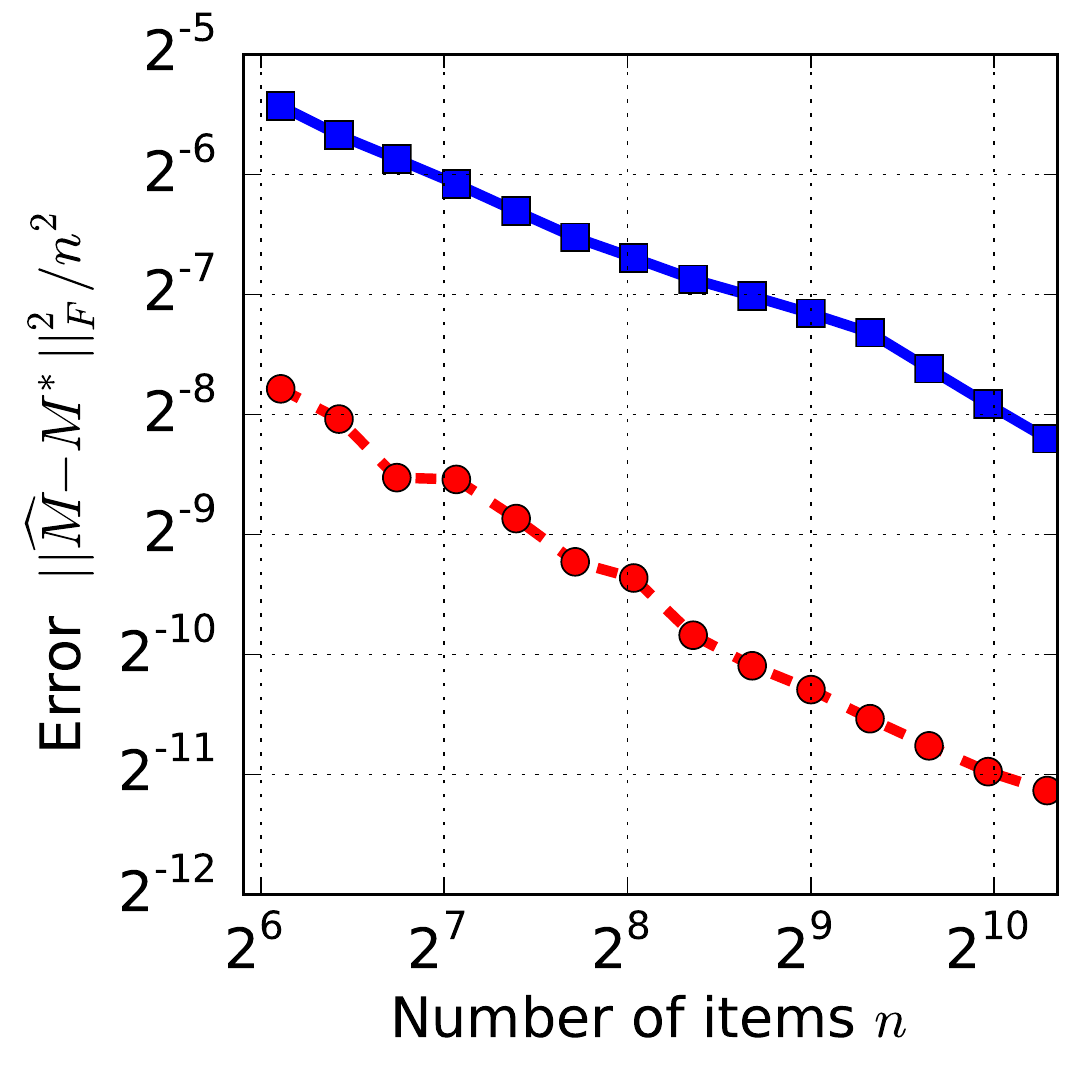}
\caption{Uniform}
\label{fig:simulations_uniform}
\end{subfigure}
\begin{subfigure}{.32 \textwidth}
\includegraphics[width=\textwidth]{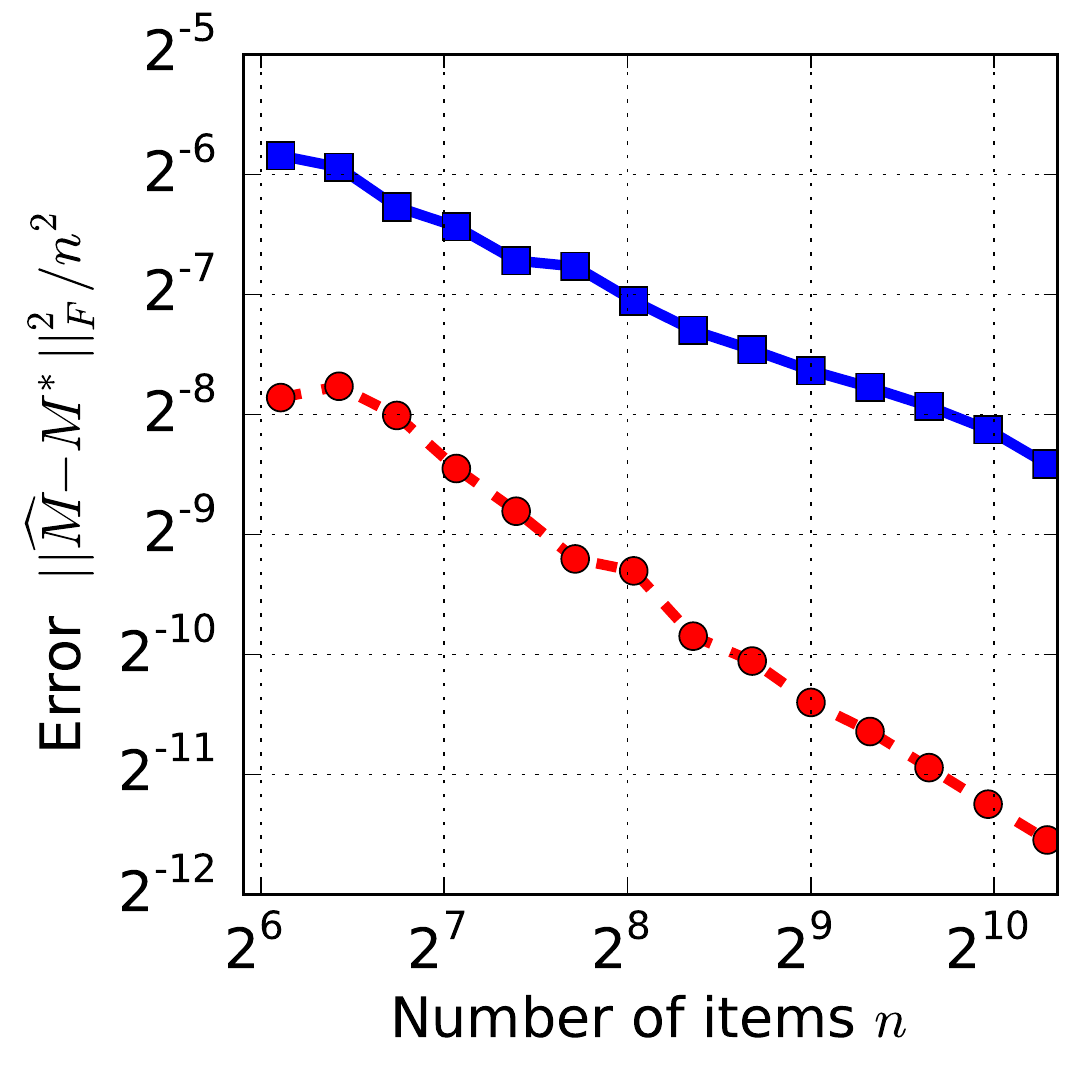}
\caption{Thurstone}
\label{fig:simulations_parametric_thurstone}
\end{subfigure}
\begin{subfigure}{.32 \textwidth}
\includegraphics[width=\textwidth]{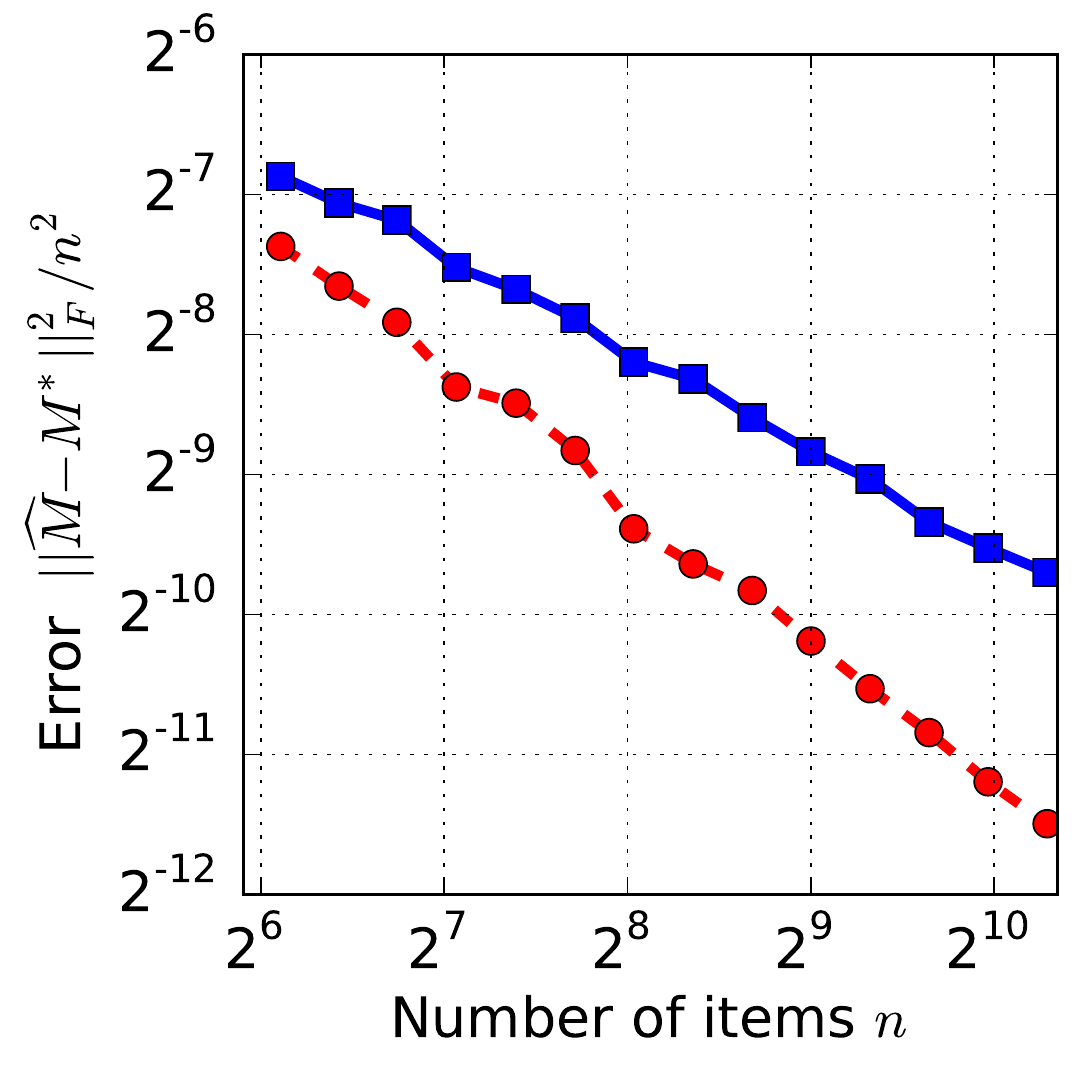}
\caption{BTL}
\label{fig:simulations_parametric_BTL_uniform}
\end{subfigure}
\begin{subfigure}{.32 \textwidth}
\includegraphics[width=\textwidth]{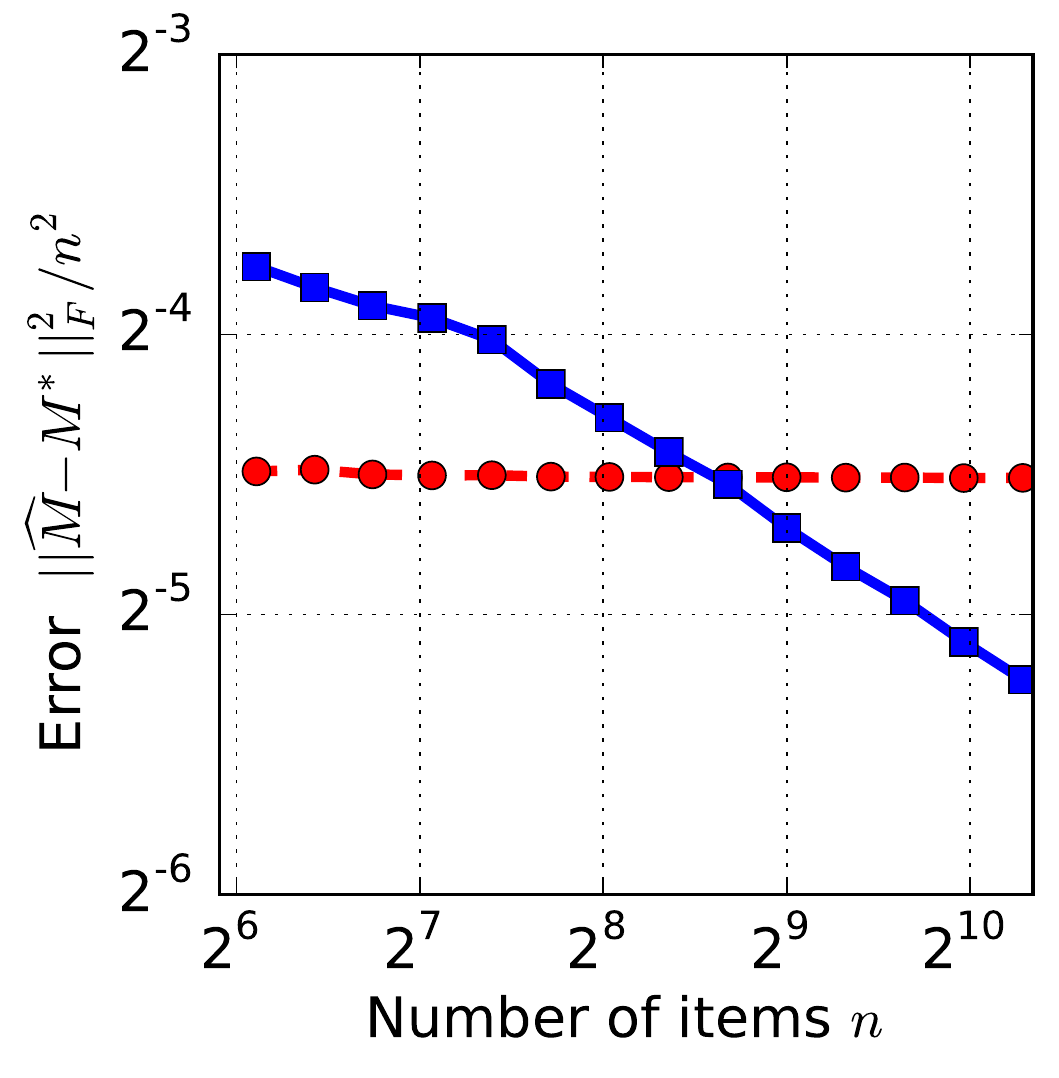}
\caption{High SNR}
\label{fig:simulations_highSNR}
\end{subfigure}
\begin{subfigure}{.32 \textwidth}
\includegraphics[width=\textwidth]{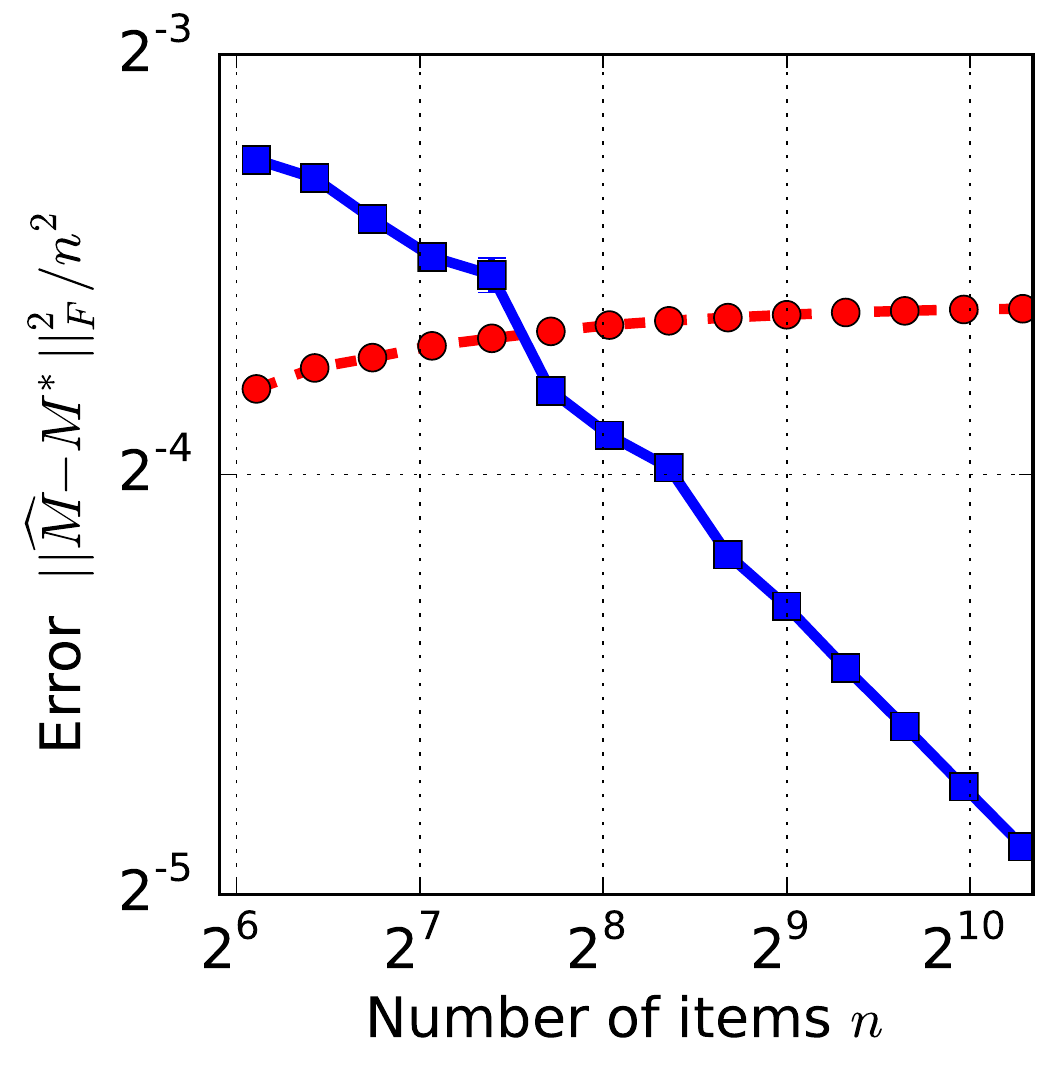}
\caption{Independent bands}
\label{fig:simulations_independent-diagonals}
\end{subfigure}
\caption{Errors of singular value thresholding (SVT) estimator and the
  Thurstone MLE under different methods to generate $\wtstar$.}
\label{fig:simulations}
\end{figure}

Figure~\ref{fig:simulations} depicts the results of the simulations
based on observations of the entire \mbox{matrix $\obs$.} Each point
is an average across $20$ trials. The error bars in most cases are too
small and hence not visible. We see that the uniform case
(Figure~\ref{fig:simulations_uniform}) is favorable for both
estimators, with the error scaling as
$\order(\frac{1}{\sqrt{\numitems}})$. With data generated from the
Thurstone model, both estimators continue to perform well, and the
Thurstone MLE yields an error of the order $\frac{1}{\numitems}$
(Figure~\ref{fig:simulations_parametric_thurstone}).  Interestingly,
the Thurstone model also fits relatively well when data is generated
via the BTL model
(Figure~\ref{fig:simulations_parametric_BTL_uniform}). This behavior
is likely a result of operating in the near-linear regime of the
logistic and the Gaussian CDF where the two curves are similar. In
these two parametric settings, the SVT estimator has squared error
strictly worse than order $\frac{1}{\numitems}$ but better than
$\frac{1}{\sqrt{n}}$.  The Thurstone model, however, yields a poor fit
for the model in the high-SNR (Figure~\ref{fig:simulations_highSNR})
and the independent bands
(Figure~\ref{fig:simulations_independent-diagonals}) cases, incurring
a constant error as compared to an error scaling as
$\order(\frac{1}{\sqrt{\numitems}})$ for the SVT estimator. We recall
that the poor performance of the Thurstone estimator was also
described previously in Proposition~\ref{PropParametricBreak} and
Figure~\ref{FigParametricBreak}.

In summary, we see that while the Thurstone MLE estimator gives minimax optimal rates of estimation when the underlying model is parametric, it can be inconsistent when the parametric assumptions are violated. On the other hand, the SVT estimator is robust to violations of parametric assumptions, and while it does not necessarily give minimax-optimal rates, it remains consistent across the entire SST class.  Finally, we remark that our theory predicts that the least squares estimator, if implementable, would outperform both these estimators in terms of statistical error.


\section{Proofs of main results}
\label{SecProofs}
This section is devoted to the proofs of our main results--namely,
Theorems 1 through 5.  Throughout these and other proofs, we use the
notation $\{c, c', c_0, C, C' \}$ and so on to denote positive
constants whose values may change from line to line.  In addition, we
assume throughout that $\numitems$ is lower bounded by a universal
constant so as to avoid degeneracies.  For any square matrix $A \in
\reals^{\numitems \times \numitems}$, we let
$\{\singularvalue{1}{A},\ldots,\singularvalue{\numitems}{A} \}$ denote
its singular values (ordered from largest to smallest), and similarly,
for any symmetric matrix $M \in \real^{\numitems \times \numitems}$,
we let $\{ \eigenvalue{1}{M},\ldots,\eigenvalue{\numitems}{M} \}$
denote its ordered eigenvalues. The identity permutation is one where item $i$ is the $i^{th}$ most preferred item, for every $i \in [\numitems]$.


\subsection{Proof of Theorem~\ref{ThmMinimax}}
\label{AppThmMinimax}

This section is devoted to the proof of Theorem~\ref{ThmMinimax},
including both the upper and lower bounds on the minimax risk in
squared Frobenius norm.

\subsubsection{Proof of upper bound}
\label{AppThmMinimaxUpper}

Define the difference matrix $\mbox{\DelHat \defn \Mhat - \Mstar}$ between $\wtstar$ and
the optimal solution $\Mhat$ to the constrained least-squares problem.
Since $\Mhat$ is optimal and $\Mstar$ is feasible, we must have
$\frobnorm{Y - \Mhat}^2 \leq \frobnorm{Y - \Mstar}^2$, and hence
following some algebra, we arrive at the \emph{basic inequality} 
\begin{align}
\label{EqnBasicOne}
\frac{1}{2}
\frobnorm{\DelHat}^2 \leq \tracer{\DelHat}{\Wmat},
\end{align}
where $\Wmat \in \real^{\numitems \times \numitems}$ is the noise
matrix in the observation model~\eqref{EqnObservationModel}, and
$\tracer{A}{B} \defn \trace(A^T B)$ denotes the trace inner product.

We introduce some additional objects that are useful in our analysis.
The class of bivariate isotonic
matrices $\bisoclass$ is defined as
\begin{align}
\label{EqnDefnbisoclass}
\bisoclass \defn \big \{ M \in [0,1]^{\numitems \times \numitems} \mid
 \mbox{$M_{k \ell} \geq M_{ij}$ whenever $k \leq i$ and $\ell \geq
  j$} \big \}.
\end{align}
For a given permutation $\pi$ and matrix $M$, we let $\pi(M)$ denote
the matrix obtained by applying $\pi$ to its rows and columns.  
We then define the set
\begin{align}
\label{eq:EqnDefnChattDiff}
\chattDiff & \defn \Big \{ \pi_1(\wt_1) - \pi_2(\wt_2) \mid \mbox{for
  some }\wt_1, \; \wt_2 \in \bisoclass, \mbox{ and perm.  $\pi_1$ and
  $\pi_2$} \Big \}.
\end{align}
corresponding to the set of difference matrices.  Note that $
\chattDiff \subset [-1, 1]^{\numitems \times \numitems}$ by
construction.  One can verify that for any $\wtstar \in
\chatterjeeclass$, we are guaranteed the inclusion
\begin{align*}
\{ \wt - \wtstar \mid \wt \in \chatterjeeclass,~\frobnorm{\wt -
  \wtstar} \leq t \} \subset \chattDiff \cap \{
\frobnorm{\chattDiffmx} \leq t \}.
\end{align*}
Consequently, the error matrix $\DelHat$ must belong to $\chattDiff$,
and so must satisfy the properties defining this set.  Moreover, as we
discuss below, the set $\chattDiff$ is star-shaped, and this property
plays an important role in our analysis.

For each choice of radius $t > 0$, define the random variable
\begin{align}
\label{EqnDefnZ}
 Z(t) & \defn \sup_{\chattDiffmx \in \chattDiff,
   \frobnorm{\chattDiffmx} \leq t} \tracer{\chattDiffmx}{\Wmat}.
\end{align}
Using our earlier basic inequality~\eqref{EqnBasicOne}, the Frobenius norm error
$\frobnorm{\DelHat}$ then satisfies the bound
\begin{align}
\label{EqnBasic}
\frac{1}{2} \frobnorm{\DelHat}^2 & \leq \tracer{\DelHat}{\Wmat} \;
\leq \; Z \big( \frobnorm{\DelHat} \big).
\end{align}
Thus, in order to obtain a high probability bound, we need to
understand the behavior of the random quantity $Z(\delta)$.

One can verify that the set $\chattDiff$ is star-shaped, meaning that
$\alpha \chattDiffmx \in \chattDiff$ for every $\alpha \in [0,1]$ and
every $\chattDiffmx \in \chattDiff$. Using this star-shaped property,
we are guaranteed that there is a non-empty set of scalars $\delcrit >
0$ satisfying the critical inequality
\begin{align}
\label{EqnCritical}
\Exs[Z(\delcrit)] & \leq \frac{\delcrit^2}{2}.
\end{align}
Our interest is in the smallest (strictly) positive solution $\delcrit$ to the
critical inequality~\eqref{EqnCritical}, and moreover, our goal is to
show that for every $t \geq \delcrit$, we have $\frobnorm{\DelHat}
\leq c \sqrt{t \delcrit}$ with probability at least $1 - c_1 e^{-c_2 \numitems
  t \delcrit}$.

For each $t > 0$, define the ``bad'' event $\AuxEvent$ as
\begin{align}
\label{EqnDefnBadevent}
\AuxEvent & = \big \{ \exists \Delta \in \chattDiff \mid
\frobnorm{\Delta} \geq \sqrt{t \delcrit} \quad \mbox{and} \quad
\tracer{\Delta}{\Wmat} \geq 2 \frobnorm{\Delta} \sqrt{t \delcrit}
\big \}.
\end{align}
Using the star-shaped property of $\chattDiff$, it follows by a
rescaling argument that
\begin{align*}
\mprob[\AuxEvent] \leq \mprob\big[Z(\delcrit) \geq 2 \delcrit \sqrt{t
    \delcrit}\big] \qquad \mbox{for all $t \geq \delcrit$.}
\end{align*}
The entries of $\Wmat$ lie in $[-1,1]$, are i.i.d. on and above the diagonal, are zero-mean, and satisfy skew-symmetry. 
Moreover, the function $\Wmat \mapsto Z(t)$ is convex and Lipschitz
with parameter $t$.  Consequently, from known concentration
bounds(e.g.,~\cite[Theorem 5.9]{Ledoux01},~\cite{samson2000concentration}) for convex Lipschitz functions, we have
\begin{align*}
\mprob \big[ Z(\delcrit) \geq \Exs[Z(\delcrit)] + \sqrt{t \delcrit}
  \delcrit \big] & \leq 2 e^{-c_1 t \delcrit} \qquad \mbox{for all $t
  \geq \delcrit$.}
\end{align*}
By the definition of $\delcrit$, we have $\Exs[Z(\delcrit)] \leq
\delcrit^2 \leq \delcrit \sqrt{t \delcrit}$ for any $t \geq \delcrit$,
and consequently
\begin{align*}
\mprob[\AuxEvent] & \leq \mprob[Z(\delcrit) \geq 2 \delcrit \sqrt{t
    \delcrit} \big] \; \leq \; 2 e^{-c_1 t \delcrit} \quad \mbox{for
  all $t \geq \delcrit$.}
\end{align*}
Consequently, either $\frobnorm{\DelHat} \leq \sqrt{t \delcrit}$, or
we have $\frobnorm{\DelHat} > \sqrt{t \delcrit}$. In the latter case, conditioning on
the complement $\AuxEvent^c$, our basic inequality implies that
$\frac{1}{2} \frobnorm{\DelHat}^2 \leq 2 \frobnorm{\DelHat} \sqrt{t
  \delcrit}$, and hence $\frobnorm{\DelHat} \leq 4 \sqrt{t \delcrit}$
with probability at least $1 - 2 e^{- c_1 t \delcrit}$.  Putting
together the pieces yields that
\begin{align}
\frobnorm{\DelHat} \leq c_0 \sqrt{t \delcrit}
\label{EqWithDelCritHP}
\end{align}
with probability at least $1- 2 e^{-c_1 t \delcrit}$ for every $t \geq
\delcrit$.  

In order to determine a feasible $\delcrit$ satisfying the critical
inequality~\eqref{EqnCritical}, we need to bound the expectation
$\Exs[Z(\delcrit)]$. We do using Dudley's entropy integral and
bounding the metric entropies of certain sub-classes of matrices.  In
particular, the remainder of this section is devoted to proving the
following claim:

\begin{lemma}
\label{LemCritical}
There is a universal constant $C$ such that
\begin{align}
\label{EqnCriticalBound}
\Exs[Z(t)] & \leq C \, \Big \{ \numitems \log^2(\numitems) + t \,
\sqrt{\numitems \log \numitems} \Big \},
\end{align}
 for all $t \in [0, 2
 \numitem]$.
\end{lemma}
\noindent Given this lemma, we see that the
critical inequality~\eqref{EqnCritical} is satisfied with
\mbox{$\delcrit = C' \sqrt{\numitems} \log \numitems$.}  Consequently,
from our bound~\eqref{EqWithDelCritHP}, there are universal positive
constants $C''$ and $c_1$ such that
\begin{align*}
\frac{\frobnorm{\DelHat}^2}{\numitems^2} & \leq C''
\frac{\log^2(\numitems)}{\numitems},
\end{align*}
with probability at least $1- 2 e^{-c_1 \numitems (\log
	\numitems)^2}$, 
which completes the proof.
\\

\noindent {\bf Proof of Lemma~\ref{LemCritical}:} 
It remains to prove Lemma~\ref{LemCritical}, and we do so by using
Dudley's entropy integral, as well as some auxiliary results on metric
entropy. We use the notation $\metent(\epsilon,\mathbb{C},\rho)$ to
denote the $\epsilon$ metric entropy of the class $\mathbb{C}$ in the
metric $\rho$.  Our proof requires the following auxiliary lemma:
\begin{lemma}
\label{LemChattMetent}
For every $\epsilon >0$, we have the metric entropy bound
\begin{align*}
\metent(\epsilon, \chattDiff, \frobnorm{.}) & \leq
9\frac{\numitems^2}{\epsilon^2} \big(\log \frac{\numitems}{\epsilon}
\big)^2 + 9\numitems \log \numitems.
\end{align*}
\end{lemma}
\noindent 
See the end of this section for the proof of this claim. Letting
$\Ball_F(t)$ denote the Frobenius norm ball of radius $t$, the
truncated form of Dudley's entropy integral inequality
(e.g.,~\cite[Corollary 2.2.8]{van1996weak}) yields
that the mean $\Exs[ Z(t)]$ is upper bounded as
\begin{align}
\Exs[ Z(t)] ] & \leq \plaincon\; \inf_{\delta \in [0,\numitems]} \Big
  \{ \numitems \delta + \int_{\frac{\delta}{2}}^{t} \sqrt{
    \metent(\epsilon,\chattDiff \cap \BallF(t),\frobnorm{.})}
  d\epsilon \Big \} \notag \\
\label{EqnEarly}
& \leq \plaincon \; \Big \{ \numitems^{-8} + \int_{\frac{1}{2}
  \numitems^{-9}}^{t} \sqrt{
  \metent(\epsilon,\chattDiff,\frobnorm{.})} d\epsilon \Big \},
\end{align}
where the second step follows by setting $\delta = \numitems^{-9}$,
and making use of the set inclusion \mbox{$(\chattDiff \cap \BallF(t))
  \subseteq \chattDiff$.} For any $\epsilon \geq \frac{1}{2}
\numitems^{-9}$, applying Lemma~\ref{LemChattMetent} yields the upper
bound
\begin{align*}
\sqrt{\metent(\epsilon, \chattDiff, \frobnorm{.})} \leq \plaincon \Big \{
\frac{\numitems}{\epsilon} \log \frac{\numitems}{\epsilon} +
\sqrt{\numitems \log \numitems} \Big \}.
\end{align*}
Over the range $\epsilon \geq \numitems^{-9}/2$, we have $\log
\frac{\numitems}{\epsilon} \leq \plaincon \log \numitems$, and hence
\begin{align*}
\sqrt{\metent(\epsilon, \chattDiff, \frobnorm{.})} \leq \plaincon \Big \{
\frac{\numitems}{\epsilon} \log \numitems + \sqrt{n \log n} \Big \}.
\end{align*}
Substituting this bound into our earlier inequality~\eqref{EqnEarly}
yields
\begin{align*}
\Exs [Z(t) ] & \leq \plaincon \Big \{ \numitems^{-8} + \big( \numitems
\log \numitems\big) \log (\numitems t) + t \sqrt{\numitems \log
  \numitems} \Big \} \\
& \stackrel{(i)}{\leq} \plaincon \Big \{ \big( \numitems \log \numitems
\big) \log (\numitems^2) + t \sqrt{\numitems \log \numitems} \Big \}
\\
& \leq \plaincon \Big \{ \numitems \log^2(\numitems) + t \sqrt{\numitems
  \log \numitems } \Big \},
\end{align*}
where step (i) uses the upper bound $t \leq 2 \numitems$. \\


\noindent The only remaining detail is to prove
Lemma~\ref{LemChattMetent}.\\


\paragraph{Proof of Lemma~\ref{LemChattMetent}:}

We first derive an upper bound on the metric entropy of the class
$\bisoclass$ defined previously in
equation~\eqref{EqnDefnbisoclass}. In particular, we do so by relating
it to the set of all bivariate monotonic functions on the square
$[0,1] \times [0,1]$.  Denoting this function class by $\monoGW$, for
any matrix $\biisomx \in \bisoclass$, we define a function
$\monoGWfn_{\biisomx} \in \monoGW$ via
\begin{align*}
\monoGWfn_\biisomx(x,y) = \biisomx_{\lceil \numitems (1-x) \rceil,
	\lceil \numitems y \rceil}.
\end{align*}
In order to handle corner conditions, we set
\mbox{$\biisomx_{0,i}=\biisomx_{1,i}$} and
\mbox{$\biisomx_{i,0}=\biisomx_{i,1}$} for all $i$. With this
definition, we have
\begin{align*}
\|\monoGWfn_\biisomx\|_2^2 = \int_{x=0}^1 \int_{y=0}^1
(\monoGWfn_\biisomx(x,y))^2 dx dy = \frac{1}{\numitems^2}
\sum_{i=1}^{\numitems} \sum_{j=1}^{\numitems} \biisomx_{i,j}^2 =
\frac{1}{\numitems^2} \frobnorm{\biisomx}^2.
\end{align*}
As a consequence, the metric entropy can be upper bounded as
\begin{align}
\metent( \epsilon ,\bisoclass,\frobnorm{.}) & \leq \;
\metent\left(\frac{\epsilon}{n}, \monoGW, \|.\|_2\right) \nonumber \\ 
&\stackrel{(i)}{\leq} \;
\frac{\numitems^2}{\epsilon^2} \big(\log \frac{\numitems}{\epsilon}
\big)^2,
\label{eq:biiso_metent}
\end{align}
where inequality (i) follows from Theorem 1.1 of Gao and
Wellner~\cite{gao2007entropy}.

We now bound the metric entropy of $\chattDiff$ in terms of the metric
entropy of $\bisoclass$. For any $\epsilon > 0$, let
$\bisoclass^\epsilon$ denote an $\epsilon$-covering set in
$\bisoclass$ that satisfies the
inequality~\eqref{eq:biiso_metent}. Consider the set
\begin{align*}
\chattDiff^{\epsilon} \defn \{ \perm_1(\biisomx_1) -
\perm_2(\biisomx_2) \mid \mbox{for some permutations $\perm_1$,
  $\perm_2$ and some $\biisomx_1, \biisomx_2 \in
  \bisoclass^{\epsilon/{2}}$} \}.
\end{align*}
For any $\chattDiffmx \in \chattDiff$, we can write $\chattDiffmx =
\perm_1(\biisomx_1') - \perm_2(\biisomx_2')$ for some permutations
$\perm_1$ and $\perm_2$ and some matrices $\biisomx_1'$ and
$\biisomx_2' \in \bisoclass$. We know there exist matrices
\mbox{$\biisomx_1, \biisomx_2 \in \bisoclass^{\epsilon/{2}}$} such
that $\frobnorm{\biisomx_1' - \biisomx_1} \leq \epsilon/{2}$ and
$\frobnorm{\biisomx_2' - \biisomx_2} \leq \epsilon/{2}$.  With these
choices, we have \mbox{$\perm_1(\biisomx_1) - \perm_2(\biisomx_2) \in
	\chattDiff^{\epsilon}$,} and moreover
\begin{align*}
\frobnorm{ \chattDiffmx - (\perm_1(\biisomx_1) - \perm_2(\biisomx_2) )
}^2 & \leq 2\frobnorm{ \perm_1(\biisomx_1) - \perm_1(\biisomx_1')}^2 +
2\frobnorm{\perm_2(\biisomx_2) - \perm_1(\biisomx_2') }^2 \\ & \leq
\epsilon^2.
\end{align*}
Thus the set $\chattDiff^\epsilon$ forms an $\epsilon$-covering set
for the class $\chattDiff$. One can now count the number of elements
in this set to find that
\begin{align*}
\covnum(\epsilon, \chattDiff,\frobnorm{.}) \leq \big(
\factorial{\numitems} \covnum(\epsilon/{2} ,
\bisoclass,\frobnorm{.}) \big)^2.
\end{align*}
Some straightforward algebraic manipulations yield the claimed result.


\subsubsection{Proof of lower bound}
\label{AppThmMinimaxLower}

We now turn to the proof of the lower bound in
Theorem~\ref{ThmMinimax}.  We may assume that the correct row/column
ordering is fixed and known to be the identity permutation. Here
we are using the fact that revealing the knowledge of this ordering cannot
make the estimation problem any harder.  Recalling the
definition~\eqref{EqnDefnbisoclass} of the bivariate isotonic class
$\bisoclass$, consider the subclass
\begin{align*}
\chatterjeeclass' \defn \{ M \in \bisoclass \mid \mbox{$M_{i,j}=1$
  when $j>i+1$ and $M_{i,j} = 1-M_{j,i}$ when $j \leq i$} \}
\end{align*}
Any matrix $M$ is this subclass can be identified with the vector $q =
q(M) \in \real^{\numitems-1}$ with elements $q_i \defn M_{i,i+1}$. The
only constraint imposed on $q(M)$ by the inclusion $M \in
\chatterjeeclass$ is that $q_i \in [ \myhalf, 1]$ for all $i = 1,
\ldots, \numitems -1$.

In this way, we have shown that the difficulty of estimating $\wtstar
\in \chatterjeeclass'$ is at least as hard as that of estimating a
vector $q \in [\myhalf, 1]^{\numitems-1}$ based on observing the
random vector $Y = \{Y_{1,2},\ldots,Y_{\numitems-1,\numitems}\}$ with
independent coordinates, and such that each $Y_{i,i+1} \sim
\operatorname{Ber}(q_i)$.  For this problem, it is easy to show that
there is a universal constant $\ULOW > 0$ such that
\begin{align*}
\inf_{\qhat } \sup_{q \in [\myhalf, 1]^{\numitems-1}} \Exs \Big[
  \|\qhat - q \|_2^2 \Big] & \geq \frac{\ULOW}{2} \numitems,
\end{align*}
where the infimum is taken over all measurable functions $Y \mapsto
\qhat$.  Putting together the pieces, we have shown that
\begin{align*}
\inf_{\wthat} \sup_{\wtstar \in \chatterjeeclass}
\frac{1}{\numitems^2} \Exs [ \frobnorm{\hat{M}-M^*}^2 ] & \geq
\frac{2}{\numitems^2} \inf_{\qhat} \sup_{q \in [0.5,1]^{\numitems-1}}
\Exs [ \|\qhat-q\|_2^2 ] \geq \frac{\ULOW}{\numitems},
\end{align*}
as claimed.

\subsection{Proof of Theorem~\ref{ThmImprovedUSVT}}
\label{AppThmImprovedUSVT}

Recall from equation~\eqref{EqnObservationModel} that we can write our observation model as $Y = \Mstar + W$,
where $W \in \real^{\numitem \times \numitem}$ is a zero-mean matrix
with entries that are drawn independently (except for the
skew-symmetry condition) from the interval $[-1,1]$.

\subsubsection{Proof of upper bound}

Our proof of the upper bound hinges upon the following two lemmas.
\begin{lemma}
\label{LemSTSVD}
If $\regparn \geq 1.01 \opnorm{\noise}$, then
\begin{align*}
\frobnorm{\Treg(Y) - \Mstar}^2 & \leq \plaincon \sum_{j=1}^\numitem \min \big \{
\regparnsq, \sigma_j^2(\Mstar) \big \},
\end{align*}
where $\plaincon$ is a positive universal constant.
\end{lemma}

\noindent Our second lemma is an approximation-theoretic result:
\begin{lemma}
\label{LemSTSVDChatterjee}
For any matrix $\Mstar \in \chatterjeeclass$ and any $s \in \{1, 2,
\ldots, \numitem-1 \}$, we have
\begin{align*}
\frac{1}{\numitems^2} \sum_{j=s+1}^\numitem \sigma^2_j(\Mstar) & \leq
\frac{1}{s}.
\end{align*}
\end{lemma}
\noindent See the end of this section for the proofs of these two
auxiliary results.\footnote{As a side note, in Section~\ref{proof:USVT_lower} we present a
construction of a matrix $\Mstar \in \chatterjeeclass$ using which we
show that the bound of Lemma~\ref{LemSTSVDChatterjee} is sharp up to a
constant factor when $s = o(\numitems)$; this result is essential in
proving the sharpness of the result of Theorem~\ref{ThmImprovedUSVT}.}

Based on these two lemmas, it is easy to complete the proof of the
theorem. The entries of $\Wmat$ are zero-mean with entries in the
interval $[-1,1]$, are i.i.d. on and above the diagonal, and satisfy
skew-symmetry.  Consequently, we may apply Theorem 3.4 of
Chatterjee~\cite{chatterjee2014matrix}, which guarantees that
\begin{align*}
\mprob \Big[ \opnorm{\Wmat} > (2 + t) \sqrt{\numitem} \Big] & \leq c
e^{- f(t) \numitem},
\end{align*}
where $c$ is a universal constant, and the quantity $f(t)$ is strictly
positive for each $t > 0$.  Thus, the choice \mbox{$\regparn = 2.1
	\sqrt{\numitems}$} guarantees that \mbox{$\regparn \geq 1.01
	\opnorm{\Wmat}$} with probability at least \mbox{$1 - c e^{-c
		\numobs}$,} as is required for applying Lemma~\ref{LemSTSVD}.
Applying this lemma guarantees that the upper bound
\begin{align*}
\frobnorm{\Treg(Y) - \Mstar}^2 & \leq c \sum_{j=1}^\numitem \min \big
\{ \numitems, \sigma_j^2(\Mstar) \big \}
\end{align*}
hold with probability at least $1 - c_1 e^{-c_2 \numitem}$. From
Lemma~\ref{LemSTSVDChatterjee}, with probability at least $1 - c_1
e^{-c_2 \numitem}$, we have
\begin{align*}
\frac{1}{\numitem^2}\frobnorm{\Treg(Y) - \Mstar}^2 & \leq c \Big\{
\frac{s}{\numitem} + \frac{1}{s} \Big \} 
\end{align*}
for all $s \in \{ 1, \ldots, \numitem \}$.  
Setting $s = \lceil \sqrt{\numitem} \rceil$ and performing some
algebra shows that
\begin{align*}
\mprob \Big[\frac{1}{\numitem^2}\frobnorm{\Treg(Y) - \Mstar}^2 >
  \frac{\UUP}{\sqrt{\numitem}} \Big] & \leq c_1 e^{-c_2 \numitem},
\end{align*}
as claimed.
Since $\frac{1}{\numitem^2}\frobnorm{\Treg(Y) - \Mstar}^2 \leq 1$,
we are also guaranteed that
\begin{align*}
\frac{1}{\numitem^2} \Exs[\frobnorm{\Treg(Y) - \Mstar}^2] & \leq
\frac{\UUP}{\sqrt{\numitem}} + c_1 e^{-c_2 \numitem} \; \leq
\frac{\UUP'}{\sqrt{\numobs}}.
\end{align*}

\paragraph{Proof of Lemma~\ref{LemSTSVD}}
Fix $\delta = 0.01$. Let $b$ be the number of singular values of $\Mstar$ above $\frac{\delta}{1+\delta} \regparn$, and let $\Mstar_b$ be the version of $\Mstar$
truncated to its top $b$ singular values.  We then have
\begin{align*}
\frobnorm{\Treg(Y) - \Mstar}^2 & \leq 2 \frobnorm{\Treg(Y) -
  \Mstar_b}^2 + 2 \frobnorm{\Mstar_b - \Mstar}^2 \\
& \leq 2 \rank(\Treg(Y) - \Mstar_b) \opnorm{\Treg(Y) - \Mstar_b}^2 + 2
\sum_{j=b + 1}^\numitem \sigma_j^2(\Mstar).
\end{align*}
We claim that $\Treg(Y)$ has rank at most $b$.  Indeed, for any $j
\geq b+1$, we have
\begin{align*}
\sigma_j(Y) & \leq \sigma_j(\Mstar) + \opnorm{\Wmat} \; \leq \; \regparn,
\end{align*} 
where we have used the facts that $\singularvalue{j}{\Mstar} \leq \frac{\delta}{1+\delta} \regparn$ for every $j \geq b+1$ and $\regparn \geq (1+\delta) \opnorm{\noise}$. As a consequence we have $\sigma_j(\Treg(Y)) = 0$, and hence
$\rank(\Treg(Y) - \Mstar_b) \leq 2 b$.  Moreover, we have
\begin{align*}
\opnorm{\Treg(Y) - \Mstar_b} & \leq \opnorm{\Treg(Y) - Y} + \opnorm{Y
  - \Mstar} + \opnorm{\Mstar - \Mstar_b}\\
& \leq \regparn + \opnorm{\Wmat} + \frac{\delta}{1+\delta} \regparn \\
& \leq 2 \regparn.
\end{align*}
Putting together the pieces, we conclude that
\begin{align*}
\frobnorm{\Treg(Y) - \Mstar}^2 & \leq 16 b \regparn^2 + 2
\sum_{j=b+1}^\numitem \sigma_j^2(\Mstar) \; \stackrel{(i)}{\leq} \; C
\sum_{j=1}^\numitem \min \{ \sigma_j^2(\Mstar), \regparn^2 \},
\end{align*}
for some constant\footnote{To be clear, the precise value of the
  constant $C$ is determined by $\delta$, which has been fixed as
  $\delta = 0.01$.}  $C$.  Here inequality (i) follows since
$\sigma_j(\Mstar) \leq \frac{\delta}{1+\delta} \regparn$ whenever $j
\geq b+1$ and $\sigma_j(\Mstar) > \frac{\delta}{1+\delta} \regparn$
whenever $j \leq b$.


\paragraph{Proof of Lemma~\ref{LemSTSVDChatterjee}}

In this proof, we make use of a construction due to
Chatterjee~\cite{chatterjee2014matrix}.  For a given matrix $\wtstar$,
we can define the vector $t \in \real^\numitems$ of row sums---namely,
with entries $t_i = \sum_{j=1}^\numitems \wtstar_{ij}$ for $i \in
[\numitems]$.  Using this vector, we can define a rank $s$
approximation $\wt$ to the original matrix $\wtstar$ by grouping the
rows according to the vector $t$ according to the following procedure:
\bcar
\item
Observing that each $t_i \in [0,\numitems]$, let us divide the full
interval $[0,\numitems]$ into $s$ groups---say of the form
$[0,\numitems/s), [\numitems/s, 2\numitems/s), \ldots
    [(s-1)\numitems/s,\numitems]$.  If $t_i$ falls into the interval
    $\alpha$ for some $\alpha \in [s]$, we then map row $i$ to the
    group $G_\alpha$ of indices.
\item For each group $G_\alpha$, we choose a particular row index $k =
  k(\alpha) \in G_\alpha$ in an arbitrary fashion.  For every other
  row index $i \in G_\alpha$, we set $\wt_{ij} = \wt_{k j}$ for all $j
  \in [\numitems]$.
\ecar

\vspace*{.04in}

By construction, the matrix $\wt$ has at most $s$ distinct rows, and
hence rank at most $s$.  Let us now bound the Frobenius norm error in
this rank $s$ approximation.  Fixing an arbitrary group index $\alpha
\in [s]$ and an arbitrary row in $i \in G_\alpha$, we then have
\begin{align*}
\sum_{j=1}^\numitems (\wtstar_{ij} - \wt_{ij})^2 \leq
\sum_{j=1}^\numitems | \wtstar_{ij} - \wt_{ij}|.
\end{align*}
By construction, we either have $\wtstar_{ij} \geq \wt_{ij}$ for every
$j \in [\numitems]$, or $\wtstar_{ij} \leq \wt_{ij}$ for every $j \in
[\numitems]$. Thus, letting $k \in G_\alpha$ denote the chosen row, we
are guaranteed that
\begin{align*}
\sum_{j=1}^\numitems |\wtstar_{ij} - \wt_{ij}| \leq |t_i - t_{k}| \leq
\frac{\numitems}{s},
\end{align*}
where we have used the fact the pair $(t_i, t_k)$ must lie in an
interval of length at most $\numitems/s$.  Putting together the pieces
yields the claim.

\subsubsection{Proof of lower bound}
\label{proof:USVT_lower}
We now turn to the proof of the lower bound in
Theorem~\ref{ThmImprovedUSVT}.  We split our analysis into two cases,
depending on the magnitude of $\regparn$.

\paragraph{Case 1:}
First suppose that $\regparn \leq \frac{\sqrt{\numitems}}{3}$. In this
case, we consider the matrix $\wtstar \defn \half 1 1^T$ in which all
items are equally good, so any comparison is simply a fair coin flip.
Let the observation matrix \mbox{$\obs \in \{0,1\}^{\numitems \times
    \numitems}$} be arbitrary.  By definition of the singular value
thresholding operation, we have $\opnorm{Y - \Treg(Y)} \leq \regparn$,
and hence the SVT estimator $\wthatUSVT = \Treg(Y)$ has Frobenius norm
at most
\begin{align*}
\frobnorm{\obs - \wthatUSVT}^2 \leq \numitems \regparn^2 & \leq
\frac{\numitems^2}{9}.
\end{align*}
Since \mbox{$\wtstar \in \{\half\}^{\numitems \times \numitems}$} and
\mbox{$\obs \in \{0,1\}^{\numitems \times \numitems}$,} we are
guaranteed that $\frobnorm{\wtstar - \obs} = \frac{\numitems}{2}$.
Applying the triangle inequality yields the lower bound
\begin{align*}
\frobnorm{\wthatUSVT - \wtstar} \; \geq \: \frobnorm{\wtstar - \obs} -
\frobnorm{\wthatUSVT - \obs} \; \geq \frac{\numitems}{2}
-\frac{\numitems}{3} \; = \; \frac{\numitems}{6}.
\end{align*}

\paragraph{Case 2:} Otherwise, we may assume that
 $\regparn > \frac{\sqrt{\numitems}}{3}$.  Consider the matrix
$\wtstar \in \real^{\numitems \times \numitems}$ with entries
\begin{align}
[\wtstar]_{ij} =
\begin{cases}
1 & \mbox{if~}i>j\\ \half & \mbox{if~}i=j\\ 0 & \mbox{if~}i<j.
\end{cases}
\label{EqnDefnNoiseless}
\end{align}
By construction, the matrix $\wtstar$ corresponds to the degenerate
case of noiseless comparisons.

Consider the matrix $\obs \in \real^{\numitems \times \numitems}$
generated according to the observation
model~\eqref{EqnObservationModel}.  (To be clear, all of its
off-diagonal entries are deterministic, whereas the diagonal is
population with i.i.d. Bernoulli variates.)  Our proof requires the
following auxiliary result regarding the singular values of
$\obs$:
\begin{lemma}
\label{LemNoiselessSpectrum}
The singular values of the observation matrix  $\obs \in \real^{\numitems \times
  \numitems}$ generated by the noiseless comparison matrix $\wtstar$
satisfy the bounds
\begin{align*} 
\frac{\numitems}{4 \pi (i+1)} - \half \leq \singularvalue{\numitems -i
  - 1}{\obs} \leq \frac{\numitems}{\pi (i-1)} + \half \qquad \mbox{for
  all integers $i \in [1, \frac{\numitems}{6}-1]$.}
\end{align*}
\end{lemma}
\noindent We prove this lemma at the end of this section.

Taking it as given, we get that $\singularvalue{\numitems -i - 1}{\obs}
\leq \frac{\sqrt{\numitems}}{3}$ for every integer $i \geq 2
\sqrt{\numitems}$, and $\singularvalue{\numitems -i}{\obs} \geq
\frac{\numitems}{50 i}$ for every integer $i \in [1,
  \frac{\numitems}{25}]$.  It follows that
\begin{align*}
\sum_{i=1}^{\numitems} (\singularvalue{i}{\obs})^2
\indicator{\singularvalue{i}{\obs} \leq \frac{\sqrt{\numitems}}{3}} &
\geq \frac{\numitems^2}{2500} \sum_{i =
  2\sqrt{\numitems}}^{\frac{\numitems}{25}} \frac{1}{i^2} \; \geq \;
\plaincon \numitems^{\frac{3}{2}},
\end{align*}
for some universal constant $\plaincon > 0$. 
Recalling that $\regparn \geq \frac{\sqrt{\numitems}}{3}$, we have the lower bound $\frobnorm{\obs -
  \wthatUSVT}^2 \geq \plaincon \numitems^{\frac{3}{2}}$.  Furthermore,
since the observations (apart from the diagonal entries) are
noiseless, we have $\frobnorm{\obs - \wtstar}^2 \leq
\frac{\numitems}{4}$.  Putting the pieces together yields the lower
bound
\begin{align*}
\frobnorm{\wthatUSVT - \wtstar} \; \geq \frobnorm{\wthatUSVT - \obs} -
\frobnorm{\wtstar - \obs} & \geq \; \plaincon \numitems^{\frac{3}{4}}
- \frac{\sqrt{\numitems}}{2} \; \geq \; \plaincon'
\numitems^{\frac{3}{4}},
\end{align*}
where the final step holds when $\numitems$ is large enough (i.e.,
larger than a universal constant).

%
\paragraph{Proof of Lemma~\ref{LemNoiselessSpectrum}:}

Instead of working with the original observation matrix $\obs$, it is
convenient to work with a transformed version.  Define the matrix
\mbox{$\obsinter \defn \obs - \diag{\obs} + I_{\numitem}$}, so that the matrix
$\obsinter$ is identical to $\obs$ except that all its diagonal
entries are set to $1$.  Using this intermediate object, define the
$(\numitems \times \numitems)$ matrix
\begin{align}
\label{EqnDefnObstil}
\obstil & \defn (\obsinter (\obsinter)^T)^{-1} - e_\numitems
e_\numitems^T,
\end{align} 
where $e_\numitems$ denotes the $\numitems^{\rm th}$ standard basis
vector.  One can verify that this matrix has entries
\begin{align*}
[\obstil]_{ij} =
\begin{cases}
1 & \mbox{if~} i = j = 1 \mbox{~or~} i = j = \numitems\\ 2 &
\mbox{if~} 1 < i = j < \numitems\\ -1 & \mbox{if~} i = j + 1
\mbox{~or~} i = j -1\\ 0 & \mbox{otherwise}.
\end{cases}
\end{align*}
Consequently, it is equal to the graph Laplacian\footnote{In
  particular, the Laplacian of a graph is given by $L = D - A$, where
  $A$ is the graph adjacency matrix, and $D$ is the diagonal degree
  matrix.} of an undirected chain graph on $\numitems$ nodes.
Consequently, from standard results in spectral graph
theory~\cite{brouwer2011spectra}, the eigenvalues of $\obstil$ are
given by $\{ 4 \sin^2(\frac{\pi i}{\numitems}) \}_{i =
  0}^{\numitems-1}$.  Recall the elementary sandwich relationship
$\frac{x}{2} \leq \sin x \leq x$, valid for every $x \in
[0,\frac{\pi}{6}]$.  Using this fact, we are guaranteed that
\begin{align}
\label{EqnInterBound}
\frac{\pi^2 i^2}{\numitems^2} \leq \eigenvalue{i+1}{\obstil} \leq
\frac{4 \pi^2 i^2}{\numitems^2} \quad \mbox{for all integers $i \in
  [1, \frac{\numitems}{6}]$.}
\end{align}

We now use this intermediate result to establish the claimed bounds on
the singular values of $\obs$.  Observe that the matrices $\obstil$
and $(\obsinter (\obsinter)^T)^{-1}$ differ only by the rank one
matrix $e_\numitems e_\numitems^T$.  Standard results in matrix
perturbation theory~\cite{thompson1976behavior} guarantee that a
rank-one perturbation can shift the position (in the large-to-small
ordering) of any eigenvalue by at most one.  Consequently, the
eigenvalues of the matrix $(\obsinter (\obsinter)^T)^{-1}$ must be
sandwiched as
\begin{align*}
\frac{\pi^2 (i-1)^2}{\numitems^2} \leq \eigenvalue{i+1}{(\obsinter
  (\obsinter)^T)^{-1}} \leq \frac{4 \pi^2 (i+1)^2}{\numitems^2} \qquad
\mbox{for all integers $i \in [1, \frac{\numitems}{6}-1]$.}
\end{align*}
It follows that the singular values of $\obsinter$ are sandwiched as
\begin{align*}
\frac{\numitems}{4 \pi (i+1)} \leq \singularvalue{\numitems - i -
  1}{\obsinter} \leq \frac{\numitems}{\pi (i-1)} \qquad \mbox{for all
  integers $i \in [1, \frac{\numitems}{6}-1]$.}
\end{align*}

Observe that $\obsinter - \obs$ is a $\{0,\half\}$-valued diagonal
matrix, and hence \mbox{$\opnorm{\obsinter - \obs} \leq \frac{1}{2}$.}
Consequently, we have \mbox{$\max_{i=1, \ldots, \numitem}
	|\sigma_i(\obs) - \sigma_i(\obsinter)| \leq \half$,} from which it
follows that
\begin{align*} 
\frac{\numitems}{4 \pi (i+1)} - \half \leq \singularvalue{\numitems -i
	- 1}{\obs} \leq \frac{\numitems}{\pi (i-1)} + \half
\end{align*}
as claimed.


\subsection{Proof of Theorem~\ref{ThmHighSNR}}
\label{AppThmHighSNR}

We now prove our results on the high SNR subclass of
$\chatterjeeclass$, in particular establishing a lower bound and then
analyzing the two-stage estimator described in
Section~\ref{SecHighSNR} so as to obtain the upper bound.


\subsubsection{Proof of lower bound}

In order to prove the lower bound, we follow the proof of the lower
bound of Theorem~\ref{ThmMinimax}, with the only difference being that
the vector $q \in \real^{\numitems-1}$ is restricted to lie in the
interval $[\half + \bias, 1]^{\numitems-1}$.


\subsubsection{Proof of upper bound}

Without loss of generality, assume that the true matrix $\Mstar$ is
associated to the identity permutation.  Recall that the second step
of our procedure involves performing constrained regression over the
set $\bisoclass(\permfas)$.  The error in such an estimate is
necessarily of two types: the usual estimation error induced by the
noise in our samples, and in addition, some form of approximation
error that is induced by the difference between $\permfas$ and the
correct identity permutation.

In order to formalize this notion, for any fixed permutation $\perm$,
consider the constrained least-squares estimator
\begin{align}
\label{EqnConstrainedLSTwo}
\wthat_\perm \in \argmin_{\wt \in \bisoclasssiva{\perm}}
\frobnorm{\obs - \wt}^2.
\end{align}
Our first result provides an upper bound on the error matrix $\wthat_\perm - \Mstar$ that involves both approximation and
estimation error terms.
\begin{lemma}
\label{lem:perm_to_mx}
There is a universal constant $\plaincon_0 > 0$ such that error in the
constrained LS estimate~\eqref{EqnConstrainedLSTwo} satisfies the
upper bound
\begin{align}
\label{EqnApproxEst}
\frac{\frobnorm{\wthat_\perm - \Mstar}^2}{\plaincon_0} & \leq
\underbrace{\frobnorm{\Mstar -
    \perm(\Mstar)}^2}_{\mbox{Approx. error}} + \underbrace{\numitem
  \log^2(\numitem)}_{\mbox{Estimation error}}
\end{align}
with probability at least $1 - c_1 e^{-c_2 \numitem}$.
\end{lemma}

There are two remaining challenges in the proof.  Since the second
step of our estimator involves the FAS-minimizing permutation
$\permfas$, we cannot simply apply Lemma~\ref{lem:perm_to_mx} to it
directly.  (The permutation $\permfas$ is random, whereas this lemma
applies to any fixed permutation).  Consequently, we first need to
extend the bound~\eqref{EqnApproxEst} to one that is uniform over a
set that includes $\permfas$ with high probability.  Our second
challenge is to upper bound the approximation error term
$\frobnorm{\Mstar - \permfas(\Mstar)}^2$ that is induced by using the
permutation $\permfas$ instead of the correct identity permutation.

In order to address these challenges, for any constant $\BRACON > 0$,
define the set
\begin{align*}
\permhatclass(\BRACON) \defn \{\perm \mid \max_{i \in [\numitems]} |i
- \perm(i)| \leq \BRACON \log \numitems\}.
\end{align*}
This set corresponds to permutations that are relatively close to the
identity permutation in the sup-norm sense.  Our second lemma shows
that any permutation in $\permhatclass(\BRACON)$ is ``good enough'' in
the sense that the approximation error term in the upper
bound~\eqref{EqnApproxEst} is well-controlled:
\begin{lemma}
\label{lem:FAS}
For any $\wtstar \in \bisoclass$ and any permutation $\perm \in
\permhatclass(\BRACON)$, we have
\begin{align}
\label{EqnApproxBound}
\frobnorm{\wtstar - \perm(\wtstar)}^2 \leq 2 \plaincon'' \numitems
\log \numitems,
\end{align}
where $\plaincon''$ is a positive constant that may depend only on $\BRACON$.
\end{lemma}

Taking these two lemmas as given, let us now complete the proof of
Theorem~\ref{ThmHighSNR}.  (We return to prove these lemmas at the end
of this section.)  Braverman and Mossel~\cite{braverman2008noisy}
showed that for the class $\chatterjeeclassbias(\bias)$, there exists
a positive constant $\BRACON$---depending on $\bias$ but independent
of $\numitems$---such that
\begin{align}
\label{EqnBraMos}
\mprob \Big[ \permfas \in \permhatclass(\BRACON) \Big] &
\geq 1 - \frac{\plaincon_3}{\numitems^2}.
\end{align}
From the definition of class $\permhatclass(\BRACON)$, there
is a positive constant $\BRACON'$ (whose value may depend only on $\BRACON$) such that its cardinality is upper
bounded as 
\begin{align*}
\operatorname{card}(\permhatclass(\BRACON)) \leq \numitems^{2 \BRACON'
	\log \numitems} \stackrel{(i)}{\leq} e^{ .5 \plaincon_2 \numitems},
\end{align*}
where the inequality (i) is valid once the number of items
$\numitems$ is larger than some universal constant.  
  Consequently, by combining
the union bound with Lemma~\ref{lem:perm_to_mx} we conclude that, with
probability at least $1 - c_1' e^{-c_2' \numitems} -
\frac{\plaincon_3}{\numitem^2}$, the error matrix $\DelFAS \defn
\wthat_{\permfas} - \Mstar$ satisfies the upper
bound~\eqref{EqnApproxEst}.  Combined with the approximation-theoretic
guarantee from Lemma~\ref{lem:FAS}, we find that
\begin{align*}
\frac{\frobnorm{\DelFAS}^2}{\plaincon_0} & \leq \frobnorm{\Mstar -
  \permfas(\Mstar)}^2 + \numitem \log^2(\numitem) \\
& \leq \BRACON'' \numitem \log \numitem +
+ \numitem \log^2(\numitem),
\end{align*}
from which the claim follows. \\

\noindent It remains to prove the two auxiliary lemmas, and we do so
in the following subsections.


\paragraph{Proof of Lemma~\ref{lem:perm_to_mx}:}

The proof of this lemma involves a slight generalization of the proof
of the upper bound in Theorem~\ref{ThmMinimax} (see
Section~\ref{AppThmMinimaxUpper} for this proof).  From the optimality
of $\wthat_\perm$ and feasibility of $\perm(\wtstar)$ for the
constrained least-squares program~\eqref{EqnConstrainedLSTwo}, we are
guaranteed that $\frobnorm{\obs - \wthat_\perm}^2 \leq \frobnorm{\obs
  - \perm(\wtstar)}^2$.  Introducing the error matrix \mbox{$\DelHatPi
  \defn \wthat_\perm - \wtstar$,} some algebraic manipulations yield
the modified basic inequality
\begin{align*}
\frobnorm{\DelHatPi}^2 \leq \frobnorm{\wtstar -
  \perm(\wtstar)}^2 + 2 \tracer{\noise}{ \wthat_\perm - \perm(\wtstar) }.
\end{align*}
Let us define $\DelHat \defn \wthat_\perm - \perm(\wtstar) $. Further, for each choice of radius $t > 0$, recall the definitions of the
random variable $Z(t)$ and event $\AuxEvent$ from
equations~\eqref{EqnDefnZ} and~\eqref{EqnDefnBadevent}, respectively.
With these definitions, we have the upper bound
\begin{align}
\label{EqnBasicHighSNR}
\frobnorm{\DelHatPi}^2 \leq \frobnorm{\wtstar -
  \perm(\wtstar)}^2 + 2 Z \big( \frobnorm{\DelHat} \big).
\end{align}
Lemma~\ref{LemChattMetent} proved earlier shows that the inequality
$\Exs[Z(\delcrit)] \leq \frac{\delcrit^2}{2}$ is satisfied by
$\delcrit = \plaincon \sqrt{\numitems} \log \numitems$. In a manner
identical to the proof in Section~\ref{AppThmMinimaxUpper}, one can
show that
\begin{align*}
\mprob[\AuxEvent] & \leq \mprob[Z(\delcrit) \geq 2 \delcrit \sqrt{t
    \delcrit} \big] \; \leq \; 2 e^{-c_1 t \delcrit} \quad \mbox{for
  all $t \geq \delcrit$.}
\end{align*}
Given these results, we break the next step into two cases depending on the magnitude of $\DelHat$.
\noindent \underline{Case I:} Suppose $\frobnorm{\DelHat} \leq \sqrt{t \delcrit}$. In this case, we have
\begin{align*}
\frobnorm{\DelHatPi}^2 \leq 2\frobnorm{\wtstar - \perm(\wtstar)}^2 + 2\frobnorm{\DelHat}^2\\
\leq 2\frobnorm{\wtstar - \perm(\wtstar)}^2 + t \delcrit.
\end{align*}

\noindent \underline{Case II:} Otherwise, we must have $\frobnorm{\DelHat} > \sqrt{t \delcrit}$. Conditioning on the complement $\AuxEvent^c$, our basic
inequality~\eqref{EqnBasicHighSNR} implies that
\begin{align*}
\frobnorm{\DelHatPi}^2 & \leq \frobnorm{\wtstar -
  \perm(\wtstar)}^2 + 4 \frobnorm{\DelHat} \sqrt{t \delcrit} \\ 
& \leq \frobnorm{\wtstar - \perm(\wtstar)}^2 +
\frac{\frobnorm{\DelHat}^2}{8} + 32 t \delcrit,\\
& \leq \frobnorm{\wtstar - \perm(\wtstar)}^2 +
\frac{2\frobnorm{\DelHatPi}^2 + 2\frobnorm{\wtstar - \perm(\wtstar)}^2}{8} + 32 t \delcrit,
\end{align*}
with probability at least $1 - 2 e^{-c_1 t \delcrit}$.  

Finally, setting $t = \delcrit = 
\plaincon \sqrt{\numitem} \log(\numitem)$ in either case and re-arranging yields the
bound~\eqref{EqnApproxEst}.


\paragraph{Proof of Lemma~\ref{lem:FAS}:}

For any matrix $\wt$ and any value $i$, let $\wt_i$ denote its $i^{\rm
  th}$ row.  Also define the clipping function $b: \mathbb{Z}
\rightarrow [\numitems]$ via $b(x) = \min \{ \max \{1, x \}, n \}$.
Using this notation, we have
\begin{align*}
\frobnorm{\wtstar - \perm(\wtstar)}^2 &= \sum_{i=1}^{\numitems}
\|\wtstar_i - \wtstar_{\perm^{-1}(i)}\|_2^2 \\
& \leq \;
\sum_{i=1}^{\numitems} \max_{0 \leq j \leq \BRACON \log \numitems}
\{ \|\wtstar_i - \wtstar_{b(i - j)}\|_2^2, \|\wtstar_i - \wtstar_{b(i
  + j)}\|_2^2 \},
\end{align*}
where we have used the definition of the set $\permhatclass(\BRACON)$ to obtain the final inequality. Since $\wtstar$
corresponds to the identity permutation, we have $\wtstar_1 \geq
\wtstar_2 \geq \cdots \geq \wtstar_\numitems$, where the inequalities
are in the pointwise sense. Consequently, we have
\begin{align*}
\frobnorm{\wtstar - \perm(\wtstar)}^2 & \leq \sum_{i=1}^{\numitems}
\max \Big \{ \|\wtstar_i - \wtstar_{b(i - \BRACON \log
  \numitems)}\|_2^2, \|\wtstar_i - \wtstar_{b(i + \BRACON \log
  \numitems)}\|_2^2 \Big \} \\
&\leq 2 \sum_{i=1}^{\numitems - \BRACON \log \numitems}
\|\wtstar_i - \wtstar_{i + \BRACON \log \numitems}\|_2^2.
\end{align*}
One can verify that the inequality \mbox{$\sum_{i=1}^{k-1} (a_i -
  a_{i+1})^2 \leq (a_1 - a_k)^2$} holds for all ordered sequences of
real numbers $a_1 \geq a_2 \geq \cdots \geq a_k$.  As stated earlier,
the rows of $\wtstar$ dominate each other pointwise, and hence we
conclude that
\begin{align*}
\frobnorm{\wtstar - \perm(\wtstar)}^2 & \leq 2 \BRACON \log \numitems
\|\wtstar_1 - \wtstar_{\numitems}\|_2^2 \; \leq \; 2 \BRACON	
\numitems \log \numitems,
\end{align*}
which establishes the claim~\eqref{EqnApproxBound}.


\subsection{Proof of Theorem~\ref{thm:parametric}}
\label{AppThmParametric}

We now turn to our theorem giving upper and lower bounds on estimating
pairwise probability matrices for parametric models.  Let us begin
with a proof of the claimed lower bound.


\subsubsection{Lower bound}

We prove our lower bound by constructing a set of matrices that are
well-separated in Frobenius norm.  Using this set, we then use an
argument based on Fano's inequality to lower bound the minimax risk.
Underlying our construction of the matrix collection is a collection
of Boolean vectors.  For any two Boolean vectors $\boo, \boo' \in
\{0,1\}^\numitems$, let $\dham(\boo, \boo') = \sum_{j=1}^\numitems
\Ind[\boo_j \neq \boo'_j]$ denote the Hamming distance between them.

\begin{lemma}
\label{LemBoolean}
For any fixed $\packdmin \in (0,1/4)$, there is a collection of
Boolean vectors $\{\boo^1, \ldots, \boo^\packnum \}$ such that
\begin{subequations}
\begin{align}
\min \big \{ \dham(\boo^j, \boo^k), \dham(\boo^j, 0) \big \} & \geq
\lceil \packdmin \numitems \rceil \qquad \mbox{for all distinct $j
  \neq k \in \{1, \ldots, \packnum\}$, and} \\
\packnum \equiv \packnum(\packdmin) & \geq \exp\Big \{ (\numitems-1)
\, \kl{2\alpha}{\frac{1}{2}} \Big \} - 1.
\end{align}
\end{subequations}
\end{lemma}

Given the collection $\{\boo^j, j \in [\packnum(\packdmin)]\}$
guaranteed by this lemma, we then define the collection of real
vectors $\{\wtparam^j, j \in [\packnum(\packdmin)] \}$ via
\begin{align*}
\wtparam^j = \delta \Big( \identity - \frac{1}{\numitems} \ones
\ones^T \Big) \boo^j \qquad \mbox{for each $j \in
  [\packnum(\packdmin)]$},
\end{align*}
where $\delta \in (0, \wmax)$ is a parameter to be specified later in
the proof. By construction, for each index $j \in
[\packnum(\packdmin)]$, we have $\inprod{\ones}{\wtparam^j} = 0$ and
$\|\wtparam^j\|_\infty \leq \delta$.  Based on these vectors, we then
define the collection of matrices $ \big\{ \wt^j, j \in
[\packnum(\packdmin)] \big \}$ via
\begin{align*}
[\wt^k]_{ij} \defn \glmcdf([\wtparam^k]_i - [\wtparam^k]_j).
\end{align*}
By construction, this collection of matrices is contained within our
parametric family.  We also claim that they are well-separated in
Frobenius norm:
\begin{lemma}
\label{LemFrobLower}
For any distinct pair $j, k \in [\packnum(\packdmin)]$, we have
\begin{align}
\label{EqnFrobLower}
\frac{\frobnorm{\wt^j - \wt^k}^2}{\numobs^2} & \geq
\frac{\packdmin^2}{4} (\glmcdf(\delta)-\glmcdf(0))^2.
\end{align}
\end{lemma}

In order to apply Fano's inequality, our second requirement is an
upper bound on the mutual information $I(Y; J)$, where $J$ is a random
index uniformly distributed over the index set $[\packnum] = \{1,
\ldots, \packnum\}$.  By Jensen's inequality, we have $I(Y; J) \leq
\frac{1}{{\packnum \choose 2}} \sum_{j \neq k}
\kl{\mprob^j}{\mprob^k}$, where $\mprob^j$ denotes the distribution of
$Y$ when the true underlying matrix is $\wt^j$.  Let us upper bound
these KL divergences.

For any pair of distinct indices $u,v \in [\numitems]^2$, let
$\diff_{uv}$ be a differencing vector---that is, a vector whose
components $u$ and $v$ are set as $1$ and $-1$, respectively,
with all remaining components equal to $0$.  We are then guaranteed
that
\begin{align*}
\inprod{\diff_{uv}}{\wtparam^j} = \delta \inprod{\diff_{uv}}{\boo^j},
\quad \mbox{and} \quad \glmcdf( \inprod{\diff_{uv}}{\wtparam^j}) \in
\big \{\glmcdf(-\delta), \glmcdf(0), \glmcdf(\delta) \big\},
\end{align*}
where $\glmcdf(\delta) \geq \glmcdf(0) \geq \glmcdf(-\delta)$ by
construction.  Using these facts, we have
\begin{align}
\kl{\mprob^j}{\mprob^k} & \stackrel{(i)}{\leq} 2 \sum_{u, v \in [\numitems]}
\frac{ \big(\glmcdf( \inprod{\diff_{uv}}{\wtparam^j})-\glmcdf(
  \inprod{\diff_{uv}}{\wtparam^k}) \big)^2} {\min\{\glmcdf(
  \inprod{\diff_{uv}}{\wtparam^k}), 1 - \glmcdf( \inprod{\diff_{uv}}
         {\wtparam^k})\}} \notag \\
& \leq 2 \numitems^2 \frac{(\glmcdf(\delta) -
           \glmcdf(-\delta))^2}{\glmcdf(-\delta)} \notag \\
\label{EqnManor}
& \leq 8 \numitems^2 \frac{( \glmcdf(\delta) - \glmcdf(0))^2}{
  \glmcdf(-\delta)},
\end{align}
where the bound $(i)$ follows from the elementary inequality $a \log \frac{a}{b} \leq (a-b) \frac{a}{b}$ for any two numbers $a, b \in (0,1)$.

This upper bound on the KL divergence~\eqref{EqnManor} and lower bound
on the Frobenius norm~\eqref{EqnFrobLower}, when combined with Fano's
inequality, imply that any estimator $\wthat$ has its worst-case risk
over our family lower bounded as
\begin{align*}
\sup_{j \in [\packnum(\packdmin)]} \frac{1}{\numitems^2} \Exs \big[
  \frobnorm{\wthat - \wt(\wtparam^j)}^2 \big] & \geq \frac{1}{8}
\packdmin^2 (\glmcdf(\delta) - \glmcdf(0))^2 \Big(1 -
\frac{\frac{8}{\glmcdf(-\delta)} \numitems^2 (\glmcdf(\delta) -
  \glmcdf(0))^2 + \log 2}{\numitems} \Big).
\end{align*}
Choosing a value of $\delta > 0$ such that $(\glmcdf(\delta)-\glmcdf(0))^2
= \frac{\glmcdf(-\delta)}{80\numitems}$ gives the claimed
result. (Such a value of $\delta$ is guaranteed to exist with $F(-\delta) \in [\frac{1}{4}, \frac{1}{2}]$ given our
assumption that $\glmcdf$ is continuous and strictly increasing.) \\

\noindent The only remaining details are to prove
Lemmas~\ref{LemBoolean} and~\ref{LemFrobLower}.

\paragraph{Proof of Lemma~\ref{LemBoolean}:}
The Gilbert-Varshamov
bound~\cite{gilbert1952comparison,varshamov1957estimate} guarantees
the existence of a collection of vectors $\{\boo^0, \ldots,
\boo^{\GVCARD-1}\}$ contained with the Boolean hypercube
$\{0,1\}^{\numitems}$ such that 
\begin{align*}
\GVCARD & \geq 2^{\numitems-1} \; \big(\sum_{\ell=0}^{\lceil \packdmin
	\numitems \rceil-1}{\numitems-1 \choose \ell} \big)^{-1}, \qquad
\mbox{and} \\ \dham(\boo^j, \boo^k) & \geq \lceil \packdmin \numitems
\rceil \qquad \mbox{for all $j \neq k$, $j, k \in [\GVCARD-1]$}.
\end{align*}
Moreover, their construction allows loss of generality that the
all-zeros vector is a member of the set---say $\boo^0 = 0$.  We are
thus guaranteed that $\dham(\boo^j, 0) \geq \lceil \packdmin \numitems
\rceil$ for all $j \in \{1, \ldots, \GVCARD-1\}$.

Since $\numitems \geq 2$ and $\alpha \in (0,\frac{1}{4})$, we have
$\frac{\lceil \alpha \numitems \rceil - 1}{\numitems - 1} \leq 2\alpha
\leq \frac{1}{2}$.  Applying standard bounds on the tail of the
binomial distribution yields
\begin{align*}
\frac{1}{2^{\numitems-1}} \sum_{\ell=0}^{\lceil \packdmin\numitems
  \rceil -1}{\numitems-1 \choose \ell } \leq \exp \Big(-(\numitems-1)
\kl{\frac{\lceil \alpha \numitems \rceil - 1}{\numitems -
    1}}{\frac{1}{2}} \Big) \leq \exp \Big(-(\numitems-1)
\kl{2\alpha}{\frac{1}{2}} \Big).
\end{align*}
Consequently, the number of non-zero code words $\packnum \defn
\GVCARD -1$ is at least
\begin{align*}
\packnum(\packdmin) \defn \exp \Big((\numitems-1)
\kl{2\alpha}{\frac{1}{2}} \Big)-1.
\end{align*}
Thus, the collection $\{\boo^1, \ldots, \boo^\packnum \}$ has the
desired properties.


\paragraph{Proof of Lemma~\ref{LemFrobLower}:}

By definition of the matrix ensemble, we have
\begin{align}
\label{EqnStarbucks}
\frobnorm{\wt(\wtparam^j)-\wt(\wtparam^k)}^2 = & \sum_{u, v \in
  [\numitems]} (\glmcdf( \inprod{\diff_{uv}}{ \wtparam^j})-\glmcdf(
\inprod{\diff_{uv}}{\wtparam^k}))^2.
\end{align}
By construction, the Hamming distances between the triplet of vectors
$\{\wtparam^j, \wtparam^k, 0 \}$ are lower bounded
$\hamming(\wtparam^j,0) \geq \packdmin \numitems$,
$\hamming(\wtparam^k,0) \geq \packdmin \numitems$ and
$\hamming(\wtparam^j,\wtparam^k) \geq \packdmin \numitems$.  We claim
that this implies that
\begin{align}
\label{EqnHanaSick}
\operatorname{card} \Big \{ u \neq v \in [\numitems]^2 \, \mid
\inprod{\diff_{uv}}{\wtparam^j} \neq
\inprod{\diff_{uv}}{\wtparam^k}\Big \} & \geq \frac{\packdmin^2}{4} \,
\numitems^2.
\end{align}
Taking this auxiliary claim as given for the moment, applying it to
Equation~\eqref{EqnStarbucks} yields the lower bound
$\frobnorm{\wt(\wtparam^1)-\wt(\wtparam^2)}^2 \; \geq \; \frac{1}{4}
\packdmin^2 \numitems^2 (\glmcdf(\delta)-\glmcdf(0))^2$, as claimed. \\

It remains to prove the auxiliary claim~\eqref{EqnHanaSick}.  We
relabel $j = 1$ and $k = 2$ for simplicity in notation. For $(y,z) \in
\{0,1\} \times \{0,1\}$, let set ${\cal I}_{yz} \subseteq [\numitems]$
denote the set of indices on which $\wtparam^1$ takes value $y$ and
$\wtparam^2$ takes value $z$.  We then split the proof into two cases:
\\

\noindent \underline{Case 1:} Suppose $\mid {\cal I}_{00} \cup {\cal
  I}_{11} \mid \geq \frac{\packdmin \numitems}{2}$. The minimum
distance condition $\hamming(\wtparam^1,\wtparam^2) \geq \packdmin
\numitems$ implies that $\mid {\cal I}_{01} \cup {\cal I}_{10} \mid
\geq \packdmin \numitems$. For any $i \in {\cal I}_{00} \cup {\cal
  I}_{11} $ and any $j \in {\cal I}_{01} \cup {\cal I}_{10} $, it must
be that $\inprod{\diff_{uv}}{\wtparam^1} \neq
\inprod{\diff_{uv}}{\wtparam^2}$. Thus there are at least
$\frac{\packdmin^2}{2} \numitems^2$ such pairs of indices.\\

\noindent \underline{Case 2:} Otherwise, we may assume that $\mid
          {\cal I}_{00} \cup {\cal I}_{11} \mid< \frac{\packdmin
            \numitems}{2}$. This condition, along with the minimum
          Hamming weight conditions $\hamming(\wtparam^1,0) \geq
          \packdmin \numitems$ and $\hamming(\wtparam^2,0) \geq
          \packdmin \numitems$, gives ${\cal I}_{10} \geq
          \frac{\packdmin \numitems}{2}$ and ${\cal I}_{01} \geq
          \frac{\packdmin \numitems}{2}$. For any $i \in {\cal
            I}_{01}$ and any $j \in {\cal I}_{10} $, it must be that
          $\inprod{\diff_{uv}}{\wtparam^1} \neq
          \inprod{\diff_{uv}}{\wtparam^2}$. Thus there are at least
          $\frac{\packdmin^2}{4} \numitems^2$ such pairs of indices.


\subsubsection{Upper bound}
\label{SecProofParametricFull}

In our earlier work~\cite[Theorem 2b]{shah2015estimation} we prove that when $\glmcdf$ is
strongly log-concave and twice differentiable, then there is a
universal constant $\UUP$ such that the maximum likelihood estimator
$\wtparamhatML$ has mean squared error at most
\begin{align}
\label{eq:param_upper_0}
\sup_{\wtparamstar \in [-\wmax,\wmax]^\numitems,
  \inprod{\wtparamstar}{1}=0} \Exs [ \myLnorm{\wtparamhatML -
    \wtparamstar}{2}^2 ] & \leq \UUP.
\end{align}
Moreover, given the log-concavity assumption, the MLE is computable in
polynomial-time.  Let $\wt(\wtparamhatML)$ and $\wt(\wtparamstar)$
denote the pairwise comparison matrices induced, via
Equation~\eqref{EqnInduce}, by $\wtparamhatML$ and $\wtparamstar$.  It
suffices to bound the Frobenius norm $\frobnorm{\wt(\wtparamhatML) -
  \wt(\wtparamstar)}$.

Consider any pair of vectors $\wtparam^1$ and $\wtparam^2$ that lie in
the hypercube $[-1,1]^\numitems$.  For any pair of indices $(i,j) \in
[\numitems]^2$, we have
\begin{align*}
((\wt(\wtparam^1))_{ij}-(\wt(\wtparam^2))_{ij})^2 &=
  (\glmcdf(\wtparam^1_i - \wtparam^1_j)-\glmcdf(\wtparam^2_i -
  \wtparam^2_j))^2 \; \leq \; \zeta^2 ((\wtparam^1_i - \wtparam^1_j )
  - (\wtparam^2_i - \wtparam^2_j))^2,
\end{align*}
where we have defined $\zeta \defn \max \limits_{z \in [-1,1]}
\glmcdf'(z)$.  Putting together the pieces yields
\begin{align}
\label{EqnParamVecToMx}
\frobnorm{\wt(\wtparam^1)-\wt(\wtparam^2)}^2 & \leq \zeta^2
(\wtparam^1 - \wtparam^2)^T (\numitems \identity - \ones \ones^T)
(\wtparam^1 - \wtparam^2) \; = \; \numitems \zeta^2 \myLnorm{\wtparam^1
  - \wtparam^2}{2}^2.
\end{align}
Applying this bound with $\wtparam^1 = \wtparamhatML$ and $\wtparam^2
= \wtparamstar$ and combining with the bound~\eqref{eq:param_upper_0}
yields the claim.

\subsection{Proof of Theorem~\ref{ThmPartialObservations}}
\label{AppThmPartialObservations}

We now turn to the proof of Theorem~\ref{ThmPartialObservations},
which characterizes the behavior of different estimators for the
partially observed case.

\subsubsection{Proof of part (a)}

In this section, we prove the lower and upper bounds stated in part
(a).

\paragraph{Proof of lower bound:}
We begin by proving the lower bound in
equation~\eqref{EqnPartialMinimax}.  The Gilbert-Varshamov
bound~\cite{gilbert1952comparison,varshamov1957estimate} guarantees
the existence of a set of vectors $\{\boo^1, \ldots, \boo^\packnum\}$
in the Boolean cube $\{0,1\}^\frac{\numitems}{2}$ with cardinality at
least $\packnum \defn 2^{\plaincon \numitems}$ such that
\begin{align*}
\dham(\boo^j, \boo^k) & \geq \lceil 0.1 \numitems \rceil \qquad
\mbox{for all distinct pairs $j, k \in [\packnum] \defn \{1, \ldots,
  \packnum\}$.}
\end{align*}
Fixing some $\delta \in (0,\frac{1}{4})$ whose value is to be specified
later, for each $k \in [\packnum]$, we define a matrix $\wt^k \in
\chatterjeeclass$ with entries
\begin{align*}
[\wt^k]_{uv} =
\begin{cases} \half + \delta & \quad \mbox{if } u \leq \frac{\numitems}{2},
      [\boo^k]_{u} = 1 \mbox{ and } v \geq \frac{\numitems}{2} \\
 \half & \quad \mbox{otherwise},
\end{cases} 
\end{align*}
for every pair of indices $u \leq v$.  We complete the matrix by
setting $[\wt^k]_{vu} = 1 - [\wt^k]_{uv}$ for all indices $u > v$. 

By construction, for each distinct pair $j, k \in [\packnum]$, we have
the lower bound
\begin{align*}
\frobnorm{\wt^j - \wt^k}^2 &= \numitems\delta^2 \|\boo^j - \boo^k\|_2^2
\; \geq \; \plaincon_0 \numitems^2 \delta^2.
\end{align*}
Let $\mprob^j$ and $\mprob^j_{uv}$ denote (respectively) the
distributions of the matrix $Y$ and entry $Y_{uv}$ when the underlying
matrix is $\wt^j$.  Since the entries of $Y$ are generated
independently, we have $\kl{\mprob^j}{\mprob^k} = \sum
\limits_{\HACKUV} \kl{\mprob^j_{uv}}{\mprob^k_{uv}}$. The matrix entry
$Y_{uv}$ is generated according to the model
\begin{align*}
\obs_{uv} = \begin{cases} 1 &\quad \mbox{w.p. }\pp\wtstar_{uv}\\ 0 &
  \quad \mbox{w.p. }\pp(1-\wtstar_{uv})\\\mbox{not observed} & \quad
  \mbox{w.p. }1-\pp.
\end{cases} 
\end{align*}
Consequently, the KL divergence can be upper bounded as
\begin{subequations}
\begin{align}
& \kl{\mprob^j_{uv}}{\mprob^k_{uv}} \nonumber \\
& = \pp \Big( \wt^j_{uv} \log \frac{
	\wt^j_{uv}}{ \wt^k_{uv}} + (1-\wt^j_{uv}) \log \frac{(1-
	\wt^j_{uv})}{(1-\wt^k_{uv})} \Big) \nonumber \\
& {\leq} \pp \Big\{ \! \wt^j_{uv} \big( \frac{
	\wt^j_{uv} - \wt^k_{uv}}{ \wt^k_{uv}}\big) + (1-\wt^j_{uv}) \big(
\frac{\wt^k_{uv}- \wt^j_{uv}}{1-\wt^k_{uv}} \big) \! \Big\}\label{EqnKLFrob2} \\
& = \pp \; \frac{( \wt^j_{uv} - \wt^k_{uv}
	)^2}{\wt^k_{uv}(1-\wt^k_{uv})}\label{EqnKLFrob3} \\
& {\leq} 16 \pp \; ( \wt^j_{uv} - \wt^k_{uv} )^2,
\label{EqnKLFrob4}
\end{align}
\end{subequations}
where inequality~\eqref{EqnKLFrob2} follows from the fact that $\log(t) \leq t - 1$
for all $t > 0$; and inequality~\eqref{EqnKLFrob4} follows 
since the numbers
$\{\wt^j_{uv}, \wt^k_{uv}\}$ both lie in the interval $[\frac{1}{4},
\frac{3}{4}]$. Putting together the pieces, we conclude that
\begin{align*}
\kl{\mprob^j}{\mprob^k} \leq \plaincon_1 \pp \frobnorm{\wt^j -
  \wt^k}^2 \; \leq \plaincon_1' \pp \numitems^2 \delta^2.
\end{align*}
Thus, applying Fano's inequality to the packing set
$\{\wt^1,\ldots,\wt^\packnum\}$ yields that any estimator $\wthat$ has
mean squared error lower bounded by
\begin{align*}
\sup_{k \in [\packnum]} \frac{1}{\numitems^2} \Exs[\frobnorm{\wthat -
  \wt^k}^2 ] \geq \plaincon_0 \delta^2 \Big(1 - \frac{ \plaincon_1' \pp
  \numitems^2 \delta^2 + \log 2}{\plaincon \numitems} \Big).
\end{align*}
Finally, choosing $\delta^2 = \frac{\plaincon_2 }{2\plaincon_1 \pp
  \numitems}$ yields the lower bound $\sup_{k \in [\packnum]}
\frac{1}{\numitems^2} \Exs[ \frobnorm{\wthat - \wt^k}^2 ] \geq \plaincon_3
\frac{1}{\numitems \pp}$.  Note that in order to satisfy the condition
$\delta \leq \frac{1}{4}$, we must have $\pp \geq \frac{16 \plaincon_2
}{2\plaincon_1 \numitems}$. 
 

\paragraph{Proof of upper bound:}

For this proof, recall the linearized form of the
observation model given in
equations~\eqref{EqnDefnObsPartial},~\eqref{EqnDefnObsPrime}, and~\eqref{EqnDefnWprimePartial}.  We begin by introducing some
additional notation. Letting $\permall$ denote the set of all
permutations of $\numitems$ items. For each $\perm \in \permall$, we
define the set
\begin{align*}
\perm(\bisoclass) \defn \big \{ M \in [0,1]^{\numitems \times
  \numitems} \mid \mbox{$M_{k \ell} \geq M_{ij}$ whenever $\perm(k)
  \leq \perm(i)$ and $\perm(\ell) \geq \perm(j)$} \big \},
\end{align*}
corresponding to the subset of SST matrices that are faithful to the
permutation $\perm$.  We then define the estimator $\wt_\perm \in
\argmin \limits_{\wt \in \perm(\bisoclass)} \frobnorm{\obs' - \wt}^2$,
in terms of which the least squares
estimator~\eqref{EqnDefnLSEPartial} can be rewritten as
\begin{align*}
\wthat \in \argmin \limits_{\perm \in \permall} \frobnorm{\obs' -
  \wt_\perm}^2.
\end{align*}
Define a set of permutations $\permall' \subseteq \permall$ as
\begin{align*}
\permall' \defn \{\perm \in \permall \mid \frobnorm{ \obs' -
  \wt_\perm}^2 \leq \frobnorm{ \obs' - \wtstar }^2\}.
\end{align*}
Note
that the set $\permall'$ is guaranteed to be non-empty since the
permutation corresponding to $\wthat$ always lies in $\permall'$.
 We claim that for any $\perm \in \permall'$,
we have
\begin{align} 
\label{EqnPartialPermReq}
\mprob \big( \frobnorm{\wt_\perm - \wtstar}^2 \leq \UUP
\frac{\numitems}{\pp} \log^2 \numitems \big) \geq 1 - e^{- 3 \numitems
  \log \numitems},
\end{align}
for some positive universal constant $\UUP$. Given this bound, since
there are at most $e^{\numitems \log \numitems}$ permutations in the
set $\permall'$, a union bound over all these permutations applied
to~\eqref{EqnPartialPermReq} yields
\begin{align*}
\mprob \Big( \max_{\perm \in \permall'} \frobnorm{\wt_\perm -
  \wtstar}^2 > \UUP \frac{\numitems}{\pp} \log^2 \numitems \Big) \leq
e^{-2 \numitems \log \numitems}.
\end{align*}
Since $\wthat$ is equal to $\wt_{\perm}$ for some $\perm \in \permall'$, this tail bound yields the claimed
result.

The remainder of our proof is devoted to proving the
bound~\eqref{EqnPartialPermReq}. By definition, any permutation $\perm
\in \permall'$ must satisfy the inequality
\begin{align*}
\frobnorm{ \obs - \wt_\perm}^2 \leq \frobnorm{ \obs - \wtstar }^2.
\end{align*}
Letting $\DelHat_\perm \defn \wt_\perm - \wtstar$ denote the error
matrix, and using the linearized form~\eqref{EqnDefnObsPrime} of the
observation model, some algebraic manipulations yield the basic
inequality
\begin{align}
\label{EqnBasicPartial}
\frac{1}{2} \frobnorm{\DelHat_\perm}^2 & \leq \frac{1}{\pp}
\tracer{W'}{\DelHat_\perm}.
\end{align}
Now consider the set of matrices
\begin{align}
\label{eq:EqnDefnChattDiffPartial}
\chattDiff(\perm) \defn \Big \{ \alpha(\wt - \wtstar) \mid & \ \wt \in
\perm(\bisoclass), \ \alpha \in [0,1] \Big \},
\end{align}
and note that $\chattDiff(\perm) \subseteq [-1,1]^{\numitems \times
  \numitems}$.  (To be clear, the set $\chattDiff(\perm)$ also depends
on the value of $\wtstar$, but considering $\wtstar$ as fixed, we omit
this dependence from the notation for brevity.)  For each choice of
radius $t > 0$, define the random variable
\begin{align}
\label{EqnDefnZPartial}
Z_\perm(t) & \defn \sup_{ \substack{\chattDiffmx \in
    \chattDiff(\perm),\\ \frobnorm{\chattDiffmx} \leq t}} ~~
\frac{1}{\pp} \tracer{\chattDiffmx}{\Wmat'}.
\end{align}
Using the basic inequality~\eqref{EqnBasicPartial}, the Frobenius norm
error $\frobnorm{\DelHat_\perm}$ then satisfies the bound
\begin{align}
\label{EqnDefnDelcritPartial}
\frac{1}{2} \frobnorm{\DelHat_\perm}^2 & \leq \frac{1}{\pp}
\tracer{\Wmat'}{\DelHat_\perm} \; \leq \; Z_\perm \big(
\frobnorm{\DelHat_\perm} \big).
\end{align}
Thus, in order to obtain a high probability bound, we need to
understand the behavior of the random quantity $Z_\perm(t)$.

One can verify that the set $\chattDiff(\perm)$ is star-shaped,
meaning that $\alpha \chattDiffmx \in \chattDiff(\perm)$ for every
$\alpha \in [0,1]$ and every $\chattDiffmx \in
\chattDiff(\perm)$. Using this star-shaped property, we are guaranteed
that there is a non-empty set of scalars $\delcrit > 0$ satisfying the
critical inequality
\begin{align}
\label{EqnCriticalPartial}
\Exs[Z_\perm(\delcrit)] & \leq \frac{\delcrit^2}{2}.
\end{align}
Our interest is in an upper bound to the smallest (strictly) positive
solution $\delcrit$ to the critical
inequality~\eqref{EqnCriticalPartial}, and moreover, our goal is to
show that for every $t \geq \delcrit$, we have $\frobnorm{\DelHat}
\leq c \sqrt{t \delcrit}$ with high probability.

For each $t > 0$, define the ``bad'' event
\begin{align}
\label{EqnDefnBadeventPartial}
\AuxEvent & = \big \{ \exists \Delta \in \chattDiff(\perm) \mid
\frobnorm{\Delta} \geq \sqrt{t \delcrit} \quad \mbox{and} \quad
\frac{1}{\pp} \tracer{\Delta}{\Wmat'} \geq 2 \frobnorm{\Delta} \sqrt{t
  \delcrit} \big \}.
\end{align}
Using the star-shaped property of $\chattDiff(\perm)$, it follows by a
rescaling argument that
\begin{align*}
\mprob[\AuxEvent] \leq \mprob[Z_\perm(\delcrit) \geq 2 \delcrit
  \sqrt{t \delcrit}] \qquad \mbox{for all $t \geq \delcrit$.}
\end{align*}
The following lemma helps control the behavior of the random variable
$Z_\perm(\delcrit)$.
\begin{lemma}
\label{LemZpermPartial}
For any $\delta >0$, the mean of $Z_\perm(\delta)$ is bounded as
\begin{align*}
\Exs [ Z_\perm(\delta) ] \leq \UUP \frac{\numitems}{\pp} \log^2
\numitems,
\end{align*}
and for every $u>0$, its tail probability is bounded as
\begin{align*}
\mprob \Big(Z_\perm(\delta) > \Exs[Z_\perm(\delta)] + u \Big) \leq \exp\Big( \frac{-\UHP u^2 \pp}{\delta^2 + \Exs[Z_\perm(\delta)] + u} \Big),
\end{align*}
where $\UUP$ and $\UHP$ are positive universal constants.
\end{lemma}
From this lemma, we have the tail bound
\begin{align*}
\mprob \Big(Z_\perm(\delcrit) > \Exs[Z_\perm(\delcrit)] + \delcrit
\sqrt{t \delcrit} \Big) \leq \exp\Big( \frac{-\UHP (\delcrit \sqrt{t \delcrit})^2 \pp}{ \delcrit^2 + \Exs[Z_\perm(\delcrit)] +
  (\delcrit \sqrt{t \delcrit})} \Big), \quad \mbox{for all $t \geq
  \delcrit$.}
\end{align*}
By the definition of $\delcrit$ in equation~\eqref{EqnCriticalPartial}, we have
$\Exs[Z(\delcrit)] \leq \delcrit^2 \leq \delcrit \sqrt{t \delcrit}$
for any $t \geq \delcrit$, and consequently
\begin{align*}
\mprob[\AuxEvent] & \leq \mprob[Z(\delcrit) \geq 2 \delcrit \sqrt{t
    \delcrit} \big] \; \leq \; \exp\Big( \frac{-\UHP (\delcrit \sqrt{t \delcrit})^2 \pp }{3\delcrit \sqrt{t \delcrit} } \Big), \quad \mbox{for all $t \geq \delcrit$.}
\end{align*}
Consequently, either $\frobnorm{\DelHat_\perm} \leq \sqrt{t
  \delcrit}$, or we have $\frobnorm{\DelHat_\perm} > \sqrt{t
  \delcrit}$. In the latter case, conditioning on the complement
$\AuxEvent^c$, our basic inequality implies that $\frac{1}{2}
\frobnorm{\DelHat_\perm}^2 \leq 2 \frobnorm{\DelHat_\perm} \sqrt{t
  \delcrit}$ and hence $\frobnorm{\DelHat_\perm} \leq 4 \sqrt{t
  \delcrit}$.  Putting together the pieces yields that
\begin{align}
\mprob \big( \frobnorm{\DelHat_\perm} \leq 4 \sqrt{t \delcrit} \big)
\geq 1 - \exp \big( -\UHP' \delcrit \sqrt{t \delcrit} \pp \big),  \quad \mbox{for all $t \geq \delcrit$.}
\label{EqWithDelCritHPPartial}
\end{align}

Finally, from the bound on the expected value of $Z_\perm(t)$ in
Lemma~\ref{LemZpermPartial}, we see that the critical
inequality~\eqref{EqnCriticalPartial} is satisfied for $\delcrit =
\sqrt{\frac{\UUP \numitems}{\pp}}\log \numitems$. Setting $t =
\delcrit = \sqrt{\frac{\UUP \numitems}{\pp}}\log \numitems$
in~\eqref{EqWithDelCritHPPartial} yields
\begin{align}
\mprob \Big( \frobnorm{\DelHat_\perm} \leq 4 \frac{\UUP
  \numitems}{\pp} \log^2 \numitems \Big) \geq 1 - \exp\Big( -3 \numitems \log \numitems \Big),
\end{align}
for some universal constant $\UUP>0$, thus proving the bound~\eqref{EqnPartialPermReq}.\\

\noindent It remains to prove Lemma~\ref{LemZpermPartial}.


\paragraph*{Proof of Lemma~\ref{LemZpermPartial}}
\underline{Bounding $\Exs [Z_\perm(\delta)]$}: We establish an upper bound on $\Exs [
  Z_\perm(\delta)]$ by using Dudley's entropy integral, as well as
some auxiliary results on metric entropy. We use the notation
$\metent(\epsilon,\mathbb{C},\rho)$ to denote the $\epsilon$ metric
entropy of the class $\mathbb{C}$ in the metric $\rho$.  Introducing
the random variable $\widetilde{Z}_\perm \defn \sup \limits_{ \chattDiffmx \in
  \chattDiff(\perm)} \tracer{\chattDiffmx}{\Wmat'}$, note that we have
$\Exs[Z_\perm(\delta)] \leq \frac{1}{\pp}
\Exs[\widetilde{Z}_\perm]$. The truncated form of Dudley's entropy
integral inequality yields
\begin{align}
\label{EqnDudleyPartial}
\Exs[ \widetilde{Z}_\perm] & \leq \plaincon \; \Big \{ \numitems^{-8}
+ \int_{\frac{1}{2} \numitems^{-9}}^{2 \numitems} \sqrt{
  \metent(\epsilon,\chattDiff(\perm),\frobnorm{.})} d\epsilon \Big \},
\end{align}
where we have used the fact that the diameter of the set $\chattDiff(\perm)$ is at most $2 \numitems$ in the Frobenius norm.

From our earlier bound~\eqref{eq:biiso_metent}, we are
guaranteed that for each $\epsilon > 0$, the metric entropy is upper
bounded as
\begin{align*}
\metent \Big( \epsilon, \{\alpha \wt \mid \wt \in \bisoclass, \alpha
\in [0,1]\}, \frobnorm{\cdot} \Big) & \leq 8\frac{\numitems^2}{
  \epsilon^2} \big(\log \frac{\numitems}{ \epsilon} \big)^2.
\end{align*}
Consequently, we have
\begin{align*}
\metent(\epsilon,\chattDiff(\perm),\frobnorm{.}) \leq
16\frac{\numitems^2}{ \epsilon^2} \big(\log \frac{\numitems}{
  \epsilon} \big)^2.
\end{align*} 
Substituting this bound on the metric entropy of
$\chattDiff(\perm)$ and the inequality $\epsilon \geq \half
\numitems^{-9}$ into the Dudley bound~\eqref{EqnDudleyPartial} yields
\begin{align*}
\Exs [\widetilde{Z}_\perm ] & \leq \plaincon \numitems (\log
\numitems)^2.
\end{align*}
 The inequality $\Exs[Z_\perm(\delta)] \leq \frac{1}{\pp}
 \Exs[\widetilde{Z}_\perm]$ then yields the claimed result.~\\

\underline{Bounding the tail probability of $Z_\perm(\delta)$}: In order to establish the claimed tail bound, we
use a Bernstein-type bound on the supremum of empirical processes due
to Klein and Rio~\cite[Theorem 1.1c]{klein2005concentration}, which we
state in a simplified form here.
\begin{lemma}
\label{LemKleinRio}
Let $X \defn (X_1,\ldots,X_m)$ be any sequence of zero-mean,
independent random variables, each taking values in $[-1,1]$. Let
$\mathcal{V} \subset [-1,1]^m$ be any measurable set of $m$-length
vectors.  Then for any $u>0$, the supremum $X^\dagger = \sup_{v \in
  \mathcal{V} } \inprod{X}{v}$ satisfies the upper tail bound
\begin{align*}
\mprob \big( X^\dagger > \Exs[ X^\dagger] + u \big) \leq \exp
\Big(\frac{ - u^2 }{2\sup_{v \in \mathcal{V} } \Exs[\inprod{v}{X}^2] +
  4\Exs[ X^\dagger] + 3 u} \Big).
\end{align*}
\end{lemma}
We now invoke Lemma~\ref{LemKleinRio} with the choices
\mbox{$\mathcal{V} = \chattDiff(\perm) \cap \Ball(\delta)$,} \mbox{$m
	= (\numitems \times \numitems)$,} \mbox{$X = \noise'$,} and
\mbox{$X^\dagger = \pp Z_\perm(\delta)$.} The matrix $\noise'$ has
zero-mean entries belonging to the interval $[-1, +1]$, and are
independent on and above the diagonal (with the entries below
determined by the skew-symmetry condition).  Then we have
\mbox{$\Exs[X^\dagger] \leq \pp \Exs[Z_\perm(\delta) ]$} and
\mbox{$\Exs[\tracer{\chattDiffmx}{\noise'}^2] \leq 4 \pp
	\frobnorm{\chattDiffmx}^2 \leq 4 \pp \delta^2$} for every
\mbox{$\chattDiffmx \in \mathcal{V}$.} With these assignments, and
some algebraic manipulations, we obtain that for every $u > 0$,
\begin{align*}
\mprob \Big[ Z_\perm(\delta) > \Exs[Z_\perm(\delta)] + u \Big]
\leq \exp\Big( \frac{ - u^2 \pp }{8 \delta^2 +
	4\Exs[Z_\perm(\delta)] + 3u} \Big),
\end{align*}
as claimed.


\subsubsection{Proof of part (b)}
In order to prove the bound~\eqref{EqnPartialSVT}, we analyze the SVT
estimator $\Tregp(\obs')$ with the threshold $\regparnp =
3 \sqrt{\frac{\numitems}{\pp}}$.  Naturally then, our analysis is
similar to that of complete observations case from
Section~\ref{AppThmImprovedUSVT}.  Recall our formulation of the
problem in terms of the observation matrix $\obs'$ along with the
noise matrix $\noise'$ from
equations~\eqref{EqnDefnObsPartial},~\eqref{EqnDefnObsPrime}
and~\eqref{EqnDefnWprimePartial}. The result of Lemma~\ref{LemSTSVD}
continues to hold in this case of partial observations, translated to
this setting. In particular, if $\regparnp \geq \frac{1.01}{\pp} \opnorm{\noise'}$,
then
\begin{align*}
\frobnorm{\Tregp(\obs') - \wtstar}^2 & \leq \plaincon_1
\sum_{j=1}^\numitem \min \big \{ \regparnp^2, \sigma_j^2(\wtstar) \big
\},
\end{align*}
where
$\plaincon_1 > 0$ is a universal constant.

We now upper bound the operator norm of the noise matrix $\noise'$. 
Define a $(2\numitems \times 2\numitems)$ matrix
\begin{align*}
\noise'' = 
\begin{bmatrix}
0 & \noise' \\
(\noise')^T & 0
\end{bmatrix}.
\end{align*}
From equation~\eqref{EqnDefnWprimePartial} and the construction above, we have that the matrix $\noise''$ is symmetric, with mutually independent entries above the diagonal that have a mean of zero and are bounded in absolute value by $1$. Consequently, known results in random matrix theory (e.g., see~\cite[Theorem 2.3.21]{tao2012topics}) yield the bound $\opnorm{\noise''} \leq 2.01 \sqrt{2\numitems}$ with probability at least $1 - \numitems^{-\plaincon_2}$, for some universal constant  $\plaincon_2>1$. One can also verify that $\opnorm{\noise''} = \opnorm{\noise'}$, thereby yielding the bound
\begin{align*}
\mprob \Big[ \opnorm{\Wmat'} > 2.01 \sqrt{2 \numitem \pp}  \Big] & \leq
\numitems^{-\plaincon_2}.
\end{align*}
%
%

With our choice $\regparnp = 3 \sqrt{\frac{\numitems}{\pp}}$, the event $\{\regparnp \geq
\frac{1.01}{\pp} \opnorm{\Wmat'} \}$ holds with probability at least
$1- \numitems^{-\plaincon_2}$.  Conditioned on this event, the
approximation-theoretic result from Lemma~\ref{LemSTSVDChatterjee}
gives
\begin{align*}
\frac{1}{\numitems^2} \frobnorm{\Tregp(\obs') - \Mstar}^2 & \leq c
\Big( \frac{s \regparnp^2}{\numitems^2} + \frac{1}{s} \Big)
\end{align*}
with probability at least $1 - \numitems^{-\plaincon_2}$.  Substituting
$\regparnp = 3 \sqrt{\frac{\numitems}{\pp}}$ in this bound and
setting $s = \sqrt{\pp \numitems}$ yields the claimed result.


\subsubsection{Proof of part (c)}

As in our of proof of the fully observed case from
Section~\ref{SecProofParametricFull}, we consider the two-stage
estimator based on first computing the MLE $\wtparamhatML$ of
$\wtparamstar$ from the observed data, and then constructing the
matrix estimate $\wt(\wtparamhatML)$ via Equation~\eqref{EqnInduce}.
Let us now upper bound the mean-squared error associated with this
estimator.

Our observation model can be (re)described in the following way.
Consider an Erd\H{o}s-R\'enyi graph on $\numitems$ vertices with each
edge drawn independently with a probability $\pp$. For each edge in
this graph, we obtain one observation of the pair of vertices at the
end-points of that edge. Let $\LapErdos$ be the (random) Laplacian
matrix of this graph, that is, $\LapErdos = D - A$ where $D$ is an
$(\numitems \times \numitems)$ diagonal matrix with $[D]_{ii}$ being
the degree of item $i$ in the graph (equivalently, the number of
pairwise comparison observations that involve item $i$) and $A$ is the
$(\numitems \times \numitems)$ adjacency matrix of the graph. Let
$\eigenvalue{2}{\LapErdos}$ denote the second largest eigenvalue of
$\LapErdos$.  From Theorem 2(b) of our paper~\cite{shah2015estimation}
on estimating parametric models,\footnote{Note that the Laplacian
  matrix used in the statement of~\cite[Theorem
    2(b)]{shah2015estimation} is a scaled version of the matrix
  $\LapErdos$ introduced here, with each entry of $\LapErdos$ divided
  by the total number of observations.} for this graph, there is a
universal constant $\plaincon_1$ such that the maximum likelihood
estimator $\wtparamhatML$ has mean squared error upper bounded as
\begin{align*}
\Exs [ \myLnorm{\wtparamhatML - \wtparamstar}{2}^2 \mid \LapErdos]
\leq \plaincon_1 \frac{ \numitems }{ \eigenvalue{2}{\LapErdos}}.
\end{align*}
The estimator $\wtparamhatML$ is computable in a time polynomial in $\numitems$.

Since $\pp \geq \plaincon_0 \frac{(\log \numitems)^2}{\numitems}$,
known results on the eigenvalues of random
graphs~\cite{oliveira2009concentration,chung2011spectra, kolokolnikov2014algebraic}
imply that
\begin{align}
\mprob\Big[\eigenvalue{2}{\LapErdos} \geq \plaincon_2 \numitems
  \pp\Big] & \geq 1 - \frac{1}{\numitems^4}
\end{align}
for a universal constant $\plaincon_2$ (that may depend on
$\plaincon_0$).  As shown earlier in Equation~\eqref{EqnParamVecToMx},
for any valid score vectors $\wtparam^1$, $\wtparam^2$, we have
$\frobnorm{\wt(\wtparam^1)-\wt(\wtparam^2)}^2 \leq \numitems \zeta^2
\myLnorm{\wtparam^1 - \wtparam^2}{2}^2$ where $\zeta \defn \max_{z \in
  [-1,1]} \glmcdf'(z)$ is a constant independent of $\numitems$ and
$\pp$. Putting these results together and performing some simple
algebraic manipulations leads to the upper bound
\begin{align*}
\frac{1}{\numitems^2} \Exs \Big[ \frobnorm{\wt(\wtparamhatML) -
    \wtstar}^2 \Big ] & \leq \frac{\plaincon_3 \zeta^2}{\numitems
  \pp},
\end{align*}
which establishes the claim.


\section{Discussion}
\label{SecDiscussion}

In this paper, we analyzed a flexible model for pairwise comparison
data that includes various parametric models, including the BTL and
Thurstone models, as special cases.  We analyzed various estimators
for this broader matrix family, ranging from optimal estimators
to various polynomial-time estimators, including forms of
singular value thresholding, as well as a multi-stage method based on a
noisy sorting routine.  We show that this SST model permits far more robust estimation as compared to popular parametric models, while surprisingly, incurring little penalty for this significant generality.\footnote{In Appendix~\ref{AppModWeakTransitivity} we show that under weaker notions of stochastic transitivity, the pairwise-comparison probabilities are unestimable.} Our results thus present a strong motivation towards the use of such general stochastic transitivity based models. 

All of the results in this paper focused on
estimation of the matrix of pairwise comparison probabilities in the
Frobenius norm.  Estimation of probabilities in other metrics, such as
the KL divergence or estimation of the ranking in the Spearman's footrule or Kemeny distance, follow as corollaries of our results (see Appendix~\ref{SecOtherMetrics}). Establishing the best possible rates for polynomial-time
algorithms over the full class $\chatterjeeclass$ is a challenging
open problem.
 
We evaluated a computationally efficient estimator based on
thresholding the singular values of the observation matrix that is
consistent, but achieves a suboptimal rate. In our analysis of this
estimator, we have so far been conservative in our choice of the
regularization parameter, in that it is a fixed choice. Such a fixed
choice has been prescribed in various theoretical works on the soft or
hard-thresholded singular values (see, for instance, the
papers~\cite{chatterjee2014matrix,gavish2014optimal}).  In practice, the
entries of the effective noise matrix $\Wmat$ have variances that
depend on the unknown matrix, and the regularization parameter may be
obtained via cross-validation. The effect of allowing a data-dependent
choice of the regularization parameter remains to be studied, although
we suspect it may improve the minimax risk by a constant factor at
best.

Finally, in some applications, choices can be systematically
intransitive, for instance when objects have multiple features and
different features dominate different pairwise comparisons. In these
situations, the SST assumption may be weakened to one where the
underlying pairwise comparison matrix is a mixture of a small number of
SST matrices. The results of this work may form building blocks to
address this general setting; we defer a detailed analysis to future
work.


\paragraph{Acknowledgments:}
This work was partially supported by ONR-MURI grant DOD-002888, AFOSR
grant FA9550-14-1-0016, NSF grant CIF-31712-23800, and ONR MURI grant
N00014-11-1-0688. The work of NBS was also supported in part by a
Microsoft Research PhD fellowship.

\bibliographystyle{alpha_initials} \bibliography{bibtex}


\appendix
\begin{center}
\bf \large APPENDICES
\end{center}


\section{Relation to other error metrics}
\label{SecOtherMetrics}

In this section, we show how estimation of the
pairwise-comparison-probability matrix $\wtstar$ under the squared
Frobenius norm implies estimates and bounds under other error
metrics. In particular, we investigate relations between estimation of
the true underlying ordering under the Spearman's footrule and the
Kemeny metrics, and estimation of the matrix $\wtstar$ under the
Kullback-Leibler divergence metric.

\subsection{Recovering the true ordering}

Recall that the SST class assumes the existence of some true ordering
of the $\numitems$ items. The pairwise-comparison probabilities are
then assumed to be faithful to this ordering. In this section, we
investigate the problem of estimating this underlying ordering.

In order to simplify notation, we assume without loss of generality
that this true underlying ordering is the identity permutation of the $\numitems$
items, and denote the identity permutation as $\permid$. Recall the set $\bisoclass$ of \emph{bivariate isotonic matrices}, that is, SST matrices that are faithful to the identity permutation:
\begin{align*}
\bisoclass = \{ \wt \in [0,1]^{\numitem
  \times \numitem} \mid M_{ij} = 1-
M_{ji} \mbox{ for all $(i,j) \in [\numitems]^2$, and $M_{i \ell} \geq
M_{j \ell}$ whenever $i < j$.} \}
\end{align*} 
Then we have that $\wtstar \in \bisoclass$. Let $\perm$ be any permutation of the $\numitems$ items. For any matrix $\wt
\in \reals^{\numitems \times \numitems}$ and any integer $i \in
    [\numitems]$ we let $\wt_i$ denote the $i^{th}$ row of $\wt$.

Two of the most popular metrics of measuring the error between two
such orderings are the Spearman's footrule and the Kemeny (or Kendall
tau) distance, defined as follows. Spearman's footrule measures the
total displacement of all items in $\perm$ as compared to
$\permid$, namely
\begin{align*}
\mbox{Spearman's footrule}(\perm, \permid) \defn
\sum_{i=1}^{\numitems} \mid \perm(i) - i \mid.
\end{align*}
On the other hand, the Kemeny distance equals the total number of
pairs whose relative positions are different in the two orderings,
namely,
\begin{align*}
\mbox{Kemeny}(\perm, \permid) \defn \sum_{1 \leq i < j \leq \numitems} \indicator{ \sign(\perm(i) - \perm(j)) \neq \sign(i - j)},
\end{align*}
where ``$\sign$'' denotes the sign function, that is, $\sign(x) = 1$
if $x>0$, $\sign(x) = -1$ if $x<0$ and $\sign(x) = 0$ if $x=0$. The
Kemeny distance is also known as the Kendall tau metric.

Before investigating the two aforementioned metrics, we remark on one
important aspect of the problem of estimating the order of the
items. Observe that if the rows of $\wtstar$ corresponding to some
pair of items $(i,j)$ are very close to each other (say, in a
pointwise sense), then it is hard to estimate the relative position of
item $i$ with respect to item $j$. On the other hand, if the two rows
are far apart then differentiating between the two items is
easier. Consequently, it is reasonable to consider a metric that
penalizes errors in the inferred permutation based on the relative values of the rows of
$\wtstar$. To this end, we define a reweighted version of Spearman's
footrule as
\begin{align*}
\mbox{Matrix-reweighted Spearman's footrule}_{\wtstar}(\perm,
\permid) \defn \frobnorm{\perm(\wtstar) - \wtstar}^2 =
\sum_{i=1}^{\numitems} \Lnorm{\wtstar_{\perm(i)} -
  \wtstar_{i}}{2}^2.
\end{align*}

Given these definitions, the following proposition now relates the
squared Frobenius norm metric to the other aforementioned metrics.
\setcounter{proposition}{2}
\begin{subtheorem}{proposition}
\label{PropOtherMetrics}
 \setcounter{\subtheoremcounter}{0}
 \begin{proposition}
\label{PropWeighted}
Any two matrices $\wtstar \in \bisoclass$, and $\wt \in
\chatterjeeclass$ with $\perm$ as its underlying permutation, must satisfy the following bound on the
matrix-reweighted Spearman's footrule:
\begin{align*}
\frobnorm{\wtstar - \perm(\wtstar)}^2  \leq 4 \frobnorm{\wtstar - \wt}^2.
\end{align*}
\end{proposition}
\begin{proposition}
\label{PropSpearman}
Consider any matrix $\wtstar \in \bisoclass$ that satisfies
$\Lnorm{\wtstar_i - \wtstar_{i+1}}{2}^2 \geq \gamma^2$ for some constant
$\gamma>0$ and for every $i \in [\numitems-1]$. Then for any
permutation $\perm$, the Spearman's footrule distance from the
identity permutation is upper bounded as
\begin{align*}
\sum_{i=1}^{\numitems} \mid i - \perm(i) \mid \leq \frac{1}{\gamma^2} \frobnorm{\wtstar - \perm(\wtstar)}^2.
\end{align*}
Conversely, there exists a matrix $\wtstar \in \bisoclass$ that satisfies $\Lnorm{\wtstar_i - \wtstar_{i+1}}{2}^2 = \gamma^2$ for every $i \in [\numitems - 1]$  such that for every permutation $\perm$, the Spearman's footrule distance from the
identity permutation is lower bounded as
\begin{align*}
\sum_{i=1}^{\numitems} \mid i - \perm(i) \mid \geq \frac{1}{4\gamma^2} \frobnorm{\wtstar - \perm(\wtstar)}^2.
\end{align*}
\end{proposition}
\begin{proposition}[\cite{diaconis1977spearman}]
\label{PropKemeny}
The Kemeny distance of any permutation $\perm$ from the identity permutation $\permid$ is sandwiched as
\begin{align*}
\half \sum_{i=1}^{\numitems} \mid i - \perm(i) \mid \leq \sum_{1
  \leq i < j \leq \numitems} \indicator{ \sign(\perm(i) -
  \perm(j)) \neq \sign(i - j)} \leq \sum_{i=1}^{\numitems} \mid i -
\perm(i) \mid.
\end{align*}
\end{proposition}
\end{subtheorem}
As a consequence of this proposition, an upper bound on the error in
estimation of $\wtstar$ under the squared Frobenius norm yields
identical upper bounds (with some constant factors) under the other
three metrics.

\noindent A few remarks are in order:
\begin{enumerate}[leftmargin = *]
\item[(a)] Treating $\wtstar$ as the true pairwise comparison probability matrix and $\wt$ as its estimate, Proposition~\ref{PropWeighted} assumes that $\wt$ also lies in the
  matrix class $\chatterjeeclass$. This set-up is known as proper
  learning in some of the machine learning literature.
\item[(b)] The $\gamma$-separation condition of
  Proposition~\ref{PropSpearman} is satisfied in the models assumed in
  several earlier works~\cite{braverman2008noisy,
    wauthier2013efficient}.
\end{enumerate}

The remainder of this subsection is devoted to the proof of these
claims.


\subsubsection{Proof of Proposition~\ref{PropWeighted}} 

For any matrix $\wt$ and any permutation $\perm$ of $\numitems$ items,
let $\perm(\wt)$ denote the matrix resulting from permuting the rows
of $\wt$ by $\perm$. With this notation, we have
\begin{align*}
\frobnorm{ \perm(\wtstar) - \wtstar}^2 \;\leq 2 \frobnorm{
  \perm(\wtstar) - \wt}^2 + 2\frobnorm{ \wt - \wtstar }^2 
& = 2 \frobnorm{ \wtstar - \perm^{-1}(\wt)}^2 + 2\frobnorm{
  \wt - \wtstar }^2.
\end{align*}
We now show that
\begin{align}
\label{EqSamePermGood}
\frobnorm{ \wtstar - \perm^{-1}(\wt)}^2 \leq \frobnorm{\wtstar -
  \wt}^2,
\end{align}
which would then imply the claimed result.  As shown below, the
inequality~\eqref{EqSamePermGood} is a consequence of the fact that
$\wtstar$ and $\perm^{-1}(\wthat)$ both lie in the SST class and
have the same underlying ordering of the rows.  More generally, we
claim that for any two matrices $\wt \in \bisoclass$ and $\wt' \in
\bisoclass$,
\begin{align}
\permid \in \argmin_{\widetilde{\perm}} \frobnorm{\wt - \widetilde{\perm}(\wt')}^2,
\label{EqSamePermBest}
\end{align}
where the minimization is carried out over all permutations of
$\numitems$ items. To see this, consider any two matrices $\wt$ and
$\wt'$ in $\bisoclass$ and let $\perm'$ be a minimizer of
$\frobnorm{\wt - \perm(\wt')}^2$. If $\perm' \neq \permid$, then there
must exist some item $i \in [\numitems-1]$ such that item $(i+1)$ is
ranked higher than item $i$ in $\perm'$. Consequently,
\begin{align*}
\Lnorm{\wt_i - \wt'_{i+1}}{2}^2  + \Lnorm{\wt_{i+1} - \wt'_{i} }{2}^2 -
\Lnorm{\wt_i & - \wt'_i}{2}^2  - \Lnorm{\wt_{i+1} - \wt'_{i+1}}{2}^2 \\
& =
2 \tracer{ \wt_i - \wt_{i+1} }{ \wt'_i - \wt'_{i+1} } \geq 0,
\end{align*}
where the final inequality follows from the fact that $\wt \in
\bisoclass$ and $\wt' \in \bisoclass$. It follows that the new
permutation obtained by swapping the positions of items $i$ and
$(i+1)$ in $\perm'$ (which now ranks item $i$ higher than item
$(i+1)$) is also a minimizer of $\frobnorm{\wt - \perm(\wt')}^2$. A
recursive application of this argument yields that $\permid$ is also a
minimizer of $\frobnorm{\wt - \perm(\wt')}^2$.


\subsubsection{Proof of Proposition~\ref{PropSpearman}}
We first prove the upper bound on the Spearman's footrule metric. Due
to the monotonicity of the rows and the columns of $\wtstar$, we have
the lower bound
\begin{align*}
\frobnorm{\wtstar - \perm(\wtstar)}^2 \geq \sum_{\ell = 1
}^{\numitems} \Lnorm{\wtstar_\ell - \wtstar_{\perm(\ell)}}{2}^2.
\end{align*}
Now consider any $\ell \in [\numitems]$ such that $\perm(\ell) >
\ell$. Then we have
\begin{align*}
\Lnorm{\wtstar_\ell - \wtstar_{\perm(\ell)}}{2}^2 = \Lnorm{ \sum_{i =
    \ell}^{\perm(\ell)-1} (\wtstar_i - \wtstar_{i+1}) }{2}^2 \stackrel{(i)}{\geq}
\sum_{i = \ell}^{\perm(\ell)-1} \Lnorm{ \wtstar_i - \wtstar_{i+1}
}{2}^2 \stackrel{(ii)}{\geq} \gamma^2 |\perm(i) - i|,
\end{align*}
where the inequality (i) is a consequence of the fact that for every
$i \in [\numitems-1]$, every entry of the vector $(\wtstar_i -
\wtstar_{i+1})$ is non-negative, and the inequality (ii) results from the assumed $\gamma$-separation
condition on the rows of $\wtstar$. An
identical argument holds when $\perm(\ell) < \ell$. This argument completes the proof of the upper bound.

We now move on to the lower bound on Spearman's footrule. To this end, consider the matrix $\wtstar \in \bisoclass$ with its entries given as:
\begin{align*}
[\wtstar]_{ij} =
\begin{cases}
\frac{1}{2} + \frac{\gamma}{\sqrt{2}} & \quad \mbox{if $i<j$}\\
\frac{1}{2}  & \quad \mbox{if $i=j$}\\
\frac{1}{2} - \frac{\gamma}{\sqrt{2}} & \quad \mbox{if $i>j$}.
\end{cases}
\end{align*}
One can verify that this matrix $\wtstar$ satisfies the required condition $\Lnorm{\wtstar_i - \wtstar_{i+1}}{2}^2 = \gamma^2$ for every $i \in [\numitems - 1]$. One can also compute that this matrix also satisfies the condition $\frobnorm{\wtstar - \perm(\wtstar)} = 4\gamma^2 \sum_{\ell = 1}^{\numitems} | \ell - \perm(\ell)|$, thereby yielding the claim.

\subsubsection{Proof of Proposition~\ref{PropKemeny}}

 It is well known~\cite{diaconis1977spearman} that the Kemeny distance
 and Spearman's footrule distance between two permutation lie within a
 factor of $2$ of each other.


\subsection{Estimating comparison probabilities under Kullback-Leibler 
divergence} 
\label{AppKL}

Let $\mprob_\wt$ denote the probability distribution of the
observation matrix $Y \sim \{0,1\}^{\numitems \times \numitems}$
obtained by independently sampling entry $Y_{ij}$ from a Bernoulli
distribution with parameter $\wt_{ij}$. The Kullback-Leibler (KL)
divergence between $\mprob_\wt$ and $\mprob_{\wt'}$ is given by
\begin{align*}
\kl{\mprob_{\wt}}{\mprob_{\wt'}} = \wt_{ij} \log \frac{ \wt_{ij} }{
  \wt'_{ij} } + ( 1 - \wt_{ij} ) \log \frac{ 1 - \wt_{ij} }{ 1 -
  \wt'_{ij} } .
\end{align*}

Before we establish the connection with the squared Frobenius norm, we
make one assumption on the pairwise comparison probabilities that is
standard in the literature on estimation from pairwise
comparisons~\cite{negahban2012iterative, hajek2014minimax,
  shah2015estimation, chen2015spectral}. We assume that every entry of
$\wtstar$ is bounded away from $\{0,1\}$. In other words, we assume
the existence of some known constant-valued parameter $\epsilon \in
(0,\half]$ whose value is independent of $\numitems$, such that
  $\wtstar_{ij} \in (\epsilon, 1 - \epsilon)$ for every pair $(i,j)$.
  Given this assumption, for any estimator $\wt$ of $\wtstar$, we
  clip each of its entries and force them to lie in the interval
  $(\epsilon, 1 - \epsilon)$.\footnote{This clipping step does not
    increase the estimation error.} The following proposition then
  relates the Kullback-Leibler divergence metric to estimation under
  the squared Frobenius norm.
\begin{proposition}
\label{PropSandwich}
The probability distributions induced by any two probability matrices
$\wtstar$ and $\wt$ must satisfy the sandwich inequalities:
\begin{align*}
\frobnorm{\wt - \wtstar}^2 \leq
\kl{\mprob_{\wt}}{\mprob_{\wtstar}} \leq
\frac{1}{\epsilon(1-\epsilon)} \frobnorm{\wt - \wtstar}^2,
\end{align*}
where for the upper bound we have assumed that every entry of the
matrices lies in $(\epsilon, 1 - \epsilon)$.
\end{proposition}
The proof of the proposition follows from standard upper and lower
bounds on the natural logarithm~\eqref{EqnKLFrob3}. As a consequence of this result, any upper or
lower bound on $\frobnorm{\wt - \wtstar}^2$ therefore automatically
implies an identical upper or lower bound on
$\kl{\mprob_{\wt}}{\mprob_{\wtstar}}$ up to constant factors.


\section{Proof of Proposition~\ref{PropParametricBreak}}
\label{AppPropParametricBreak}
We show that the matrix $\wtstar$ specified in
Figure~\ref{FigParametricBreak}a satisfies the conditions required by
the proposition.  It is easy to verify that $\wtstar \in
\chatterjeeclass$, so that it remains to prove the
approximation-theoretic lower bound~\eqref{EqnParametricBreak}.  In
order to do so, we require the following auxiliary result:
\begin{lemma}
\label{lem:parametric_necessary}
Consider any matrix $\wt$ that belongs to $\paramclass(\glmcdf)$ for a
valid function $\glmcdf$. Suppose for some collection of four distinct
items $\{i_1, \ldots, i_4\}$, the matrix $\wt$ satisfies the
inequality $\wt_{i_1 i_2} > \wt_{i_3 i_4}$. Then it must also satisfy
the inequality $\wt_{i_1 i_3} \geq \wt_{i_2 i_4}$.
\end{lemma}

We return to prove this lemma at the end of this section.  Taking it
as given, let us now proceed to prove the lower
bound~\eqref{EqnParametricBreak}.  For any valid $\glmcdf$, fix an
arbitrary member $\wt$ of a class $\paramclass(\glmcdf)$, and let
$\wtparam \in \real^\numitems$ be the underlying weight vector (see
the definition~\eqref{EqnInduce}).

 Pick any item in the set of first $\frac{\numitems}{4}$ items
 (corresponding to the first $\frac{\numitems}{4}$ rows of $\wtstar$)
 and call this item as ``$1$''; pick an item from the next set of
 $\frac{\numitems}{4}$ items (rows) and call it item ``$2$''; item
 ``$3$'' from the next set and item ``$4$'' from the final set.  Our
 analysis proceeds by developing some relations between the pairwise
 comparison probabilities for these four items that must hold for
 every parametric model, that are strongly violated by $\wtstar$.  We
 divide our analysis into two possible relations between the entries
 of $\wt$.

\noindent {\underline{Case I:}} First suppose that $\wt_{12} \leq
\wt_{34}$.  Since $\wtstar_{12} = 6/8$ and $\wtstar_{34}=5/8$ in our
construction, it follows that
\begin{align*}
(\wt_{12} - \wtstar_{12})^2 + (\wt_{34} - \wtstar_{34} )^2 \geq
  \frac{1}{256}.
\end{align*}

\noindent {\underline{Case II:}} Otherwise, we may assume that
$\wt_{12} > \wt_{34}$. Then Lemma~\ref{lem:parametric_necessary}
implies that \mbox{$\wt_{13} \geq \wt_{24}$.}  Moreover, since
$\wtstar_{13} = 7/8$ and $\wtstar_{24}= 1$ in our construction, it
follows that
\begin{align*}
(\wt_{13} - \wtstar_{13})^2 + (\wt_{24} - \wtstar_{24} )^2 \geq
  \frac{1}{256}.
\end{align*}

\vspace*{.05in} 
Aggregating across these two exhaustive cases, we find that 
\begin{align*}
\sum \limits_{(u,v) \in \{1,2,3,4\}} (\wt_{uv} - \wtstar_{uv})^2 \geq
\frac{1}{256}.
\end{align*}
Since this bound holds for any arbitrary selection of items from the
four sets, we conclude that $\frac{1}{\numitems^2} \frobnorm{\wt -
  \wtstar}^2$ is lower bounded by a universal constant $\plaincon > 0$
as claimed.

Finally, it is easy to see that upon perturbation of any of the
entries of $\wtstar$ by at most $\frac{1}{32}$---while still ensuring
that the resulting matrix lies in $\chatterjeeclass$---the
aforementioned results continue to hold, albeit with a worse
constant. Every matrix in this class satisfies the claim of this
proposition.


\paragraph{Proof of Lemma~\ref{lem:parametric_necessary}:}

It remains to prove Lemma~\ref{lem:parametric_necessary}. Since $\wt$
belongs to the parametric family, there must exist some valid function
$\glmcdf$ and some vector $\wtparam$ that induce $\wt$ (see
Equation~\eqref{EqnInduce}). Since $\glmcdf$ is non-decreasing, the
condition $\wt_{i_1 i_2} > \wt_{i_3 i_4}$ implies that 
\begin{align*}
\wtparam_{i_1} - \wtparam_{i_2} > \wtparam_{i_3} - \wtparam_{i_4}.
\end{align*}
Adding $\wtparam_{i_2} - \wtparam_{i_3}$ to both sides of this
inequality yields $\wtparam_{i_1} - \wtparam_{i_3} > \wtparam_{i_2} -
\wtparam_{i_4}$.  Finally, applying the non-decreasing function
$\glmcdf$ to both sides of this inequality gives yields $\wt_{i_1 i_3}
\geq \wt_{i_2 i_4}$ as claimed, thereby completing the proof.


\section{Minimizing feedback arc set over entire SST class}\label{SecFASoverAllSST}
 Our analysis in Theorem~\ref{ThmHighSNR} shows that the two-step estimator proposed in Section~\ref{SecHighSNR} works well under the stated bounds
on the entries of $\Mstar$, i.e., for $\Mstar \in
\chatterjeeclassbias(\bias)$ for a fixed $\bias$.  This two-step estimator
is based on finding a minimum feedback arc set (FAS) in the first
step. In this section, we investigate the efficacy of estimators based on minimum FAS over the full class $\chatterjeeclass$. We show that minimizing the FAS does not
work well over $\chatterjeeclass$.

The intuition is that although minimizing the feedback arc set appears
to minimize disagreements at a global scale, it makes only local
decisions: if it is known that items $i$ and $j$ are in adjacent
positions, the order among these two items is decided based solely on
the outcome of the comparison between items $i$ and $j$, and is
independent of the outcome of the comparisons of $i$ and $j$ with all
other items.

Here is a concrete example to illustrate this property. Suppose
$\numitems$ is divisible by $3$, and consider the following
$(\numitems \times \numitems)$ block matrix $\wt \in
\chatterjeeclass$:
\begin{align*}
\wt = \renewcommand\arraystretch{1.2}
\begin{bmatrix}~
\half~ & \half & 1 \\ \half & \half & \frac{3}{4} \\ 0 & \frac{1}{4} &
~\half ~\end{bmatrix},
\end{align*}
where each block is of size $(\frac{\numitems}{3} \times
\frac{\numitems}{3})$. Let $\perm^1$ be the identity permutation, and
let $\perm^2$ be the permutation $[\frac{\numitems}{3} + 1, \ldots ,
  \frac{2\numitems}{3},1,\ldots,\frac{\numitems}{3},\frac{2\numitems}{3}+1,\ldots,\numitems]$, that is, $\perm^2$ swaps the second block of $\frac{\numitems}{3}$ items with the first block. For
any permutation $\perm$ of the $\numitems$ items and any $\wt \in
\chatterjeeclass$, let $\perm(\wt)$ denote the $(\numitems \times
\numitems)$ matrix resulting from permuting both the rows and the
columns by $\perm$.

One can verify that $\frobnorm{\perm^1(\wt) - \perm^2(\wt)}^2 \geq \plaincon \numitems^2$, for some universal constant $\plaincon > 0$. Now suppose an observation $\obs$ is generated
from $\perm^1(\wt)$ as per the model~\eqref{EqnObservationModel}. Then the
distribution of the size of the feedback arc set of $\perm^1$ is
identical to the distribution of the size of the feedback arc set of
$\perm^2$. Minimizing FAS cannot distinguish between $\perm^1(\wt)$
and $\perm^2(\wt)$ at least $50\%$ of the time, and consequently, any
estimator based on the minimum FAS output cannot perform well over the
SST class.


\section{Relation to other models}\label{AppRelationsModels}
We put things in perspective to the other models considered in the
literature. We begin with two weaker versions of stochastic
transitivity that are also investigated in the literature on
psychology and social science.

\subsection{Moderate and weak stochastic transitivity} 
\label{AppModWeakTransitivity}

The model $\chatterjeeclass$ that we consider is called
strong stochastic transitivity in the literature on psychology and
social science~\cite{fishburn1973binary,davidson1959experimental}. The
two other popular (and weaker) models are those of \emph{moderate
  stochastic transitivity} $\modStocTranClass$ defined
as \begin{align*} \modStocTranClass \defn \left\{ \wt \in
  [0,1]^{\numitems \times \numitems}~\vert~\mbox{$\wt_{ik} \geq
    \min\{\wt_{ij},\wt_{jk}\}$ for every $(i,j,k)$ satisfying
    $\wt_{ij} \geq \half$ and $\wt_{jk} \geq \half$} \right\},
\end{align*}
and \emph{weak stochastic transitivity} $\weakStocTranClass$ defined
as \begin{align*} \weakStocTranClass \defn \left\{ \wt \in
  [0,1]^{\numitems \times \numitems}~\vert~\mbox{$\wt_{ik} \geq \half$
    for every $(i,j,k)$ satisfying $\wt_{ij} \geq \half$ and $\wt_{jk}
    \geq \half$}\right\}.
\end{align*}
Clearly, we have the inclusions $\chatterjeeclass \subseteq
\modStocTranClass \subseteq \weakStocTranClass$.

In Theorem~\ref{ThmMinimax}, we prove that the minimax rates of
estimation under the strong stochastic transitivity assumption are
$\tilde{\Theta}(\numitems^{-1})$. It turns out, however, that the two
weaker transitivity conditions do not permit meaningful estimation.

\begin{proposition}
\label{prop:mod_weak_stoctran}
There exists a universal constant $\plaincon>0$ such that under the
moderate $\modStocTranClass$ stochastic transitivity model,
\begin{align*}
\inf_{\Mtil} \sup_{\Mstar \in \modStocTranClass} \frac{1}{\numitems^2}
\Exs [ \frobnorm{\Mtil - \Mstar}^2 ] > \plaincon.
\end{align*}
where the infimum is taken over all measurable mappings from the
observations $\obs$ to $[0,1]^{\numitems \times \numitems}$.
Consequently, for the weak stochastic transitivity model
$\weakStocTranClass$, we also have
\begin{align*}
\inf_{\Mtil} \sup_{\Mstar \in \weakStocTranClass}
\frac{1}{\numitems^2} \Exs [ \frobnorm{\Mtil - \Mstar}^2 ] >
\plaincon,
\end{align*}
\end{proposition}
The minimax risk over these two classes is clearly the worst possible
(up to a universal constant) since for any two arbitrary matrices
$\wt$ and $\wt'$ in $[0,1]^{\numitems \times \numitems}$, we have
$\frac{1}{\numitems^2} \frobnorm{\wt - \wt'}^2 \leq 1$. For this
reason, in the paper we restrict our analysis to the strong stochastic
transitivity condition.

\subsection{Comparison with statistical models}
\label{sec:relation_models_stats}

Let us now investigate relationship of the strong stochastic
transitivity model considered in this paper with two other popular
models in the literature on statistical learning from comparative
data. Perhaps the most popular model in this regard is the class of
parametric models $\paramclass$: recall that this class is defined as
\begin{align*}
\paramclass \defn \{ \wt \mid \wt_{ij} = \glmcdf(\wtparamstar_i -
\wtparamstar_j) ~~ & \mbox{for some non-decreasing function }  \glmcdf: \reals
\rightarrow [0,1], \\ \mbox{ and vector }\wtparamstar \in \reals^\numitems
\}.
\end{align*}
The parametric class of models assumes that the function $\glmcdf$ is
fixed and known. Statistical estimation under the parametric class is
studied in several recent
papers~\cite{negahban2012iterative,hajek2014minimax,shah2015estimation}. The
setting where the function $\glmcdf$ is fixed, but unknown leads to a
semi-parametric variant. The results presented in this section also
readily apply to the semi-parametric class.

The second class is that generated from distributions over complete
rankings~\cite{diaconis1989generalization,farias2013nonparametric,
  ding2014topic}. Specifically, every element in this class is
generated as the pairwise marginal of an arbitrary probability
distribution over all possible permutations of the $\numitems$
items. We denote this class as $\marginalclass$.

The following result characterizes the relation between the classes.
\begin{proposition}
\label{prop:model_relations}
Consider any value of $\numitems > 10$. The parametric class $\paramclass$ is a strict subset of the strong
stochastic transitivity class $\chatterjeeclass$. The class
$\marginalclass$ of marginals of a distribution on total rankings is
neither a subset nor a superset of either of the classes
$\chatterjeeclass$, $\paramclass$, and $\chatterjeeclass \backslash
\paramclass$.
\end{proposition}

\begin{figure}
\centering \includegraphics[width=.46\textwidth]{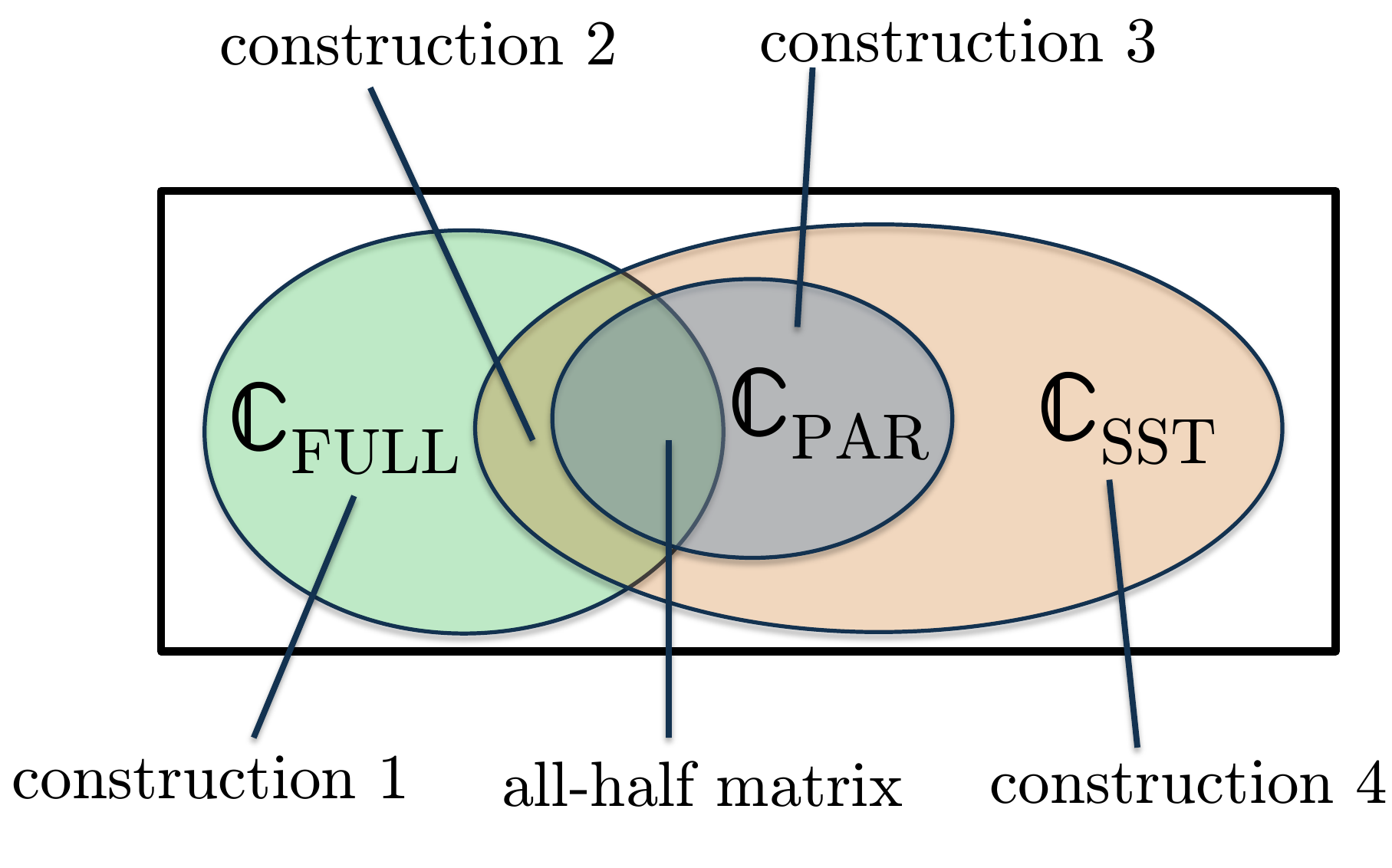}
\caption{Relations between various models of pairwise comparisons. The
  constructions proving these relations are presented as a part of the
  proof of Proposition~\ref{prop:model_relations}.}
\label{fig:model_relations}
\end{figure}

The various relationships in Proposition~\ref{prop:model_relations} are depicted pictorially in Figure~\ref{fig:model_relations}. These relations are
derived by first establishing certain conditions that matrices in the
classes considered must satisfy, and then constructing matrices that
satisfy or violate one or more of these conditions. The conditions on
$\marginalclass$ arise from the observation that the class is the
convex hull of all SST matrices that have their non-diagonal elements
in $\{0,1\}$; we derive conditions on this convex hull that leads to
properties of the $\marginalclass$ class. To handle the parametric
class $\paramclass$, we employ a necessary condition discussed earlier
in Section~\ref{SecParametricBad} and defined formally in
Lemma~\ref{lem:parametric_necessary}. The SST class
$\chatterjeeclass$ is characterized using the insights derived
throughout the paper.


\subsection{Proof of Proposition~\ref{prop:mod_weak_stoctran}}

We will derive an order one lower bound under the moderate
stochastic transitivity condition. This result automatically implies
the order one lower bound for weak stochastic transitivity.

The proof imposes a certain structure on a subset of the entries of
$\wtstar$ in a manner that $\Theta(\numitems^2)$ remaining entries are
free to take arbitrary values within the interval $[\half,1]$. This
flexibility then establishes a minimax error of $\Theta(1)$ as
claimed.

Let us suppose $\wtstar$ corresponds to the identity permutation of
the $\numitems$ items, and that this information is public
knowledge. Set the entries of $\wtstar$ above the diagonal in the
following manner. For every $i \in [\numitems]$ and every \emph{odd}
$j \in [\numitems]$, set $\wtstar_{ij} = \half$. For every $i \in
[\numitems]$ and every \emph{even} $j \in [\numitems]$, set
$\wtstar_{ji} = \half$. This information is also assumed to be public
knowledge. Let $\mathcal{S} \subset [\numitems]^2$ denote the set of
all entries of $\wtstar$ above the diagonal whose values were not
assigned in the previous step. Let $\cardinality{\mathcal{S}}$ denote
the size of set $\mathcal{S}$. The entries below the diagonal are
governed by the skew-symmetry constraints.

We first argue that every entry in $\mathcal{S}$ can take arbitrary
values in the interval $[\half,1]$, and are not constrained by each
other under the moderate stochastic transitivity condition. To this
end, consider any entry $(i,k) \in \mathcal{S}$. Recall that the
moderate stochastic transitivity condition imposes the following set
of restrictions in $\wtstar_{ik}$: for every $j$, $\wtstar_{ik} \geq
\min\{\wtstar_{ij},\wtstar_{jk}\}$. From our earlier construction we
have that for every odd value of $j$, $\wtstar_{ij} = \half$ and hence
the restriction simply reduces to $\wtstar_{ik} \geq \half$. On the
other hand, for every even value of $j$, our construction gives
$\wtstar_{jk} = \half$, and hence the restriction again reduces to
$\wtstar_{ik} \geq \half$. Given the absence of any additional
restrictions, the error $\Exs [ \frobnorm{\wthat - \wtstar}^2 ] \geq
\plaincon \cardinality{\mathcal{S}}$. Finally, observe that every
entry $(i,k)$ where $i < k$, $i$ is odd and $k$ is even belongs to the
set $\mathcal{S}$. It follows that $\cardinality{\mathcal{S}} \geq
\frac{\numitems^2}{8}$, thus proving our claim.


\subsection{Proof of Proposition~\ref{prop:model_relations}}

The constructions governing the claimed relations are enumerated in
Figure~\ref{fig:model_relations} and the details are provided below.

It is easy to see that since $\cdf$ is non-decreasing, the parametric
class $\paramclass$ is contained in the strong stochastic transitivity
class $\chatterjeeclass$. We provide a formal proof of this statement
for the sake of completeness. Suppose without loss of generality that
$\wtparam_1  \geq \cdots \geq \wtparam_\numitems$. Then we
claim that the distribution of pairwise comparisons generated through
this model result in a matrix, say $\wt$, that lies in the SST model
with the ordering following the identity permutation. This is because for any
$i>j>k$,
\begin{align*}
\wtparam_i - \wtparam_k &\geq \wtparam_i -
\wtparam_j\\ \cdf(\wtparam_i - \wtparam_k) &\geq \cdf(\wtparam_i -
\wtparam_j)\\ \wt_{ik} &\geq \wt_{ij}.
\end{align*}

We now show the remaining relations with the four constructions
indicated in Figure~\ref{fig:model_relations}. While these
constructions target some specific value of $\numitems$, the results
hold for any value $\numitems$ greater than that specific value. To
see this, suppose we construct a matrix $\wt$ for some $\numitems =
\numitems_0$, and show that it lies inside (or outside) one of these
classes. Consider any $\numitems > \numitems_0$, and define a
$(\numitems \times \numitems)$ matrix $\wt'$ as having $\wt$ as the
top-left $(\numitems_0 \times \numitems_0)$ block, $\half$ on the
remaining diagonal entries, $1$ on the remaining entries above the
diagonal and $0$ on the remaining entries below the diagonal. This
matrix $\wt'$ will retain the properties of $\wt$ in terms of lying
inside (or outside, respectively) the claimed class.

In this proof, we use the notation $i \succ j$ to represent a greater preference for $i$ as compared to $j$.

\subsubsection{Construction 1} 
 We construct a matrix $\wt$ such that $\wt \in \marginalclass$ but
$\wt \notin \chatterjeeclass$. Let $\numitems = 3$. Consider the
following distribution over permutations of $3$ items $(1,2,3)$:
\begin{eqnarray*}
\mprob(1 \succ 2 \succ 3) &= \frac{2}{5},\\
\mprob(3 \succ 1 \succ 2) &= \frac{1}{5}, \\
\mprob(2 \succ 3 \succ 1) &= \frac{2}{5}.
\end{eqnarray*}
This distribution induces the pairwise marginals
\begin{eqnarray*}
\mprob(1 \succ 2) &= \frac{3}{5}, \\ \mprob(2 \succ 3) &= \frac{4}{5},
\\ \mprob(3 \succ 1) &= \frac{3}{5}.
\end{eqnarray*}
Set $\wt_{ij} = \mathbb{P}(i \succ j)$ for every pair. By definition
of the class $\marginalclass$, we have $\wt \in \marginalclass$.

A necessary condition for a matrix $\wt$ to belong to the class
$\chatterjeeclass$ is that there must exist at least one item, say
item $i$, such that $\wt_{ij} \geq \half$ for every item $j$. One can
verify that the pairwise marginals enumerated above do not satisfy
this condition, and hence $\wt \notin \chatterjeeclass$.

\subsubsection{Construction 2} 

We construct a matrix $\wt$ such that $\wt \in \chatterjeeclass \cap
\marginalclass$ but $\wt \notin \paramclass$.  Let $\numitems=4$ and
consider the following distribution over permutations of $4$ items
$(1,2,3,4)$:
\begin{align*}
\mprob(3 \succ 1 \succ 2 \succ 4) = \frac{1}{8}, & \qquad \mprob(1 \succ
2 \succ 4 \succ 3) = \frac{1}{8} \\ 
\mprob(2 \succ 1 \succ 4 \succ 3) = \frac{2}{8} & \quad \mbox{and}
\quad \mprob(1 \succ 2 \succ 3 \succ 4) = \frac{4}{8}.
\end{align*}
One can verify that this distribution leads to the following pairwise
comparison matrix $\wt$ (with the ordering of the rows and columns
respecting the permutation $1 \succ 2 \succ 3 \succ 4$):
\begin{align*}
\wt \defn \frac{1}{8}
\begin{bmatrix}
4 & 6 & 7 & 8\\
2 & 4 & 7 & 8\\
1 & 1 & 4 & 5\\
0 & 0 & 3 & 4
\end{bmatrix}.
\end{align*}
It is easy to see that this matrix $\wt \in \chatterjeeclass$, and by
construction $\wt \in \marginalclass$. Finally, the proof of Proposition~\ref{prop:mod_weak_stoctran} shows that $\wt \notin \paramclass$, thereby completing the proof. 

\subsubsection{Construction 3} 

We construct a matrix $\wt$ such that $\wt \in \paramclass$ (and hence
$\wt \in \chatterjeeclass$) but $\wt \notin \marginalclass$. First
observe that any total ordering on $\numitems$ items can be
represented as an $(\numitems \times \numitems)$ matrix in the SST
class such that all its off-diagonal entries take values in
$\{0,1\}$. The class $\marginalclass$ is precisely the convex hull of
all such binary SST matrices.

Let $B^1,\ldots,B^{\factorial{\numitems}}$ denote all $(\numitems
\times \numitems)$ matrices in $\chatterjeeclass$ whose off-diagonal
elements are restricted to take values in the set $\{0,1\}$.  The
following lemma derives a property that any matrix in the convex hull
of $B^1,\ldots,B^{\factorial{\numitems}}$ must satisfy.
\begin{lemma}
\label{lem:binaryhullconstraint}
Consider any $\wt \in \chatterjeeclass$, and consider three items
$i,j,k \in [\numitems]$ such that $\wt$ respects the ordering
$i \succ j \succ k$. Suppose $\wt_{ij}=\wt_{jk}=\half$ and $\wt_{ik}=1$. Further
suppose that $\wt$ can be written as
\begin{align}
\wt = \sum_{\ell \in [\factorial{\numitems}]} \alpha^\ell B^\ell,
\end{align}
where $\alpha^\ell \geq 0~\forall~\ell$ and
$\sum_{\ell=1}^{\factorial{\numitems}} \alpha^\ell = 1$. Then for any
$\ell \in [\factorial{\numitems}]$ such that $\alpha^\ell > 0$, it
must be that $B_{ij}^\ell \neq B_{jk}^\ell$.
\end{lemma}
The proof of the lemma is provided at the end of this section.

Now consider the following $(7 \times 7)$ matrix $\wt \in
\chatterjeeclass$:
\begin{align}
\renewcommand\arraystretch{1.2} \wt \defn \left[
\begin{array}{ccccccc}
\colordiag& \half &1&1&1&1&1\\ \half & \colordiag & \half & \half & 1
& 1 & 1\\ 0& \half & \colordiag & \half & \half & 1 & 1 \\ 0& \half &
\half & \colordiag & \half & \half & 1 \\ 0& 0 & \half & \half &
\colordiag & \half & 1 \\ 0& 0 & 0 & \half & \half & \colordiag &
\half \\ 0& 0 &0 & 0 & 0 & \half & \colordiag
\end{array}
\right].
\label{eq:binaryhullconstraint0}
\end{align}
We will now show via proof by contradiction that $\wt$ cannot be
represented as a convex combination of the matrices
$B^1,\ldots,B^{\factorial{\numitems}}$. We will then show that $\wt
\in\paramclass$.

Suppose one can represent $\wt$ as a convex combination $\wt =
\sum_{\ell \in [\factorial{\numitems}]} \alpha^\ell B^\ell$, where
$\alpha^1,\ldots,\alpha^{\factorial{\numitems}}$ are non-negative
scalars that sum to one. Consider any $\ell$ such that $\alpha^\ell
\neq 0$. Let $B^\ell_{12} = b \in \{0,1\}$. Let us derive some more
constraints on $B^\ell$. Successively applying
Lemma~\ref{lem:binaryhullconstraint} for the following values of
$i,j,k$ implies that $B^\ell$ must necessarily have the
form~\eqref{eq:binaryhullconstraint1} shown below. Here $\bar{b} \defn
1-b$ and `$*$' denotes some arbitrary value that is irrelevant to the
discussion at hand.
\begin{itemize}
\item $i=1,j=2,k=3$ gives $B^\ell_{23} = \bar{b}$
\item $i=1,j=2,k=4$ gives $B^\ell_{24} = \bar{b}$
\item $i=2,j=3,k=5$ gives $B^\ell_{35} = {b}$
\item $i=2,j=4,k=6$ gives $B^\ell_{46} = {b}$
\item $i=3,j=5,k=6$ gives $B^\ell_{56} = \bar{b}$
\item $i=4,j=6,k=7$ gives $B^\ell_{67} = \bar{b}$.
\end{itemize}
Thus $B^\ell$ must be of the form
\begin{align}
\renewcommand\arraystretch{1.2}B^\ell = \left[
\begin{array}{ccccccc}
\colordiag& b &1&1&1&1&1\\
\half & \colordiag & \bar{b} & \bar{b} & 1 & 1 & 1\\
0& b & \colordiag & * & b & 1 & 1 \\
0& b & * & \colordiag & * & b & 1 \\
0& 0 & \bar{b} & * & \colordiag & \bar{b} & 1 \\
0& 0 & 0 & \bar{b} & b & \colordiag & \bar{b} \\
0& 0 &0 & 0 & 0 & b & \colordiag
\end{array}
\right].
\label{eq:binaryhullconstraint1}
\end{align}
Finally, applying Lemma~\ref{lem:binaryhullconstraint} with
$i=5$, $j=6$ and $k=7$ implies that $B^\ell_{67} = b$, which contradicts the
necessary condition in equation~\eqref{eq:binaryhullconstraint1}. We have thus
shown that $\wt \notin \marginalclass$.

We now show that the matrix $\wt$ constructed
in equation~\eqref{eq:binaryhullconstraint0} is contained in the class
$\paramclass$.  Consider the following function $\cdf:[-1,1]
\rightarrow [0,1]$ in the definition of a parametric class:
\begin{align*}
\cdf(x) =
\begin{cases}
0 & \qquad \mbox{if $x<-0.25$}\\
\half & \qquad \mbox{if $-0.25 \leq x \leq 0.25$}\\
1  & \qquad \mbox{if $x > 0.25$}.
\end{cases}
\end{align*}
Let $\numitems=7$ with $w_1=.9$, $w_2=.7$, $w_3=.6$, $w_4=.5$,
$w_5=.4$, $w_6=.3$ and $w_7=.1$. One can verify that under this
construction, the matrix of pairwise comparisons is identical to that
in equation~\eqref{eq:binaryhullconstraint0}.


\paragraph{Proof of Lemma~\ref{lem:binaryhullconstraint}}
In what follows, we show that $\sum_{\ell:B_{ij}^\ell=1, B_{jk}^\ell=1} \alpha^\ell = \sum_{\ell:B_{ij}^\ell=1, B_{jk}^\ell=1} \alpha^\ell = 0$. The result then follows immediately.

Consider some $\ell' \in [\factorial{\numitems}]$ such that
$\alpha^{\ell'} > 0$ and $B_{ij}^{\ell'}=0$. Since $\wt_{ik}=1$, we
must have $B_{ik}^{\ell'}=1$. Given that $B^{\ell'}$ represents a total ordering of the $\numitems$ items, that is, $B^{\ell'}$ is an SST matrix with boolean-valued its off-diagonal elements, $B_{ij}^{\ell'}=0$ and $B_{ik}^{\ell'}=1$ imply that $B_{jk}^{\ell'}=1$. We have thus shown that $B_{jk}^{\ell'}=1$ whenever $B_{ij}^{\ell'}=0$. This result has two consequences. The first consequence is that $\sum_{\ell:B_{ij}^\ell=0, B_{jk}^\ell=0} \alpha^\ell = 0$. The second consequence employs the additional fact that $\wt_{ij}=\half$ and hence $\sum_{\ell:B_{ij}^\ell=0} \alpha^\ell = \half$, and then gives $\sum_{\ell:B_{ij}^\ell=0, B_{jk}^\ell=1}
\alpha^\ell = \half$. Building on, we have \begin{align*} \half = \wt_{jk}= \sum_{\ell:B_{ij}^\ell=0, B_{jk}^\ell=1} \alpha^\ell   + \sum_{\ell:B_{ij}^\ell=1, B_{jk}^\ell=1} \alpha^\ell, \end{align*} and hence we have $\sum_{\ell:B_{ij}^\ell=1, B_{jk}^\ell=1} \alpha^\ell = 0$, thus completing the proof.

\subsubsection{Construction 4}
We construct a matrix $\wt$ such that $\wt \in \chatterjeeclass$ but
$\wt \notin \marginalclass$ and $\wt \notin \paramclass$. Consider
$\numitems = 11$. Let $\wt_2$ denote the $(4 \times 4)$ matrix of
Construction 2 and let $\wt_3$ denote the $(7 \times 7)$ matrix of
construction $3$. Consider the $(11 \times 11)$ matrix $\wt$ of the
form
\begin{align*}
\wt \defn
\begin{bmatrix}
\wt_2 & 1\\
0 & \wt_3
\end{bmatrix}.
\end{align*}
Since $\wt_2 \in \chatterjeeclass$ and $\wt_3\in \chatterjeeclass$, it
is easy to see that $\wt \in \chatterjeeclass$. Since $\wt_2 \notin
\paramclass$ and $\wt \notin \marginalclass$, it follows that $\wt
\notin \paramclass$ and $\wt \notin \marginalclass$. This construction completes the proof of Proposition~\ref{prop:model_relations}.


\end{document}